\documentclass[10pt,twocolumn,letterpaper]{article}

\usepackage[pagenumbers]{cvpr}              %

\usepackage{multirow}
\usepackage{booktabs}
\usepackage{amsmath,amsfonts,amssymb}
\usepackage{pifont}

\usepackage{graphicx}   %
\usepackage{subcaption} %
\usepackage{adjustbox}
\usepackage{array}

\usepackage[dvipsnames]{xcolor}

\definecolor{cvprblue}{rgb}{0.21,0.49,0.74}
\usepackage[pagebackref,breaklinks,colorlinks,citecolor=cvprblue]{hyperref}

\DeclareMathOperator*{\argmax}{arg\,max}

\newcommand{\cmark}{\ding{51}}
\newcommand{\xmark}{\ding{55}}

\usepackage{algorithm2e}
\RestyleAlgo{ruled}

\SetKwComment{Comment}{/* ----- }{ ----- */}

\def\paperID{3504} %
\def\confName{CVPR}
\def\confYear{2025}

\title{Point-Cache: Test-time Dynamic and Hierarchical Cache for Robust and Generalizable Point Cloud Analysis}

\author{
    Hongyu Sun$^{1,2}$, Qiuhong Ke$^{2}$, Ming Cheng$^{2}$, Yongcai Wang$^{1}\thanks{Corresponding author.}$, Deying Li$^{1}$, Chenhui Gou$^{2}$, Jianfei Cai$^{2}$\\
    $^{1}$Department of Computer Science, Renmin University of China, China\\
    $^{2}$Department of Data Science \& AI, Monash University, Australia\\
    {\tt\small \{sunhongyu,ycw,deyingli\}@ruc.edu.cn \quad \{qiuhong.ke,ming.cheng,jianfei.cai\}@monash.edu}
}

\begin{document}
\maketitle

\begin{abstract}
    This paper proposes a general solution to enable point cloud recognition models to handle distribution shifts at test time. Unlike prior methods, which rely heavily on training data (often inaccessible during online inference) and are limited to recognizing a fixed set of point cloud classes predefined during training, we explore a more practical and challenging scenario: adapting the model solely based on online test data to recognize both previously seen classes and novel, unseen classes at test time. 
To this end, we develop \textbf{Point-Cache}, a hierarchical cache model that captures essential clues of online test samples, particularly focusing on the global structure of point clouds and their local-part details. 
Point-Cache, which serves as a rich 3D knowledge base, is dynamically managed to prioritize the inclusion of high-quality samples.  
Designed as a plug-and-play module, our method can be flexibly integrated into large multimodal 3D models to support open-vocabulary point cloud recognition.  
Notably, our solution operates with efficiency comparable to zero-shot inference, as it is entirely training-free. Point-Cache demonstrates substantial gains across 8 challenging benchmarks and 4 representative large 3D models, highlighting its effectiveness. Code is available at \url{https://github.com/auniquesun/Point-Cache}.
\end{abstract}%
\section{Introduction}
\label{sec:intro}

In recent years, 3D sensors such as LiDARs and RGB-D cameras have been widely adopted in robotics and electric vehicles for their ability to provide reliable 3D geometry measurements 
~\cite{wang15large,wang19pseudo,fan21rangedet,guo21survey,wang23dsvt,wang23unitr,lu23open,huang24roco,wu24towards}. 
Point clouds are among the most direct data formats produced by these 3D sensors. Although remarkable progress has been 
made in 3D point cloud recognition, 
the success is primarily based on the assumption that the test data and the model training data are identically 
distributed~\cite{qi17pointnet,qi17pointnet2,li18pointcnn,wang19dgcnn,zhao21pt,ma22pointmlp,park23spotr,duan23condaformer,chen23pointgpt}. 
However, this assumption is frequently violated in real-world scenarios due to complex geometries, as well as sensing and processing errors~\cite{ren22modelnet-c,sun22modelnet-c,sun24point-prc}. 

\begin{figure}
    \centering
    \begin{subfigure}{0.5\linewidth}
        \centering
        \includegraphics[width=\columnwidth]{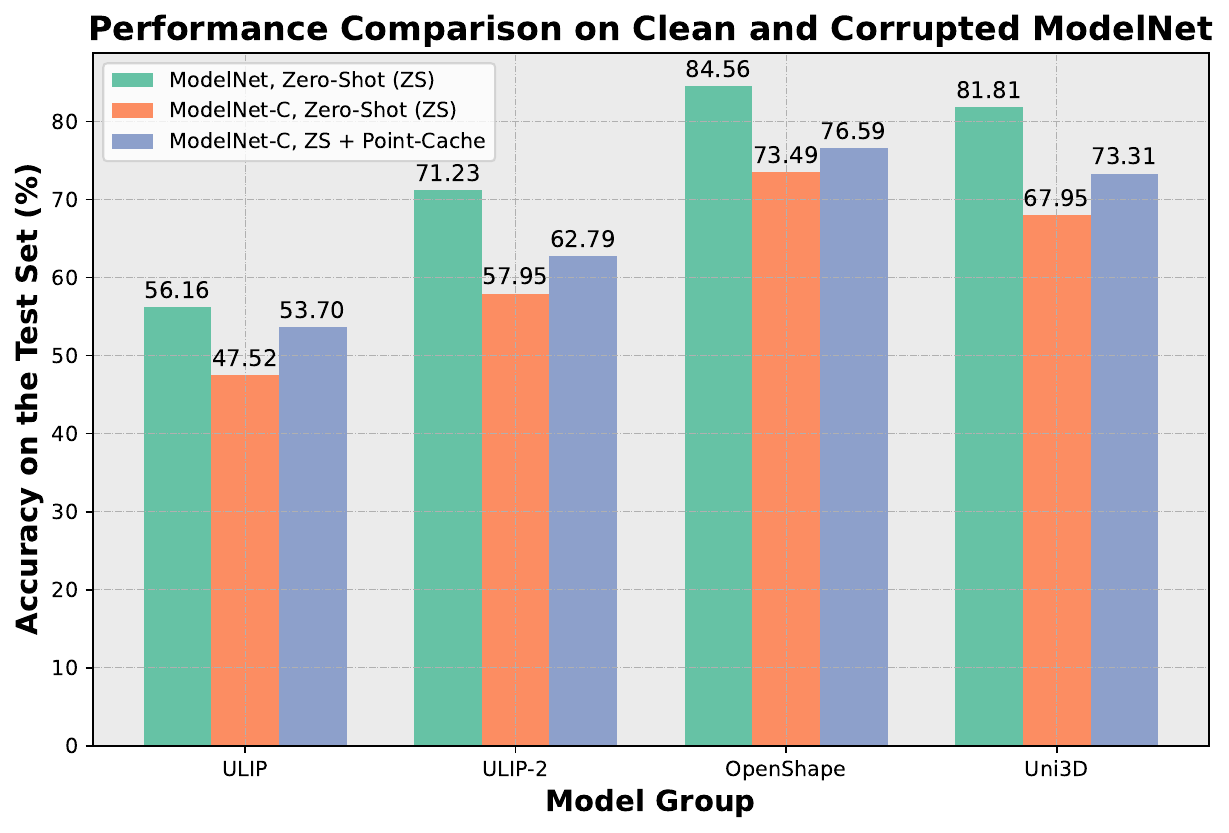}%
        \caption{ModelNet \& ModelNet-C.}
    \end{subfigure}%
    \begin{subfigure}{0.5\linewidth}
        \centering
        \includegraphics[width=\columnwidth]{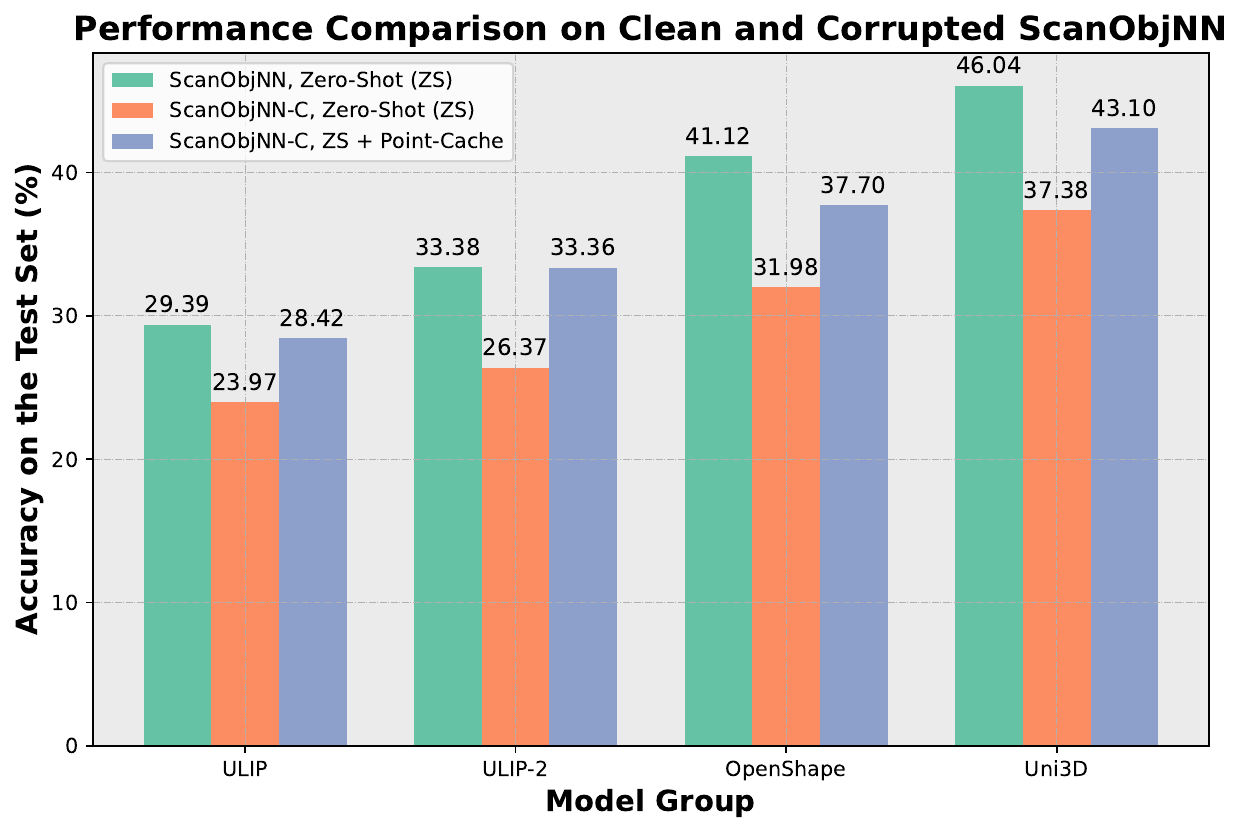}%
        \caption{ScanObjNN \& ScanObjNN-C.}
    \end{subfigure}
    \caption{\textbf{Recognition accuracy comparison on clean and corrupted point cloud datasets}. The suffix -C indicates  datasets 
    with corruptions. Models experience a severe performance drop when data corruptions arise. \textbf{Point-Cache} 
    effectively narrows down the performance gap between clean and corrupted data. The hardest split of ScanObjNN is used.}
    \label{fig:acc_on_clean_and_corrupted_data}
\end{figure}

In practice, a majority of models remain vulnerable to distribution shifts, such as out-of-distribution (OOD) samples~\cite{Objaverse,OmniObject3D,Objaversexl,sun24point-prc}, data 
corruptions~\cite{uy19sonn,ren22modelnet-c,ren22pointcloud-c,sun22modelnet-c}, and more. 
As shown in Fig.~\ref{fig:acc_on_clean_and_corrupted_data}, notable performance gaps (\eg, 11+\%) occur when models are tested on clean (ModelNet~\cite{wu15modelnet}) versus corrupted data (ModelNet-C~\cite{ren22modelnet-c}). 
These issues drive researchers to enhance the robustness and generalization of point cloud recognition methods. 
Related efforts can be grouped into three main streams. 
The first focuses on training models that can transfer to target data. 
These models learn a shared feature space between source and target domains~\cite{qin19pointdan,shen22da,fan22glrv,wei22pdg}, 
or apply meta-learning frameworks to handle geometry shifts by training multiple classifiers~\cite{huang21metasets}, 
or design learnable prompts to elicit general knowledge from pre-trained large 3D models~\cite{sun24point-prc}. 
However, possible shifts are unpredictable, and the training-based methods cannot account for all cases during training. 
As a result, these models are limited when new shifts arise at test time~\cite{yeo23rna}. 
The second stream addresses the problem through test-time adaptation strategies~\cite{mirza23mate,wang24bftt3d,shim24cloudfixer}. 
These strategies improve performance with various adaptation techniques, such as masked auto-encoding for point clouds (MATE~\cite{mirza23mate}), 
static prototype memory (BFTT3D~\cite{wang24bftt3d}), and diffusion models (CloudFixer~\cite{shim24cloudfixer}). 
Nevertheless, we find that these approaches rely heavily on training data. 
For instance, MATE~\cite{mirza23mate} requires pre-training on the entire training set, and BFTT3D~\cite{wang24bftt3d} employs  training data to build a static and offline prototype memory to store point cloud features. 
This heavy dependence on training data poses challenges, particularly when such data is unavailable or when distribution shifts occur at test time. In these cases, revisiting the training data may not be helpful to adapt the model to online test data. 
Additionally, these test-time methods can only recognize classes seen during training, which limits their practicality in generalizable point cloud analysis. 

Inspired by advancements in language-guided image recognition~\cite{radford21clip,jia21scaling,pham21basic,zhai22lit}, 
the third branch of works utilizes contrastive pre-training on large-scale triplets (point, text, image) to connect flexible language descriptions and enable open-vocabulary point cloud 
recognition~\cite{xue23ulip,xue24ulip2,liu23openshape,zhou24uni3d}. 
These large multimodal 3D models demonstrate promising zero-shot performance and provide a new paradigm for generalizable point cloud analysis. 
However, zero-shot predictions are performed independently for each test sample, with the final accuracy simply averaged across the test set. This approach does not fully exploit the distribution of test data, particularly the statistics of reliable predictions from large 3D models, which could further enhance accuracy and generalization.  
At this point, we pose the following question: 
\emph{can we take advantages of both the promising zero-shot ability of large 3D models and test-time adaptation techniques to unlock 
more robust and generalizable point cloud recognition?}

In this work, to fully exploit the data distribution during test time, we develop a cache model to store critical fingerprints of test samples. 
These fingerprints consist of key-value pairs derived by large multimodal 3D models, where the key represents the sample feature extracted by the point encoder, and the value represents the predicted (pseudo) label from the model.  
This cache model is entirely constructed from online test data without accessing any training samples. Our approach is thus more practical and challenging than prior methods~\cite{qin19pointdan,huang21metasets,shen22da,fan22glrv,wei22pdg,mirza23mate,wang24bftt3d,shim24cloudfixer,sun24point-prc}. 
Our cache model is hierarchical, as it stores not only the global features of a point cloud in a global cache but also its local-part details in a local cache, which are essential for distinguishing subtle differences among classes.  
We also empirically demonstrate that this coarse-to-fine design enhances robustness and generalization in the wild. 
Additionally, our cache model is dynamic, since we need to continuously update the cache with online test data to prioritize high-quality key-value pairs. 
The constructed cache serves as a powerful 3D knowledge base, capturing key characteristics of online test data. When a new sample arrives, we query the knowledge base, compute the affinity between the sample and cached keys, and produce a prediction by weighting the cached (pseudo) labels accordingly. Our method is training-free and flexible, requiring only a pre-trained 3D model and test data. More details will be provided in Sec.~\ref{sec:method}.

The superiority of the proposed method is supported by comprehensive comparisons with previous strong baselines across  
8 challenging benchmarks, containing up to 1,156 point cloud classes. 
Moreover, we implement the hierarchical cache as a plug-and-play module and integrate it into representative large 3D models, 
such as ULIP~\cite{xue23ulip}, ULIP-2~\cite{xue24ulip2}, OpenShape~\cite{liu23openshape} and Uni3D~\cite{zhou24uni3d}, 
to enhance test-time point cloud recognition dynamically. 
The proposed cache model brings notable and consistent gains over the baselines. 
For instance, our model yields +6.18\% and +4.84\% absolute accuracy improvements average over 7 types of corruptions in ModelNet-C~\cite{ren22modelnet-c} 
based on ULIP~\cite{xue23ulip} and ULIP-2~\cite{xue24ulip2}. 
On Objaverse-LVIS~\cite{Objaverse} that includes 1,156 categories, Point-Cache 
boosts the accuracy by +2.10\% absolute points compared to the zero-shot ULIP-2~\cite{xue24ulip2}. 
In summary, the contributions of our paper are threefold. 
\begin{itemize}
    \item To our knowledge, Point-Cache is the first to explore test-time point cloud recognition in a more practical and challenging
    setting, where only online test data is available and the model needs to generalize to both known classes and unseen new classes, which is beyond the capability of prior test-time methods. 
    \item We propose a hierarchical cache model to capture essential distribution clues of online test samples, creating an accurate profile for various point clouds from global to local perspectives. We implement the hierarchical cache as a plug-and-play component, enabling diverse 3D backbones to achieve enhanced test-time performance. 
    \item Comprehensive experiments and consistent gains across 8 public datasets and 4 large 3D models demonstrate the efficacy of Point-Cache. 
\end{itemize}

\section{Related Work}
\label{sec:related_work}

\begin{figure*}[t]
    \centering
    \includegraphics[width=\textwidth]{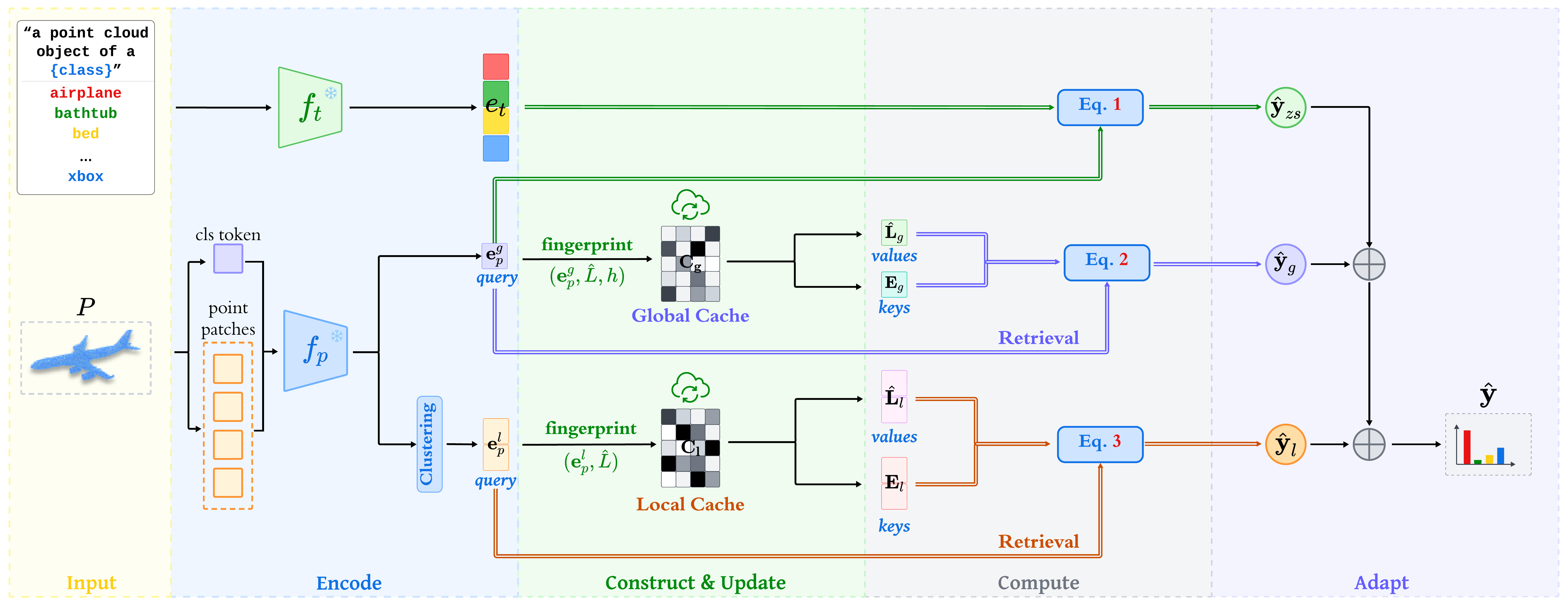}
    \caption{\textbf{The overall pipeline of Point-Cache.} The zero-shot predictions $\hat{\textbf{y}}_{zs}$ of large 3D models are 
    effectively adapted by our global cache logits $\hat{\textbf{y}}_{g}$ and local cache logits $\hat{\textbf{y}}_{l}$ to handle 
    the distribution shifts, enabling robust and generalizable point cloud analysis.}
    \label{fig:architecture}
\end{figure*}

\noindent\textbf{Test-time adaptation in 3D point cloud analysis}
receives relatively less attention compared to the 2D image community~\cite{shu22tpt,feng23difftpt,karmanov24tda,zanella24on,tsai24gda,liang20shot,sun20ttt,wang21tent,liu21ttt++,iwasawa21t3a,liang22source,zhang22memo,mirza22dua,choi22improving,gandelsman22ttt-mae}.
This technique has recently emerged in 3D point cloud tasks, including recognition~\cite{mirza23mate,wang24bftt3d,shim24cloudfixer}, registration~\cite{hatem23point-tta}, upsampling~\cite{hatem23upsampling} and object detection~\cite{wang20train,yang21st3d,zhang21srdan}.
Among these, MATE~\cite{mirza23mate}, BFTT3D~\cite{wang24bftt3d}, and CloudFixer~\cite{shim24cloudfixer} are specifically focused on test-time point cloud recognition and are thus closely related to our work. However, as previously analyzed, these methods depend heavily on training data from the source domain to construct their pipelines, which may not be available or helpful to understand 
target data distributions at test time. 
Furthermore, they lack effective strategies to leverage the distribution of test samples. By contrast, Point-Cache is constructed entirely from online test data without access to any training samples, making our setting both more challenging and more practical. Additionally, our cache model, powered by large 3D models, can generalize to both known classes and unseen new categories, an advancement that prior point cloud adaptation/generalization frameworks have not achieved~\cite{qin19pointdan,huang21metasets,shen22da,fan22glrv,wei22pdg,mirza23mate,wang24bftt3d,shim24cloudfixer}. 

\noindent\textbf{Cache models}
that store sample features in memory for further reference and reuse have been explored in fields like language modeling~\cite{vinyals16mn,grave17unbounded,merity17pointer,khandelwal20generalization}, 
image classification~\cite{vinyals16mn,snell17prototypical,finn17maml,chen18a,orhan18a,chen21meta-baseline,iwasawa21t3a,choi22improving,zhang22tip-adapter,zhang23cafo,udandarao23sus-x,karmanov24tda} and point cloud recognition~\cite{zhang23pointnn,tang24pointpeft,wang24bftt3d}. 
Our proposed cache model stands out from previous ones in several key aspects: 
(1) Our cache model is constructed based on test samples whereas previous cache models rely on a large training set for construction, as in Tip-Adapter~\cite{zhang22tip-adapter}, CaFo~\cite{zhang23cafo}, PointNN~\cite{zhang23pointnn}, Point-PEFT~\cite{tang24pointpeft} and BFTT3D~\cite{wang24bftt3d}. 
(2) Our cache model is hierarchically designed to capture both global and local-part features of a point cloud object, while existing methods store only global representations~\cite{zhang23pointnn,tang24pointpeft,wang24bftt3d,vinyals16mn,snell17prototypical,finn17maml,chen18a,chen21meta-baseline,iwasawa21t3a,choi22improving,zhang22tip-adapter,zhang23cafo}. 
We empirically demonstrate that this coarse-to-fine design improves the capture of target data features, thereby enhancing test-time performance.  
(3) Our cache model is dynamically updated with online test samples, incorporating a selective mechanism to ensure high-quality key-value pairs replace less reliable ones when the cache reaches its capacity. 
In contrast, prior cache models, which are constructed offline using the entire training set, are static and consume a substantial amount of memory~\cite{zhang23pointnn,tang24pointpeft,wang24bftt3d}. 

\noindent\textbf{Large Multi-modal 3D Models}
~\cite{zhang22pointclip,zhu23pointclip2,xue23ulip,xue24ulip2,liu23openshape,zhou24uni3d,kuo23f-vlm,yu23convolutions} spark a wave of open-vocabulary point cloud understanding by 
pre-training on large-scale datasets~\cite{Objaverse} in recent years. 
For instance, ULIP-2~\cite{xue24ulip2}, OpenShape~\cite{liu23openshape} and Uni3D~\cite{zhou24uni3d} pre-train the 3D encoders on million-scale (point, image, text) triplets using contrastive learning. These pre-trained point cloud encoders are then transferred to downstream 3D tasks like classification and retrieval, demonstrating strong zero-shot performance that surpasses traditional 3D models~\cite{qi17pointnet,qi17pointnet2,wang19dgcnn,xiang21curvenet,zhao21pt,ma22pointmlp} 
which are typically tailored to specific datasets~\cite{wu15modelnet,shapenet2015,uy19sonn}. 
This new paradigm presents a promising direction for generalizable point cloud processing. Our work is complementary to these models: rather than developing another large 3D model, we build upon their capabilities to enhance robustness and generalization for point cloud analysis at test time. Our hierarchical cache, backed by these large 3D models, generalizes to unseen classes beyond training, addressing a significant limitation of current test-time methods for point cloud recognition~\cite{mirza23mate,wang24bftt3d,shim24cloudfixer}. 
\section{Method}
\label{sec:method}

This section elaborates on the details of Point-Cache,  
the overall pipeline of our approach is depicted in Fig.~\ref{fig:architecture}.

\noindent\textbf{Preliminary.}
We begin by reviewing the zero-shot inference process of existing large multimodal 3D models~\cite{zhang22pointclip,zhu23pointclip2,xue23ulip,liu23openshape,xue24ulip2,zhou24uni3d}. 
These models use separate encoders to map inputs from a point cloud $P \in \mathbb{R}^{N\times3}$ and text $T$ (\eg, ``a point cloud object of a \verb|{class}|'') to a shared feature space. Denoting the point cloud encoder as $f_p(\cdot)$ and the text encoder as $f_t(\cdot)$, we can obtain the global feature $\textbf{e}_p^g = f_p(P) \in \mathbb{R}^d$ of the point cloud $P$ and the text feature $\textbf{e}_t = f_t(T) \in \mathbb{R}^d$, where $d$ is the feature dimension. 
To recognize the input point cloud $P$ in a zero-shot manner, these models replace ``\verb|{class}|'' in text $T$ with each of possible $C$ category names 
and compute the class probability $\hat{\textbf{y}}_{zs}$ using Eq.~\ref{eq:zero_shot_pred}.
\begin{equation}
    \hat{\textbf{y}}_{zs} = \{\hat{y}_i|_{i=1}^C\}, \quad \hat{y}_i = \frac{\exp(sim(\textbf{e}_p^g, \textbf{e}_t^i) / \tau)}{\sum_{j=1}^{C}\exp(sim(\textbf{e}_p^g, \textbf{e}_{t}^{j}) / \tau)}
    \label{eq:zero_shot_pred}
\end{equation}
Here $\textbf{e}_{t}^{i}$ indicates the text encoding of the $i$-th category. 
$sim(\cdot, \cdot)$ measures the cosine similarity between inputs, and $\tau$ is a temperature coefficient. 
Finally, the point cloud $P$ is classified into the category with the highest probability, indexed by $\hat{L} = \argmax\limits_{i} \{\hat{y}_i|_{i=1}^C\}$. 

As described in Sec.~\ref{sec:intro}, 
our hierarchical cache stores the fingerprints of online test samples, which consists of key-value pairs. 
The key is the test sample feature $\textbf{e}_p^g$ extracted by the point encoder of a pre-trained large 3D model, and the 
value is the class label $\hat{L}$ predicted by the zero-shot inference. 
In addition, we compute the entropy $h = -\sum_{i=1}^C \hat{y}_i\log\hat{y}_i$ of the zero-shot prediction $\hat{\textbf{y}}_{zs}$, 
as it is essential to assess the confidence with which the large 3D model produces this prediction. 
This entropy value is employed to filter out high-quality samples in our cache model. 

\subsection{Text \& Point Cloud Encoding}
We can obtain the text encoding $\mathbf{e}_t$ of the input $T$ and the global feature $\mathbf{e}_p^g$ of a point cloud $P$ 
as introduced in the preliminary. 
However, it is \emph{non-trivial} to represent the local-part features $\mathbf{e}_p^l$ of the point cloud $P$ due to the following challenges analyzed in Fig.~\ref{fig:challenges_of_part_feature}:

\begin{figure}[ht]
    \centering
    \includegraphics[width=0.87\columnwidth]{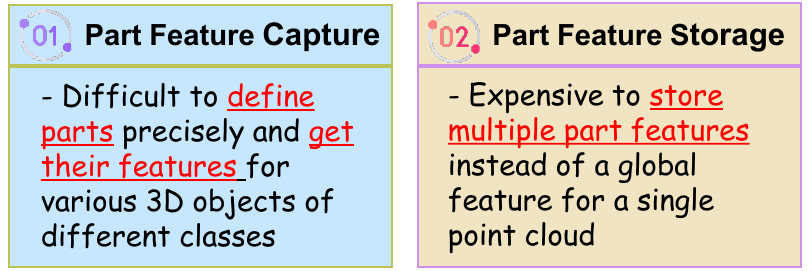}
    \caption{\textbf{Challenges in encoding part feature} for various 3D objects of different classes: Part Feature Capture \& Storage.}
    \label{fig:challenges_of_part_feature}
\end{figure}

To address these challenges, we made a careful analysis and developed effective solutions, 
seeing Fig.~\ref{fig:point_cloud_encoding} in the Supplementary for intuitive illustration.  
(1) A straightforward approach might involve using an additional pre-trained segmentation network to segment point cloud parts, but this would be computationally prohibitive during online inference. Instead, we propose a fuzzy definition of parts.  
Specifically, since the architecture of the point encoder of large 3D models is 
based on Transformer~\cite{vaswani17transformer,dosovitskiy21vit,yu22pointbert} and the input is 
a sequence of point patches, we avoid strict part definitions according to physical semantics 
(\eg, head, body, wing, and tail in an airplane), 
and instead treat the output point patches from the last encoder layer as local part representations. 
(2) A second challenge is that a long point patch sequence can become resource-intensive; for example, the point encoder in ULIP-2~\cite{xue24ulip2}
generates 512 point patches per object, which would require substantial memory. 
To mitigate this problem, we propose to summarize these point patches into $m$ centers 
$\mathbf{e}_p^l \in \mathbb{R}^{m\times d}$ using K-Means clustering ($m<10$), 
reducing memory usage by two orders of magnitude compared to storing all patches. 

\subsection{Construction \& Update of Hierarchical Cache}
\label{sec:cache_construction}

To thoroughly explore the distribution of test data, we develop a hierarchical cache to record the fingerprints of online test samples. This hierarchical cache includes both a global cache, which stores overall information, and a local cache, which captures detailed part-level information for each point cloud. The coarse-to-fine structure enables more precise profiling of test data and distinguishes our approach from previous cache models used in image recognition~\cite{vinyals16mn,snell17prototypical,finn17maml,chen18a,orhan18a,chen21meta-baseline,iwasawa21t3a,choi22improving,zhang22tip-adapter,zhang23cafo,udandarao23sus-x,karmanov24tda} 
and point cloud analysis~\cite{zhang23pointnn,tang24pointpeft,wang24bftt3d}. 
The complete construction process of hierarchical cache is outlined in Alg.~\ref{alg:hierarchical_cache_building_and_adaptation} of the Supplementary.

\noindent\textbf{Global cache} is represented  by $\textbf{C}_g = (\textbf{E}_g, \hat{\textbf{L}}_g, \textbf{h}_g)$, 
storing (feature, label, entropy) triplets for a set of point cloud objects. 
To manage storage efficiently, we design a selection mechanism to identify high-quality test samples to place into the cache. 

\emph{Construction}: 
We set an upper bound $K$ on the number of samples per class in the cache, allowing at most $K$ fingerprints (shots) can be placed in each category. 
Initially, the global cache is empty. When a new test sample $P$ arrives, we generate its fingerprint 
$(\textbf{e}_p^g, \hat{L}, h)$ and assign it to class $\hat{L}$. 

\emph{Update}: 
With the number of online samples grows, we check whether the number of shots for class $\hat{L}$ has hit the upper bound. 
If it is not the case, the fingerprint of current sample is appended to class $\hat{L}$. Otherwise, we locate the 
fingerprint with highest entropy within class $\hat{L}$ and compare it to the entropy of current test sample. 
If the current test sample has a lower entropy, we replace the highest-entropy fingerprint with the new one; 
otherwise, the current sample is skipped. 
When the cache is full, $\textbf{E}_g$ has dimensions ${CK\times d}$, 
$\hat{\textbf{L}}_g$ has dimensions ${CK\times 1}$, and $\textbf{h}_g$ has dimensions ${CK\times 1}$.

\noindent\textbf{Local cache.} 
Although the global feature of a point cloud captures expressive information, it lacks the local details of the 3D object, which are critical for point cloud recognition, 
especially when shapes share an overall appearance but differ in local parts. 
The local cache addresses this limitation by storing part-level information. 
We denote the local cache as $\textbf{C}_l = (\textbf{E}_l, \hat{\textbf{L}}_l)$,  
which store the fingerprints of local parts of point cloud objects. Here $\textbf{E}_l$ represents part features of the test samples
and $\hat{\textbf{L}}_l$ indicates their labels. 

Accordingly, we define the local-part fingerprint of a test sample $P$ as $(\textbf{e}_p^l, \hat{L})$, where 
$\textbf{e}_p^l \in \mathbb{R}^{m\times d}$ corresponds to local-part features and $\hat{L}$ is the class label of 
$P$. 

\emph{Construction \& Update}: 
Note that the sample selective mechanism of the local cache follows the same rules as the global cache, 
meaning the local-part fingerprint of a test sample is only stored if it qualifies for inclusion in the global cache. 
So there is no need to record the entropy $h$ again in the local cache.  
When the cache is full, $\textbf{E}_l$ has dimensions $(m\cdot CK)\times d$ and $\hat{\textbf{L}}_l$ has dimensions $CK\times 1$.

\subsection{Test-time Adaptation by Hierarchical Cache}
\label{sec:cache_prediction}

With the hierarchical cache serving as a rich knowledge base of online point cloud data, 
we can adapt the zero-shot predictions of large 3D models for new samples, effectively addressing distribution shifts that occur at test time. 
The adaptation follows a coarse-to-fine process as below. 

\noindent\textbf{Test-time adaptation by global cache.} 
During adaptation, a new sample $Q$ is treated as a query to retrieve knowledge from the global cache,  
which we use to produce the adaptation logits. Specifically, the retrieval process is simulated by 
calculating the affinity $A_g$ between the query and cached features $\textbf{E}_{g}$. 
This allows the new sample to establish connections with cached high-quality samples, providing it with a contextual understanding of the online test data. 
We then generate the adaptation logits $\hat{\textbf{y}}_g$ by 
weighed ensembling of cache labels $\hat{\textbf{L}}_{g}$, as shown in Eq.~\ref{eq:global_cache_pred}.
\begin{equation}
    A_g = \exp(-\beta_g(1 - \textbf{e}_q^g \textbf{E}_{g}^\top)),\ \hat{\textbf{L}}_{g} = \textrm{OH}(\hat{\textbf{L}}_{g}),\ \hat{\textbf{y}}_g =  A_{g} \hat{\textbf{L}}_{g}
    \label{eq:global_cache_pred}
\end{equation}
where $\textbf{e}_q^g \in \mathbb{R}^{1\times d}$ represents the global feature of sample $Q$, and $\textbf{E}_{g} \in \mathbb{R}^{CK\times d}$ denotes
the cached keys. The `OH' operation transforms $\hat{\textbf{L}}_{g}$ into a one-hot encoding. 
The term $\textbf{e}_q^g \textbf{E}_{g}^\top \in \mathbb{R}^{1\times CK}$ measures the cosine similarity between the query and all keys, 
with $\beta_g$ acting as a hyper-parameter to modulate the sharpness 
and the exponential operation ensuring similarity values fall within (0, 1]. 
The affinity $A_g \in \mathbb{R}^{1\times CK}$, which grasps the relationship between the query and cached keys, 
is utilized to ensemble the transformed one-hot labels $\hat{\textbf{L}}_{g} \in \mathbb{R}^{CK\times C}$, yielding  
global cache adaptation logits $\hat{\textbf{y}}_{g} \in \mathbb{R}^{1\times C}$. 

\noindent\textbf{Test-time adaptation by local cache.} 
While global cache adaptation is effective, it may not handle fine-grained distribution shifts at test time. 
To achieve a more precise adjustment, we exploit local part knowledge.
The local-part features $\textbf{e}_q^{l}$ of the point cloud $Q$ serve as queries to the 
local cache. During retrieval, we compute the affinity $A_{l}$ between $\textbf{e}_q^{l}$ and all cached part features  
$\textbf{E}_{l}$. The affinity $A_{l}$ is then used to weight the cache labels $\hat{\textbf{L}}_{l}$, producing local cache adaptation logits $\hat{\textbf{y}}_{l}$ 
as in Eq.~\ref{eq:local_cache_pred}.
\begin{equation}
    A_l = \exp(-\beta_l(1 - \textbf{e}_q^{l} \textbf{E}_{l}^\top)),\ \hat{\textbf{L}}_{l} = \textrm{OH}(\hat{\textbf{L}}_{l}),\ \hat{\textbf{y}}_{l} = \phi(A_{l} \hat{\textbf{L}}_{l})
    \label{eq:local_cache_pred}
\end{equation}
where $\textbf{e}_q^{l} \in \mathbb{R}^{m\times d}$, $\textbf{E}_{l} \in \mathbb{R}^{(m\cdot CK)\times d}$,
and $\hat{\textbf{L}}_{l} \in \mathbb{R}^{(m\cdot CK)\times C}$. The function $\phi(\cdot)$ is an average pooling operation
over $m$ parts to produce class logits. 
The term $e_q^{l} \textbf{E}_{l}^\top \in \mathbb{R}^{m\times (m\cdot CK)}$ measures the cosine similarity between 
the part features of sample $Q$ and all part features in the cache. Likewise, the similarity is 
modulated by the coefficient $\beta_l$ and scaled with the exponential function, resulting in the affinity 
$A_l \in \mathbb{R}^{m\times (m\cdot CK)}$. The term $A_{l} \hat{\textbf{L}}_{l} \in \mathbb{R}^{m\times C}$ 
represents the class logits for all $m$ parts, which are transformed by a pooling operation $\phi(\cdot)$ to derive 
the final local cache adaptation logits $\hat{\textbf{y}}_l \in \mathbb{R}^{1\times C}$.

\noindent\textbf{Overall prediction} combines three components 
as shown in Eq.~\ref{eq:overall_pred}. The zero-shot prediction $\hat{\textbf{y}}_{zs}$ of large multimodal 3D model is effectively adapted using our global cache prediction $\hat{\textbf{y}}_g$ and local cache prediction $\hat{\textbf{y}}_l$, balanced by the coefficients $\alpha_{g}$ and $\alpha_{l}$. 
\begin{equation}
    \hat{\textbf{y}} = \hat{\textbf{y}}_{zs} + \alpha_g \hat{\textbf{y}}_{g} + \alpha_l \hat{\textbf{y}}_{l}
    \label{eq:overall_pred}
\end{equation}
\section{Experiments}
\label{sec:experiments}

\begin{table*}[t]
   \footnotesize
   \centering
   \caption{\textbf{Comparison of recognition accuracy on ModelNet-C that contains 7 types of corruptions}. 
   Each clean point cloud has 1024 points and the corruption severity level is 2. The last column is the average over 7 corruption types. 
   The best results are in \textbf{bold} and the second best are \underline{underlined}. The setting applies to the following tables unless otherwise specified.}
   \label{tab:modelnet_c_robustness}
   \begin{tabular}{l c c c c c c c c c}
      \toprule
      \multirow{2}{*}{Method} & \textbf{Clean Data} & \multicolumn{7}{c}{\textbf{Corruption Type}} & \multirow{2}{*}{\textbf{Avg.}} \\\cline{3-9}
            & ModelNet & Add Global & Add Local & Drop Global & Drop Local & Rotate & Scale & Jitter & \\
      \midrule
      ULIP~\cite{xue23ulip} & 56.16 & 33.55 & 43.92 & 54.70 & 50.89 & 55.27 & 50.20 & 44.08 & 47.52 \\
      \ +\textbf{Global Cache}(Ours) & \underline{62.12} & \underline{45.79} & \textbf{47.98} & \underline{56.85} & \underline{53.89} & \underline{60.25} & \underline{54.34} & \underline{48.91} & \underline{52.56} \\ 
      \ +\textbf{Hierarchical Cache}(Ours) & \textbf{64.22} & \textbf{46.15} & \underline{47.85} & \textbf{59.16} & \textbf{56.00} & \textbf{61.47} & \textbf{55.35} & \textbf{49.92} & \textbf{53.70} \\ 
      \midrule
      ULIP-2~\cite{xue24ulip2} & 71.23 & 65.15 & 54.62 & 68.76 & 57.98 & 70.30 & 67.10 & 21.76 & 57.95 \\ 
      \ +\textbf{Global Cache}(Ours) & \underline{73.95} & \underline{67.02} & \underline{59.32} & \underline{71.35} & \underline{61.59} & \underline{72.37} & \underline{68.40} & \underline{28.20} & \underline{61.18} \\
      \ +\textbf{Hierarchical Cache}(Ours) & \textbf{74.53} & \textbf{68.11} & \textbf{61.26} & \textbf{73.22} & \textbf{63.65} & \textbf{73.34} & \textbf{70.42} & \textbf{29.50} & \textbf{62.79} \\ 
      \midrule
      O-Shape~\cite{liu23openshape} & \textbf{84.56} & 71.64 & 67.79 & 81.56 & 73.58 & 82.01 & 78.48 & 59.36 & 73.49 \\
      \ +\textbf{Global Cache}(Ours) & \underline{84.52} & \underline{74.72} & \underline{72.77} & \textbf{82.41} & \underline{75.12} & \textbf{83.18} & \textbf{78.93} & \underline{67.91} & \underline{76.43} \\
      \ +\textbf{Hierarchical Cache}(Ours) & 84.04 & \textbf{74.84} & \textbf{73.70} & \underline{82.21} & \textbf{76.26} & \underline{82.66} & \underline{78.12} & \textbf{68.35} & \textbf{76.59} \\ 
      \midrule
      Uni3D~\cite{zhou24uni3d} & 81.81 & 72.45 & 56.36 & 68.15 & 67.18 & 79.94 & 75.36 & 56.24 & 67.95 \\ %
      \ +\textbf{Global Cache}(Ours) & \underline{83.14} & \underline{76.13} & \underline{66.49} & \underline{71.43} & \underline{69.81} & \underline{81.52} & \underline{75.85} & \underline{61.43} & \underline{71.81} \\
      \ +\textbf{Hierarchical Cache}(Ours) & \textbf{83.87} & \textbf{77.51} & \textbf{71.15} & \textbf{72.16} & \textbf{70.75} & \textbf{81.77} & \textbf{77.31} & \textbf{62.52} & \textbf{73.31} \\ 
      \bottomrule
   \end{tabular}
\end{table*}

\begin{table*}[t]
   \footnotesize
   \centering
   \caption{\textbf{Comparison of recognition accuracy across a suite of datasets}. S-PB\_RS\_T50 is the hardest split of ScanObjectNN. O-LVIS: Objaverse-LVIS. Omni3D: OmniObject3D. The number under each dataset indicates the number of points (pts) for each object. In Omni3D, each point cloud can be represented by a different number of points. 
   The last column is the average accuracy on these datasets.}
   \label{tab:multi_dataset_generalization}
   \begin{tabular}{l c c c c c c c c }
      \toprule
      \multirow{2}{*}{Method} & ModelNet40 & S-PB\_RS\_T50 & O-LVIS & \multicolumn{3}{c}{Omni3D} & \multirow{2}{*}{\textbf{Avg.}} \\\cline{5-7} %
      & (10000 pts) & (2048 pts) & (10000 pts) & (1024 pts & 4096 pts & 16384 pts) \\
      \midrule
      ULIP~\cite{xue23ulip} & 58.75 & 46.44 & 6.24 & 8.39 & 7.75 & 7.28 & 22.48 \\
      \ +\textbf{Global Cache}(Ours) & \underline{61.22} & \underline{50.21} & \textbf{7.02} & \underline{10.00} & \underline{9.36} & \underline{8.43} & \underline{24.37} \\
      \ +\textbf{Hierarchical Cache}(Ours) & \textbf{62.93} & \textbf{51.80} & \textbf{7.02} & \textbf{10.47} & \textbf{9.75} & \textbf{8.90} & \textbf{25.15} \\
      \midrule
      ULIP-2~\cite{xue24ulip2} & 72.97 & 47.13 & 30.26 & 26.36 & 29.20 & 26.58 & 38.75 \\
      \ +\textbf{Global Cache}(Ours) & \underline{74.51} & \underline{51.70} & \textbf{32.65} & \underline{28.51} & \underline{31.10} & \underline{28.53} & \underline{41.17} \\
      \ +\textbf{Hierarchical Cache}(Ours) & \textbf{75.53} & \textbf{54.98} & \underline{32.36} & \textbf{29.37} & \textbf{31.24} & \textbf{29.44} & \textbf{42.15} \\
      \midrule
      Uni3D~\cite{zhou24uni3d} & 88.41 & 65.19 & \textbf{55.42} & 31.52 & 41.98 & 41.86 & 54.09 \\ %
      \ +\textbf{Global Cache}(Ours) & \underline{88.86} & \textbf{68.51} & 53.36 & \underline{34.97} & \underline{45.13} & \underline{45.19} & \underline{56.00} \\
      \ +\textbf{Hierarchical Cache}(Ours) & \textbf{89.18} & \underline{68.24} & \underline{55.19} & \textbf{35.82} & \textbf{45.60} & \textbf{45.89} & \textbf{56.65} \\
      \bottomrule
   \end{tabular}
\end{table*}

\subsection{Experimental Settings}
\label{subsec:experiment_settings}

\noindent\textbf{Datasets.} 
To evaluate the \emph{robustness} of point cloud recognition, we choose four public datasets that contain various point cloud corruptions:  
ModelNet-C~\cite{ren22modelnet-c} and three variants of ScanObjectNN-C~\cite{uy19sonn}.  
ModelNet-C~\cite{ren22modelnet-c} introduces 7 atomic corruptions, 
including adding global outliers, adding local 
outliers, dropping global structure, dropping local parts, rotation, scaling, and jittering. Other corruptions can be 
derived from these atomic corruptions, as discussed in \cite{ren22modelnet-c}. We apply these atomic corruptions to the three 
variants of ScanObjectNN~\cite{uy19sonn} and produce their corrupted versions. 
To assess \emph{generalization} on unseen new data, we test the proposed method on four challenging benchmarks: 
OmniObject3D~\cite{OmniObject3D} (216 classes), Objaverse-LVIS~\cite{Objaverse} (1,156 classes), the hardest variant of ScanObjectNN~\cite{uy19sonn} and widely used ModelNet40~\cite{wu15modelnet}. Recognition accuracy (\%) is the primary evaluation metric. 

\noindent\textbf{Implementation Details.}
We choose ULIP~\cite{xue23ulip}, ULIP-2~\cite{xue24ulip2}, OpenShape~\cite{liu23openshape} and Uni3D~\cite{zhou24uni3d} as 
the large multimodal 3D models for experiments. Their pre-trained weights are loaded and then frozen. 
Global and local-part features are generated by modifying the forward pass function without altering the model 
architecture. 
The maximum number of samples per class in the cache, $K$, is set to 3, and 
each point cloud is clustered into $m=3$ local parts. 
Both the balance factors $\alpha_g$ and $\alpha_l$ are set to 4.0 while the coefficients $\beta_g$ and $\beta_l$ are set to 3.0. 
We will examine the design choices in the ablation studies. 
To optimize memory and computation, we convert the input point clouds and model parameters into `dtype' \verb|fp16| in Pytorch.
All experiments are conducted on two 4090 GPUs. Additional details can be found in the Supplementary. 

\subsection{Test-time Robustness and Generalization} 
\label{subsec:comparison_experiments_robustness}
\noindent{\textbf{Robustness.}} The robustness analysis of our training-free test-time hierarchical caches takes place on 4 point cloud corruption datasets~\cite{ren22modelnet-c,uy19sonn}. 
As observed in Tab. \ref{tab:modelnet_c_robustness}, the zero-shot accuracy of various large 3D models 
on ModelNet-C is considerably boosted by our global cache model, with increases such as +5.04\% for ULIP~\cite{xue23ulip}, +3.23\% for ULIP-2~\cite{xue24ulip2}, +2.94\% for OpenShape~\cite{liu23openshape} and +3.86\% for Uni3D~\cite{zhou24uni3d}, averaged across 7 corruption types. 
By combining global and local cache together, our hierarchical cache model attains further improvements, highlighting the benefits of incorporating  
a local cache. 
Notably, these gains extend beyond corrupted data, as our hierarchical cache model also improves recognition accuracy on the clean ModelNet dataset, 
\eg, +8.06\% absolute increase over ULIP's zero-shot performance. 
An exception is OpenShape, which maintains a slightly higher zero-shot accuracy on clean data than our cache model. 
However, when considering the 7 corruption types, our global and hierarchical cache models outperform OpenShape by 
2.94\% and 3.10\%, respectively, demonstrating improved robustness against data corruption at test time. 
Consistent improvements are also showcased on the variants of ScanObjectNN, shown in Tab.~\ref{tab:s_obj_only_c_robustness},~\ref{tab:s_obj_bg_c_robustness} and ~\ref{tab:s_hardest_c_robustness} in the Supplementary. 

\label{subsec:comparison_experiments_generalization}

\noindent{\textbf{Generalization.}} 
We evaluated the generalization capability of Point-Cache on four representative benchmarks that were unseen during the pre-training of the large 3D models. These benchmarks generally do not contain data corruptions.  
While ModelNet40~\cite{wu15modelnet} and ScanObjectNN~\cite{uy19sonn} contain a limited number of classes, 
OmniObject3D~\cite{OmniObject3D} and Objaverse-LVIS~\cite{Objaverse} represent a broader range of 3D concepts, covering 216 and 1,156 classes, respectively.
As shown in Tab.~\ref{tab:multi_dataset_generalization}, Point-Cache
consistently enhances the recognition accuracy across datasets and 3D backbones. 
For instance, our hierarchical cache built on ULIP and ULIP-2 yields absolute improvements of 6.31\% and 5.51\% , 
respectively, average over 6 scores. 
On ScanObjectNN, our hierarchical cache significantly raises the zero-shot prediction of ULIP and ULIP-2,
from 27.20\% to 51.80\% for ULIP and from 36.22\% to 54.98\% for ULIP-2. 
On the challenging Objaverse-LVIS (O-LVIS), our global cache with ULIP-2 successfully realizes a 2.39\% absolute 
gains in zero-shot accuracy across 1,156 classes, a notable achievement. 
Furthermore, our cache model enables Uni3D to attain an 89.18\% accuracy on ModelNet40 
without any training, effectively matching the fully-supervised PointNet (89.20\%). 

\subsection{Memory Usage and Throughput}
\noindent\textbf{Memory.} 
We use Uni3D~\cite{zhou24uni3d} as the baseline and adapt its zero-shot predictions with Point-Cache. 
The experiments are conducted on 3 datasets as shown in Tab.~\ref{tab:memory_comparison_uni3d}. 
The batch size is set to 1 because we need to update the cache and adapt the class logits according to every single sample. 
The results indicate our global cache and 
hierarchical cache consume similar or slightly higher memory than the baseline on each dataset. 
Furthermore, as the number of classes increases rapidly, the memory usage grows at a slower rate. 
This is primarily because the memory consumption is dominated by the large number of parameters 
in Uni3D (\eg, 1,016.5M). In contrast, the hierarchical cache built on O-LVIS uses only approximately 7.1M parameters, 
which is negligible compared to Uni3D. Detailed calculations are provided in the Supplementary. 

\begin{table}[ht]
   \footnotesize
   \centering
   \caption{\textbf{Comparison of memory usage (MB)}. 
   The batch size is set to 1 and the used device is an RTX 4090. The number under each dataset 
   indicates \#Classes. Hierar: Hierarchical Cache.}
   \begin{tabular}{l c c c c}
      \toprule
      \multirow{2}{*}{Method} & ModelNet-C & Omni3D & O-LVIS & \#Params\\
      & (40) & (216) & (1156) & (M)\\
      \midrule
      Uni3D & \textbf{5,062} & \textbf{5,062} & \textbf{5,062} & 1,016.5 \\ 
      \ +\textbf{Global}(Ours) & \textbf{5,062} & \underline{5,064} & \underline{5,070} & 1,016.5 \\
      \ +\textbf{Hierar}(Ours) & \underline{5,064} & 5,068 & 5,090 & 1,016.5\\
      \bottomrule
   \end{tabular}
   \label{tab:memory_comparison_uni3d}
\end{table}

\noindent\textbf{Throughput.} 
We evaluate the throughput of Point-Cache on ModelNet40, where throughput is measured as the number of test samples 
processed per second (abbreviated as $t/s$). The results are reported in Tab.~\ref{tab:efficiency_comparison_mn40}. Point-Cache runs slightly slower than 
zero-shot inference due to the additional overhead introduced by cache updates, logits computation, \etc
However, the throughput degradation is minimal, \eg, a 0.04 $t/s$ drop for OpenShape with the global cache, 
and a 0.07 $t/s$ drop for ULIP-2 with the hierarchical cache. These results prove that Point-Cache brings remarkable 
accuracy improvements with negligible computational cost. 

\begin{table}[ht]
   \footnotesize
   \centering
   \caption{\textbf{Comparison of throughput ($t/s$) for different models on ModelNet40}.
   The batch size is set to 1 and the device is a 4090 GPU. 
   The results are averaged over all test samples.}
   \begin{tabular}{l r r r}
      \toprule
      Method & Zero-shot & +\textbf{Global Cache} & +\textbf{Hierar Cache} \\
      \midrule
      ULIP  & \textbf{11.25} & \underline{11.21} & 11.18 \\ %
      ULIP-2 & \textbf{11.18} & \underline{11.14} & 11.11 \\
      OpenShape & \textbf{9.80} & \underline{9.76} & 9.74 \\
      Uni3D & \textbf{9.78} & \underline{9.75} & 9.73 \\
      \bottomrule
   \end{tabular}
   \label{tab:efficiency_comparison_mn40}
\end{table}

\subsection{Ablation Study}
\label{subsec:ablation_study}

\begin{figure}[ht]
   \centering
   \begin{subfigure}{0.25\linewidth}
      \centering
      \includegraphics[width=\columnwidth]{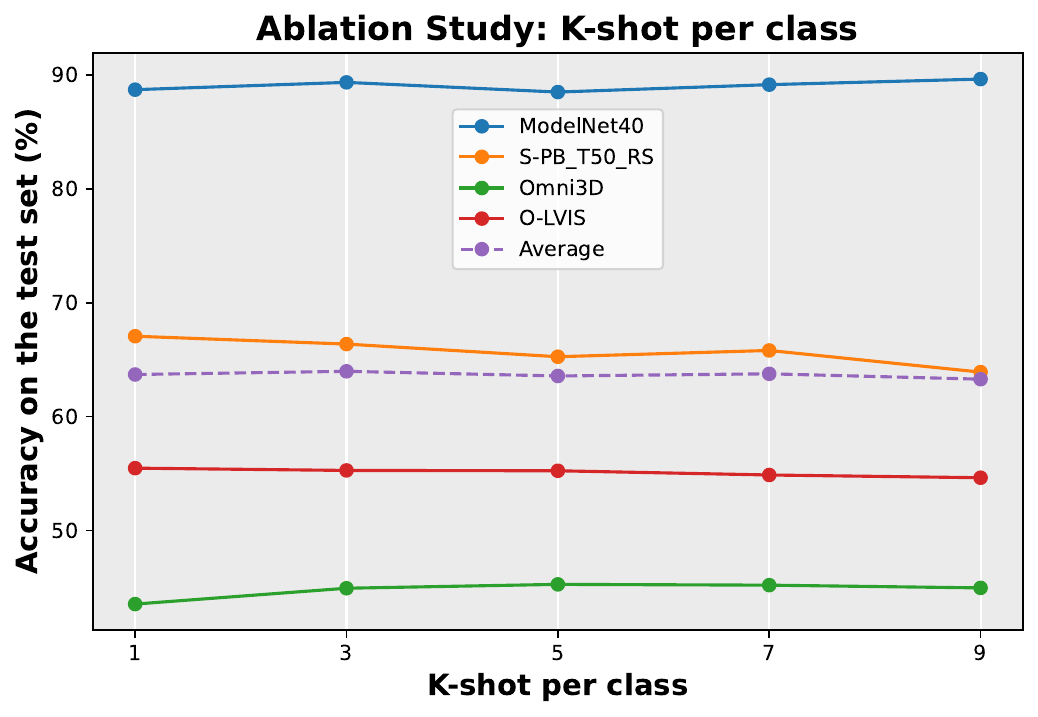}
      \caption{shot size $K$}
      \label{fig:ablate_k_shot}
   \end{subfigure}%
   \begin{subfigure}{0.25\linewidth}
      \centering
      \includegraphics[width=\columnwidth]{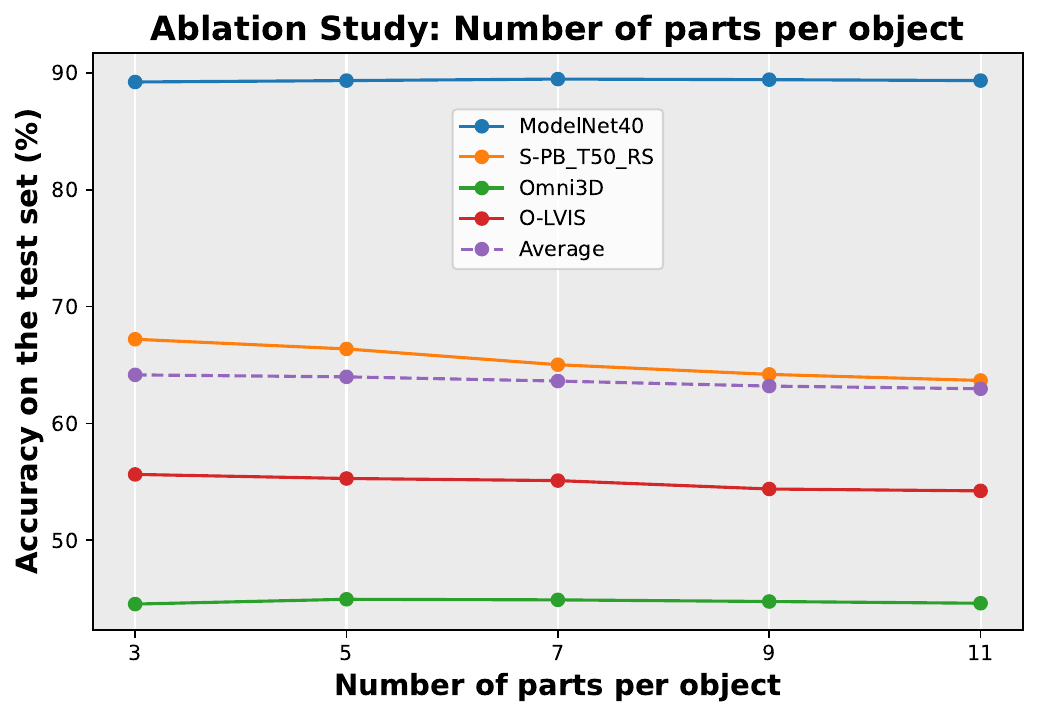}
      \caption{$m$ parts}
      \label{fig:ablate_n_cluster}
   \end{subfigure}%
   \begin{subfigure}{0.25\linewidth}
      \centering
      \includegraphics[width=\columnwidth]{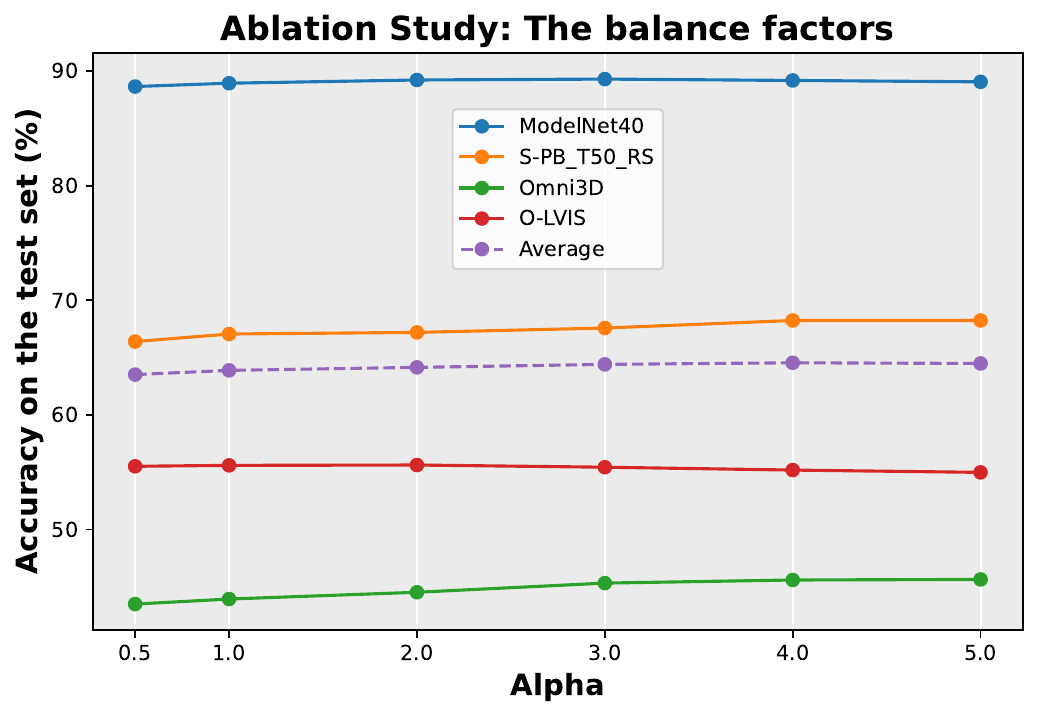}
      \caption{$\alpha_g$ and $\alpha_l$}
      \label{fig:ablate_alpha}
   \end{subfigure}%
   \begin{subfigure}{0.25\linewidth}
      \centering
      \includegraphics[width=\columnwidth]{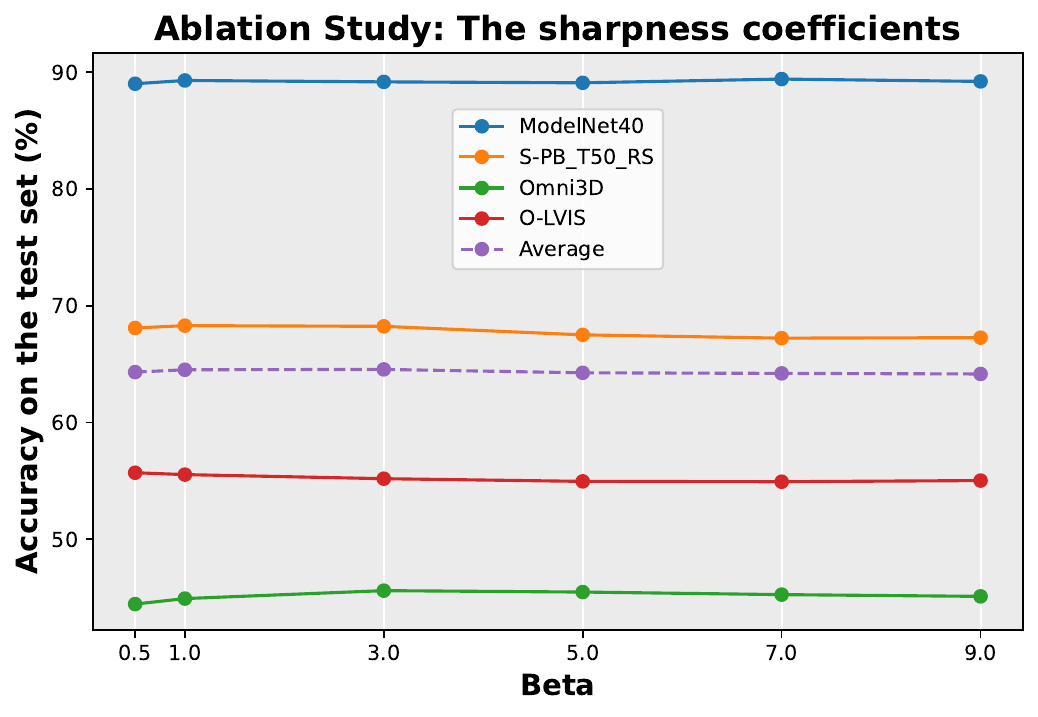}
      \caption{$\beta_g$ and $\beta_l$}
      \label{fig:ablate_beta}
   \end{subfigure}%
   \caption{\textbf{Ablation studies on the hyper-parameters in the cache design}, including the shot size $K$ per class, the number of parts $m$ per object, 
   the balance factors in the final prediction logits and the sharpness coefficients in affinity computation.}
   \label{fig:ablation_study}
\end{figure}

We use Uni3D with our hierarchical cache and 4 public datasets for ablation analysis, as shown in Fig.~\ref{fig:ablation_study}. 

\noindent\textbf{Shot size per class in the global cache.} 
The upper bound $K$ on the number of samples per class is a critical variable as it affects the total size of the hierarchical cache. 
We analyze its impact on the test-time recognition accuracy on 4 datasets, shown in Fig.~\ref{fig:ablate_k_shot}. As $K$ increases, accuracy trend differ across 
datasets; for example, accuracy increases on Omni3D but decreases on S-PB\_T50\_RS. We adopt $K=3$ as it produces 
the best average accuracy (63.98\%). The results indicate that our cache model does not require a large number 
of samples to achieve strong performance. 

\noindent\textbf{Number of local parts per 3D object.}
To optimize memory and computation, we cluster hundreds of point patches of a 3D object into $m$ parts. 
We investigate the effect of this variable on test-time performance, as exhibited in Fig.~\ref{fig:ablate_n_cluster}. 
For three out of the four datasets, namely S-PB\_T50\_RS, Omni3D, and O-LVIS, 
accuracy generally decreases as the number of parts increases, suggesting that 
a point cloud does not need to be represented with an excessive number of parts. 
We choose $m=3$ by default, as this setting reaches the best average accuracy across the 4 datasets. 

\noindent\textbf{The balance factors.}
The factors $\alpha_g$ and $\alpha_l$ control the relative weights of the global cache prediction $\hat{\textbf{p}}_g$ 
and local cache prediction $\hat{\textbf{p}}_l$. To simplify experimentation, we set $\alpha_g = \alpha_l$ in the ablation 
study. Candidate values are $\{0.5, 1.0, 2.0, 3.0, 4.0, 5.0\}$. As shown in Fig.~\ref{fig:ablate_alpha}, increasing these factors improves performance on S-PB\_T50\_RS and Omni3D, while it has minimal effect on ModelNet40 and O-LVIS. We set the factors to 4.0, which yields the optimal average accuracy across all four datasets.

\noindent\textbf{The sharpness coefficients.}
The coefficients $\beta_g$ and $\beta_l$ are used to modulate the sharpness of the query-key affinity. 
Here we analyze their impact on test-time accuracy on 4 benchmarks. Likewise, we let 
$\beta_g = \beta_l$ and the candidate values are $\{0.5, 1.0, 3.0, 5.0, 7.0, 9.0\}$. As Fig.~\ref{fig:ablate_beta} displays, accuracy initially improves slightly with increasing sharpness but 
then decreases as sharpness becomes too high.
We set the sharpness to 3.0 that yields the optimal result average over 4 datasets.

\subsection{Visualization}

\begin{figure}[ht]
   \centering
   \begin{subfigure}{0.25\linewidth}
      \centering
      \includegraphics[width=\columnwidth]{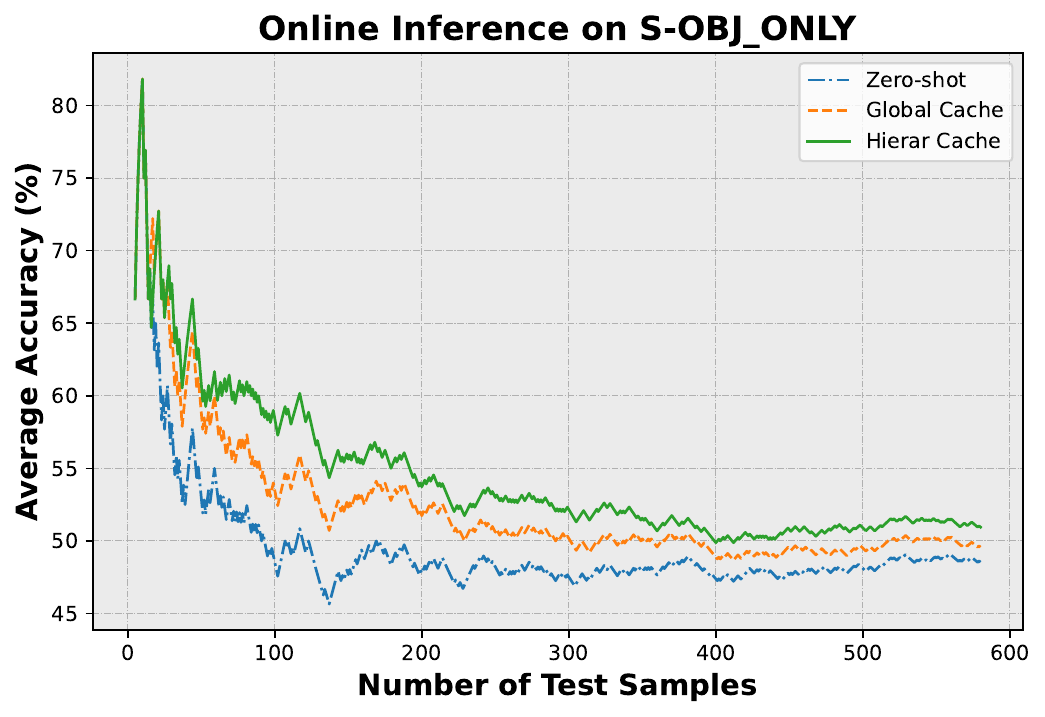}
      \caption{ULIP.}
      \label{fig:ulip1_sonn_c_obj_only_rotate_2_acc}
   \end{subfigure}%
   \begin{subfigure}{0.25\linewidth}
      \centering
      \includegraphics[width=\columnwidth]{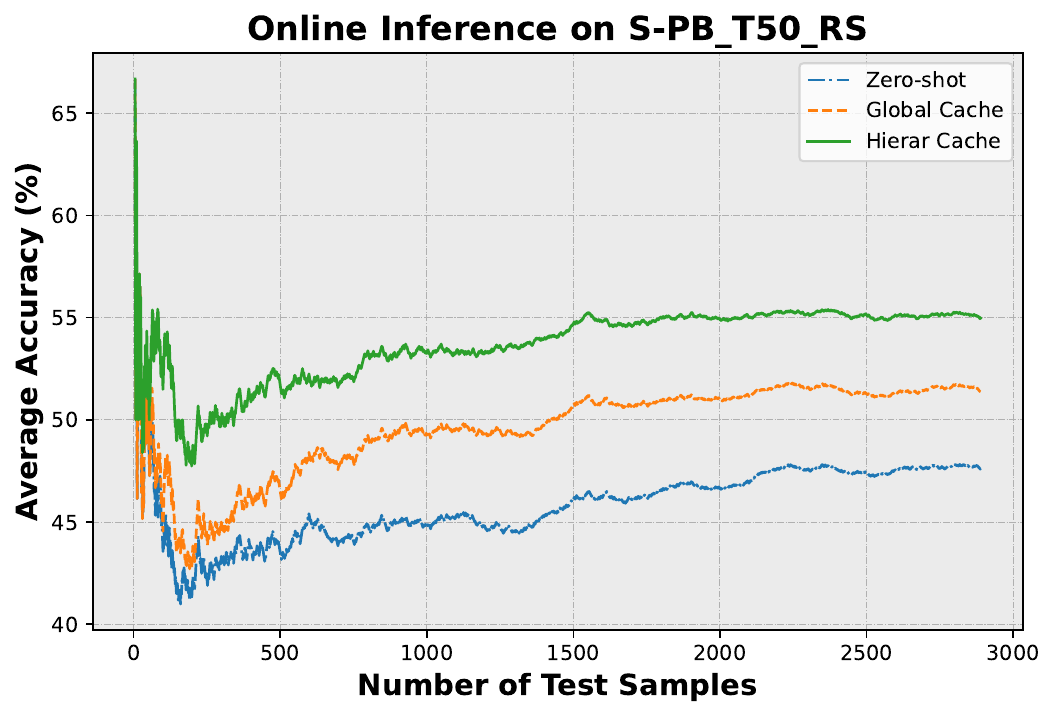}
      \caption{ULIP-2.}
      \label{fig:ulip2_scanobjnn_hardest_acc}
   \end{subfigure}%
   \begin{subfigure}{0.25\linewidth}
      \centering
      \includegraphics[width=\columnwidth]{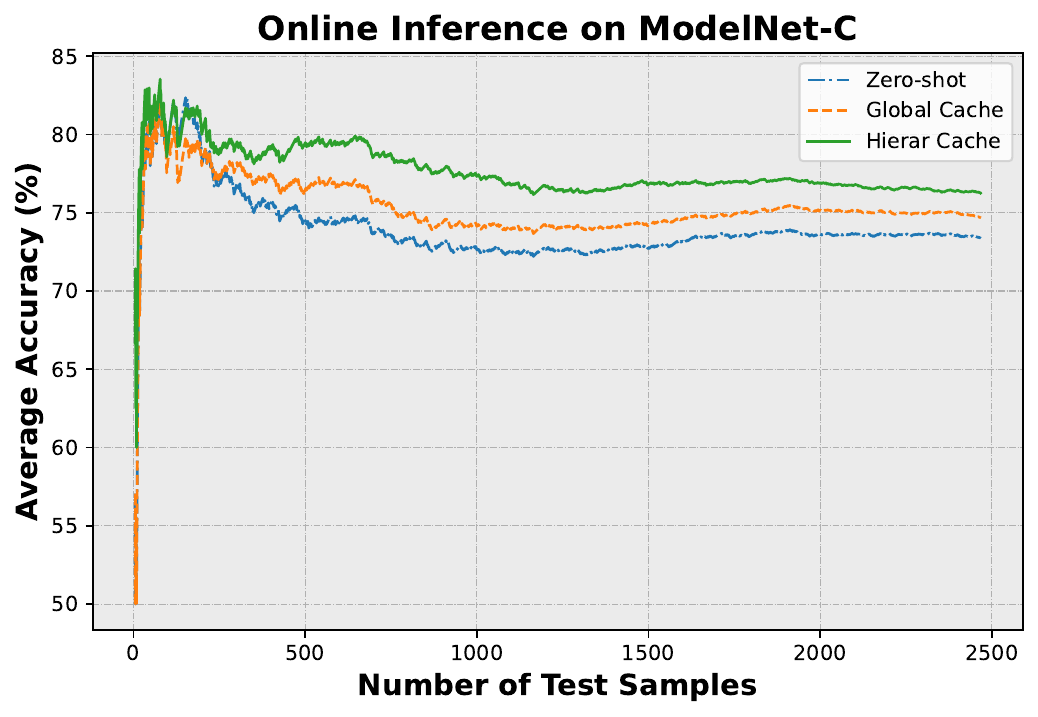}
      \caption{OpenShape.}
      \label{fig:os_mn_c_dropout_local_2_acc}
   \end{subfigure}%
   \begin{subfigure}{0.25\linewidth}
      \centering
      \includegraphics[width=\columnwidth]{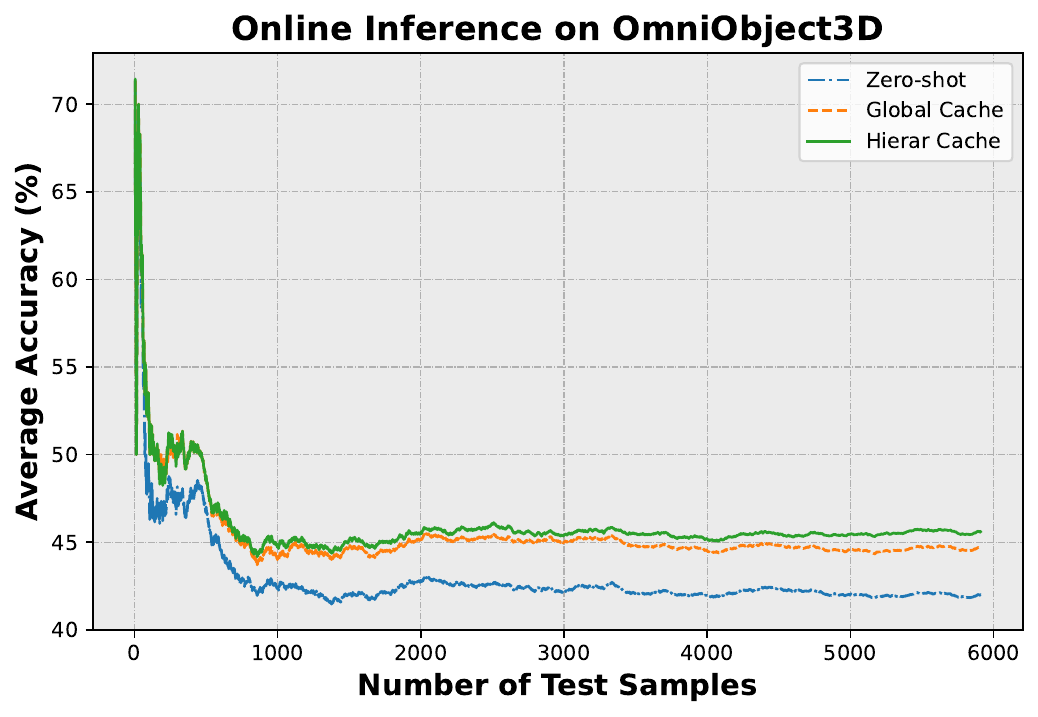}
      \caption{Uni3D.}
      \label{fig:uni3d_omniobject3d_4096pts_acc}
   \end{subfigure}%
   \caption{\textbf{The average recognition accuracy of accumulated samples during online inference}. The curve changes significantly in the 
   initial stage due to the small number of samples. Models with our global and hierarchical cache receives perceptible performance gains.}
   \label{fig:vis_online_acc}
\end{figure}

\noindent\textbf{Online Inference.}
We visualize the online inference process of Point-Cache using different large 3D models and compare three model variants: 
zero-shot prediction, the model with our global cache, and the model with our hierarchical cache, 
referring to Fig.~\ref{fig:vis_online_acc}. 
During inference, we calculate the average recognition accuracy over accumulated samples. Due to the small number of 
samples at the initial stage, the average accuracy may fluctuate drastically, so we discard the statistics for the first 5 samples. 
Overall, methods employing the hierarchical cache consistently outperform other variants by a clear margin across 
various large 3D models and datasets. 

\begin{figure}[ht]
   \centering
      \begin{subfigure}{0.25\textwidth}
      \centering
      \includegraphics[width=\columnwidth]{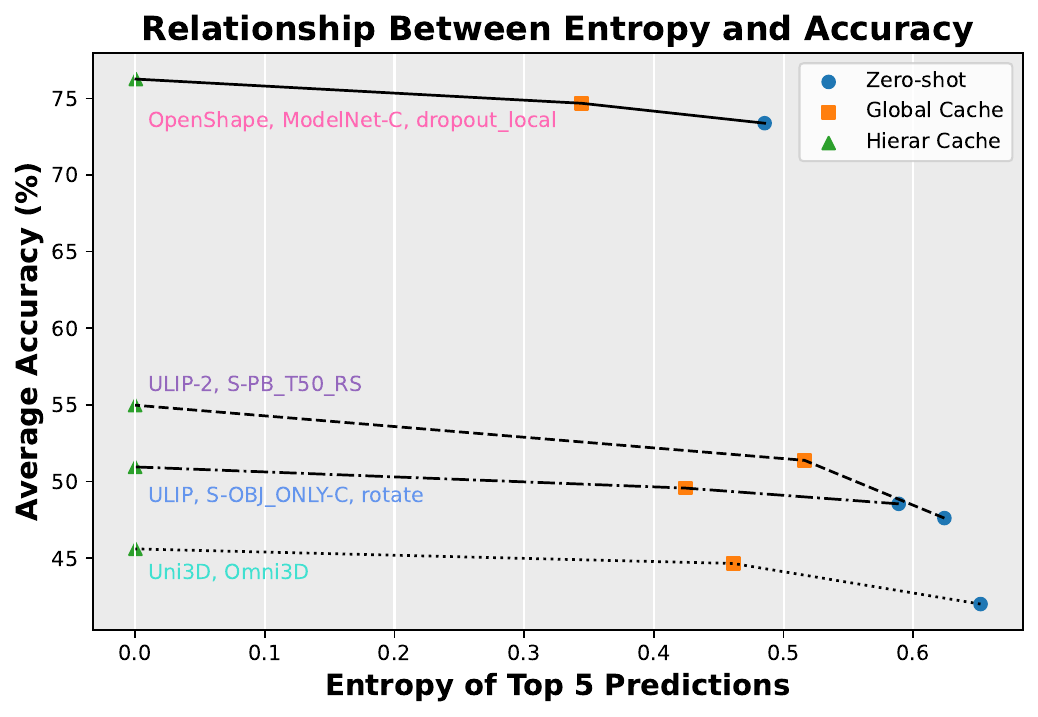}
      \caption{Group One.}
      \label{fig:ent_acc_relationship_1}
   \end{subfigure}%
   \begin{subfigure}{0.25\textwidth}
      \centering
      \includegraphics[width=\columnwidth]{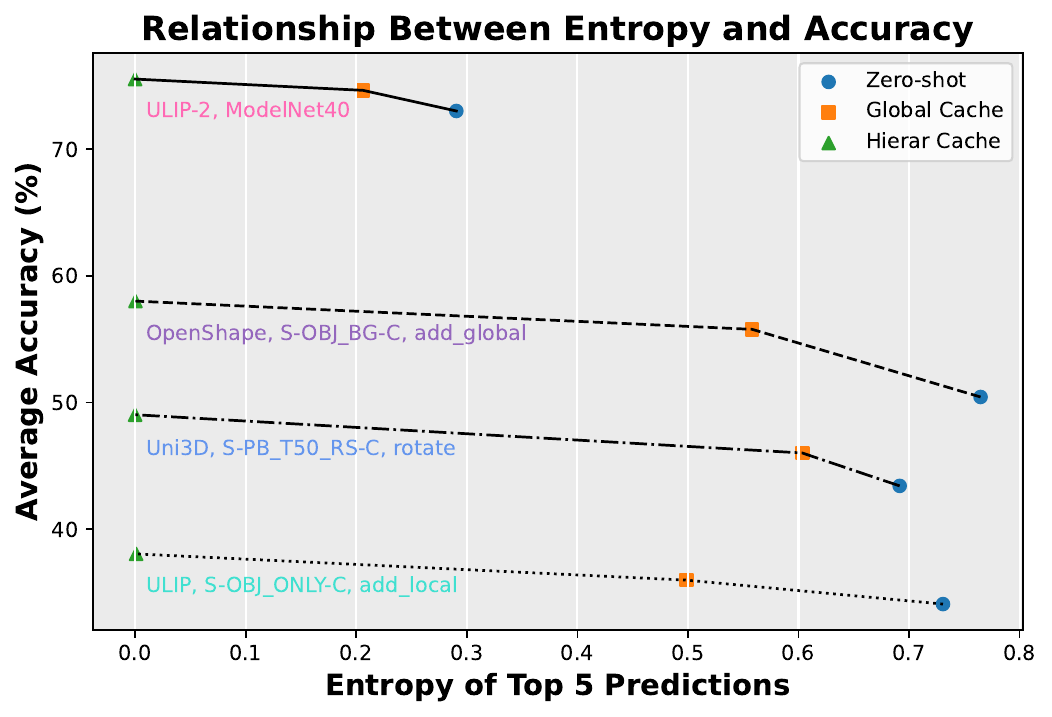}
      \caption{Group Two.}
      \label{fig:ent_acc_relationship_2}
   \end{subfigure}%
   \caption{\textbf{Relationship between entropy and accuracy.} Models with our global and hierarchical cache  
   consistently exhibit lower entropy and higher accuracy across various benchmarks.}
   \label{fig:vis_ent_acc}
\end{figure}

\noindent\textbf{Entropy and Accuracy.} 
Fig.~\ref{fig:vis_ent_acc} illustrates the relationship between entropy and accuracy for different models on various datasets. 
During online inference, we first compute the entropy of top 5 predictions for each test sample and  
then average the entropy across all accumulated samples. Similarly, the corresponding average accuracy can be calculated. 
The results reveal that Point-Cache effectively reduces the uncertainty of the zero-shot predictions made by large 3D models, 
enhances prediction confidence, and consequently lifts recognition accuracy. 
Additionally, we observe that the hierarchical cache model is more effective in reducing entropy compared to the global cache model, further underscoring the importance of incorporating the local cache.

\begin{figure}[ht]
   \centering
   \begin{subfigure}{0.125\textwidth}
      \centering
      \includegraphics[width=\columnwidth]{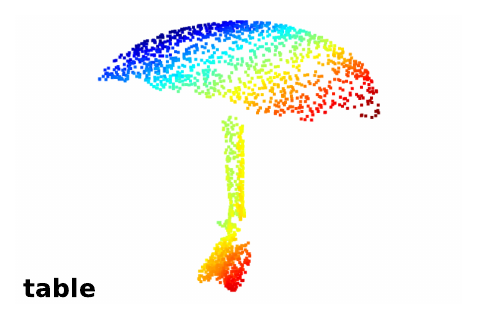}
      \caption{Ground truth.}
      \label{fig:vis_ulip2_sonn_hardest_gt}
   \end{subfigure}%
   \begin{subfigure}{0.125\textwidth}
      \centering
      \includegraphics[width=\columnwidth]{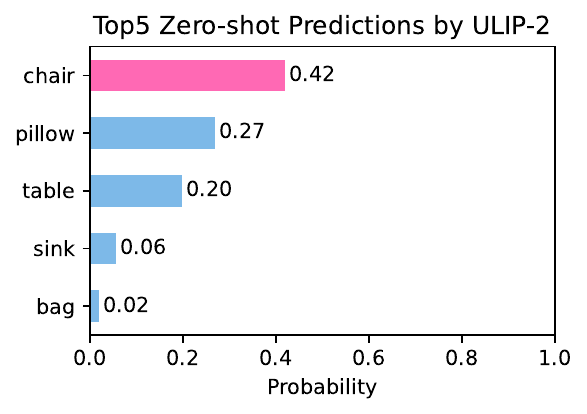}
      \caption{ULIP-2.}
      \label{fig:ulip2_sonn_hardest_zero_table}
   \end{subfigure}%
   \begin{subfigure}{0.125\textwidth}
      \centering
      \includegraphics[width=\columnwidth]{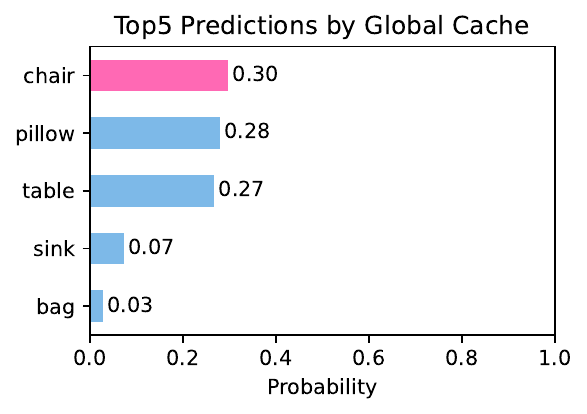}
      \caption{ULIP-2+\textbf{GC}.}
      \label{fig:ulip2_sonn_hardest_global_table}
   \end{subfigure}%
   \begin{subfigure}{0.125\textwidth}
      \centering
      \includegraphics[width=\columnwidth]{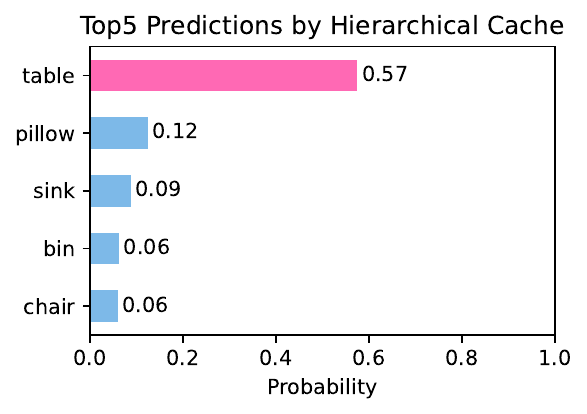}
      \caption{ULIP-2+\textbf{HC}.}
      \label{fig:ulip2_sonn_hardest_hierar_table}
   \end{subfigure}%

   \begin{subfigure}{0.125\textwidth}
      \centering
      \includegraphics[width=\columnwidth]{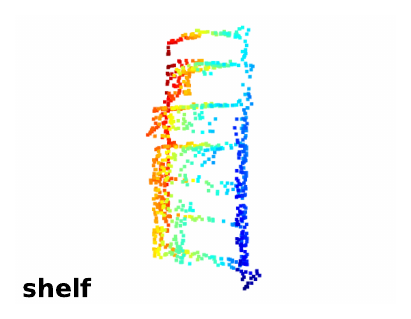}
      \caption{Ground truth.}
      \label{fig:vis_ulip1_sonn_c_obj_only_rotate_2_gt}
   \end{subfigure}%
   \begin{subfigure}{0.125\textwidth}
      \centering
      \includegraphics[width=\columnwidth]{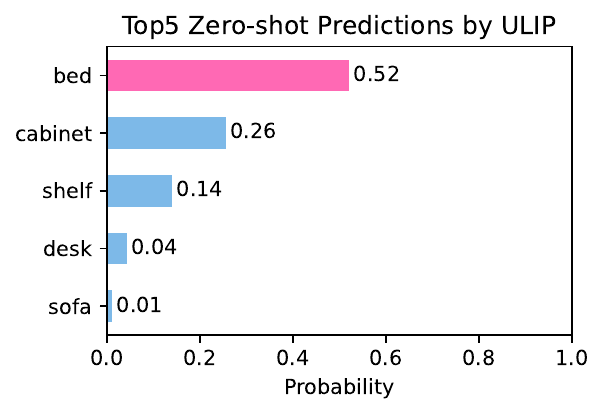}
      \caption{ULIP.}
      \label{fig:vis_ulip1_sonn_c_obj_only_rotate_2_zero_shelf}
   \end{subfigure}%
   \begin{subfigure}{0.125\textwidth}
      \centering
      \includegraphics[width=\columnwidth]{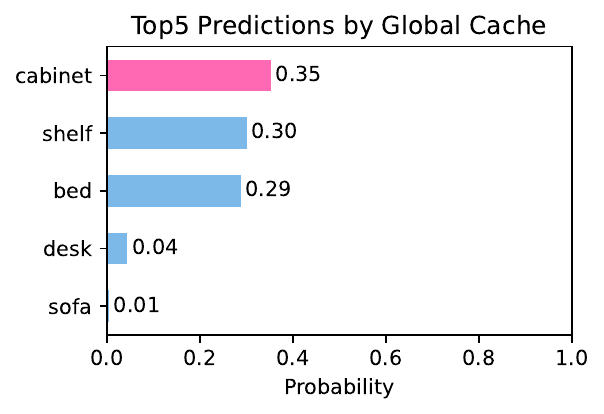}
      \caption{ULIP+\textbf{GC}.}
      \label{fig:vis_ulip1_sonn_c_obj_only_rotate_2_global_shelf}
   \end{subfigure}%
   \begin{subfigure}{0.125\textwidth}
      \centering
      \includegraphics[width=\columnwidth]{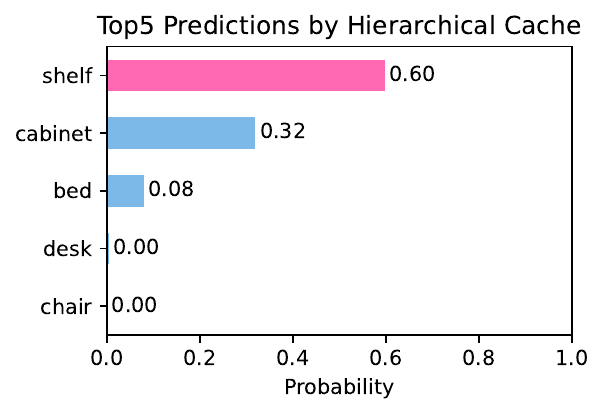}
      \caption{ULIP+\textbf{HC}.}
      \label{fig:vis_ulip1_sonn_c_obj_only_rotate_2_hierar_shelf}
   \end{subfigure}%
   
   \begin{subfigure}{0.125\textwidth}
      \centering
      \includegraphics[width=\columnwidth]{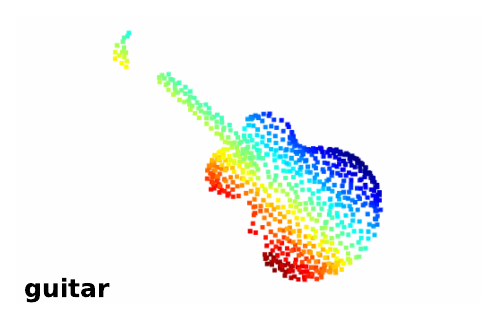}
      \caption{Ground truth.}
      \label{fig:vis_os_mn_c_dropout_local_hierar2}
   \end{subfigure}%
   \begin{subfigure}{0.125\textwidth}
      \centering
      \includegraphics[width=\columnwidth]{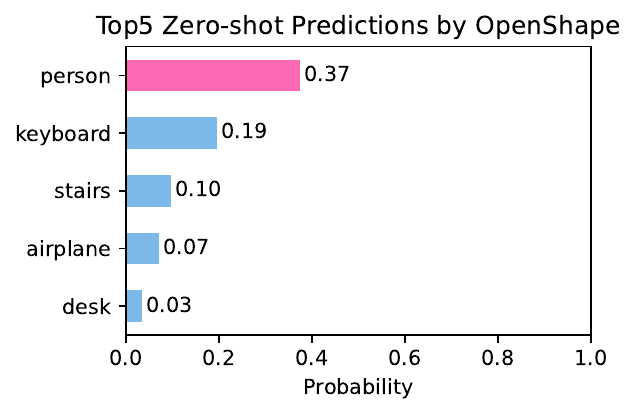}
      \caption{O-Shape.}
      \label{fig:os_mn_c_dropout_local_zero_guitar}
   \end{subfigure}%
   \begin{subfigure}{0.125\textwidth}
      \centering
      \includegraphics[width=\columnwidth]{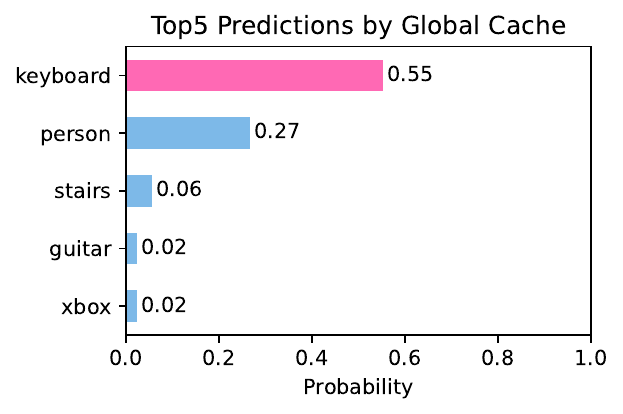}
      \caption{O-Shape+\textbf{GC}.}
      \label{fig:os_mn_c_dropout_local_global_guitar}
   \end{subfigure}%
   \begin{subfigure}{0.125\textwidth}
      \centering
      \includegraphics[width=\columnwidth]{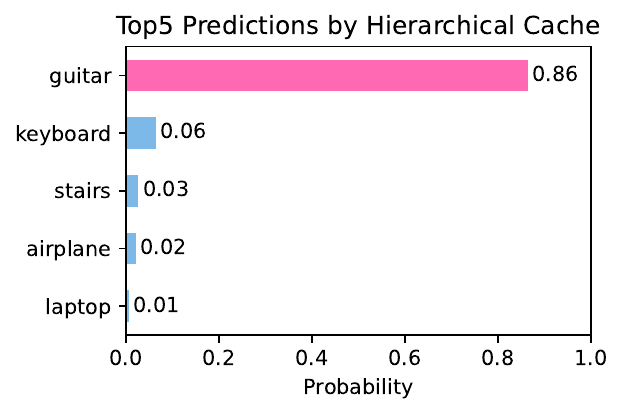}
      \caption{O-Shape+\textbf{HC}.}
      \label{fig:os_mn_c_dropout_local_hierar_guitar}
   \end{subfigure}%

   \begin{subfigure}{0.125\textwidth}
      \centering
      \includegraphics[width=\columnwidth]{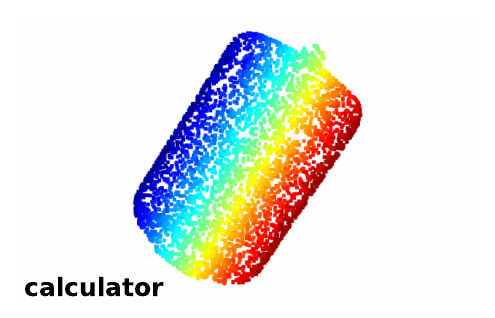}
      \caption{Ground truth.}
      \label{fig:vis_uni3d_omni3d_4096_hierar2}
   \end{subfigure}%
   \begin{subfigure}{0.125\textwidth}
      \centering
      \includegraphics[width=\columnwidth]{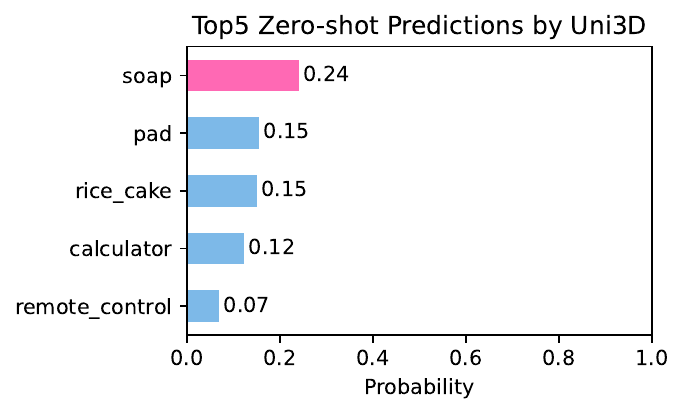}
      \caption{Uni3D.}
      \label{fig:uni3d_omni3d_4096_zero_calculator}
   \end{subfigure}%
   \begin{subfigure}{0.125\textwidth}
      \centering
      \includegraphics[width=\columnwidth]{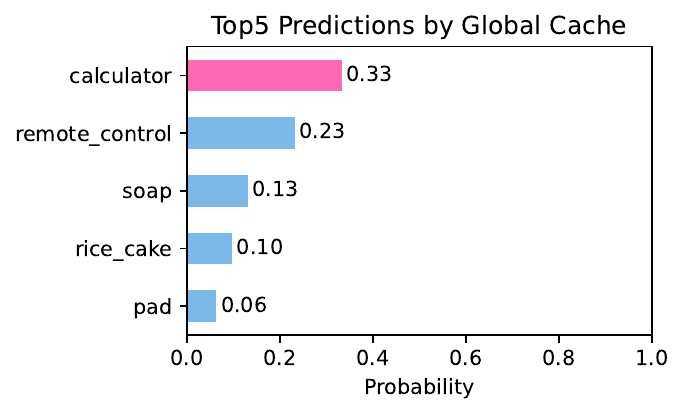}
      \caption{Uni3D+\textbf{GC}.}
      \label{fig:uni3d_omni3d_4096_global_calculator}
   \end{subfigure}%
   \begin{subfigure}{0.125\textwidth}
      \centering
      \includegraphics[width=\columnwidth]{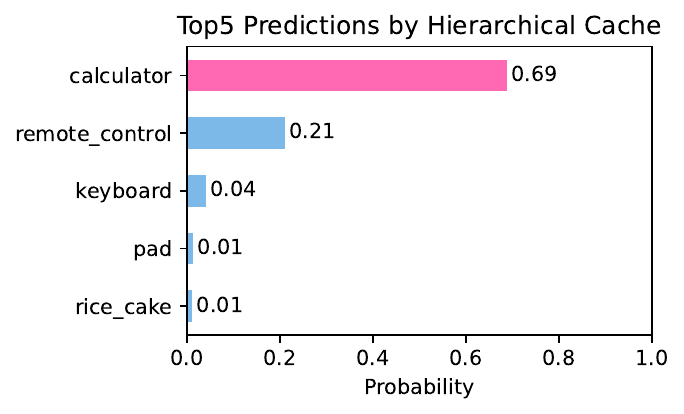}
      \caption{Uni3D+\textbf{HC}.}
      \label{fig:uni3d_omni3d_4096_hierar_calculator}
   \end{subfigure}%
   \caption{\textbf{The zero-shot predictions before and after adaptation by Point-Cache.} \textbf{GC}: global cache. \textbf{HC}: hierarchical cache.}
   \label{fig:vis_lm3d_dataset_adaptation}
\end{figure}

\noindent\textbf{Qualitative Analysis.} 
Fig.~\ref{fig:vis_lm3d_dataset_adaptation} presents the step-by-step adaptation process of Point-Cache. 
The first three rows demonstrate cases where the global cache model attempts to 
adjust the zero-shot predictions of various large 3D models but fails.
Instead, the hierarchical cache model successfully refines the class logits based on the global cache. 
In the last row, our global cache model corrects the zero-shot prediction in the first step, 
and the hierarchical cache model subsequently retains the top class while increasing its confidence.
The examples showcased here are not cherry-picked. Additional quantitative and qualitative results 
can be found in the Supplementary. 

\section{Conclusion}
\label{sec:conclusion}

This paper explores test-time point cloud recognition in a practical and challenging setting, 
where only test data and the pre-trained models are available. We develop Point-Cache,  
a powerful 3D knowledge base that enhances robustness and generalization at test time. 
It records the fingerprints of online test samples through a global and a local cache, 
creating a more accurate profile of 3D data. 
Designed as a plug-and-play module, 
Point-Cache can be seamlessly integrated into various large 3D models without compromising efficiency, 
enabling open-vocabulary point cloud recognition 
for both known and unseen classes, effectively addressing limitations of prior test-time methods.

\section*{Acknowledgement}
\label{sec:acknowledgement}

We thank all anonymous reviewers and area chairs for their time and valuable feedback. This work was partially supported by China Scholarship Council (CSC) under the Grant No.~202306360147 and Australian Research Council funding under the Grant No.~DE250100030. Dr.~Qiuhong Ke is partially supported by the Australian Research Council funding Discovery Early Career Researcher Award (DE250100030). Dr.~Deying Li was partially supported by the National Natural Science Foundation of China under the Grant No.~12071478. Dr.~Yongcai Wang was partially supported by the National Natural Science Foundation of China under the Grant No.~61972404, the Public Computing Cloud at Renmin University of China, and the Blockchain Lab, School of Information, Renmin University of China.

{
    \small
    \bibliographystyle{ieeenat_fullname}
    \bibliography{main}
}

\clearpage
\setcounter{page}{1}
\maketitlesupplementary

\section{Test-time Dynamic and Hierarchical Cache Construction and Adaptation}

The overall pipeline of Point-Cache is described in Alg.~\ref{alg:hierarchical_cache_building_and_adaptation}. This pipeline consists of five steps, corresponding to the five modules illustrated in 
Fig.~\ref{fig:architecture} of the main paper. Below, we explain several key operations in the algorithm. 

\begin{algorithm}[hbt!]
    \caption{Test-time Dynamic and Hierarchical Cache Construction and Adaptation}\label{alg:hierarchical_cache_building_and_adaptation}
    \SetEndCharOfAlgoLine{}\Comment*[l]{1. Input}
    \KwData{online test data, point cloud descriptions $T$, number of classes $C$, upper bound $K$, hyperparameters $\tau, \alpha_g, \beta_g, \alpha_l, \beta_l$}
    \KwResult{adapted class logits $\hat{\textbf{y}}$}
    \While{test sample Q}{
      \SetEndCharOfAlgoLine{}\Comment*[l]{2. Encode}
        $\textbf{e}_q^g, \textbf{e}_q^l = f_p(Q)$;

        $\textbf{e}_t = f_t(T)$;

        compute $\hat{\textbf{y}}_{zs} = \{\hat{y}_i|_{i=1}^C\}$ using Eq.~\ref{eq:zero_shot_pred};

        obtain class $\hat{L} = \argmax\limits_{i} \{\hat{y}_i|_{i=1}^C\}$ ;

        $num$ = count(\textbf{C}$_g$, $\hat{L}$) ;
        
        compute entropy $h = -\sum_{i=1}^{C}\hat{y}_{i}\log \hat{y}_{i}$;

      \SetEndCharOfAlgoLine{}\Comment*[l]{3. Update}
      \eIf{$num<K$}{
        put ($\textbf{e}_q^g, \hat{L}, h$) into global cache $\textbf{C}_g$;

        put ($\textbf{e}_q^l, \hat{L}$) into local cache $\textbf{C}_l$;

      }{$h_{max}$ = retrieve($\textbf{C}_g$, $\hat{L}$);

        ($\textbf{e}_p^{g,max}, \hat{L}, h_{max}$) = locate($\textbf{C}_g$, $\hat{L}$, $h_{max}$);

        ($\textbf{e}_p^{l,max}, \hat{L}$) = locate($\textbf{C}_l$, $\hat{L}$, $h_{max}$);

        \If{$h<h_{max}$}{
          ($\textbf{e}_p^{g,max}, \hat{L}, h_{max}$) $\leftarrow$ ($\textbf{e}_q^g, \hat{L}, h$);

          ($\textbf{e}_p^{l,max}, \hat{L}$) $\leftarrow$ ($\textbf{e}_q^l, \hat{L}$);

        }
      }
      \SetEndCharOfAlgoLine{}\Comment*[l]{4. Compute}
      compute $\hat{\textbf{y}}_g$ using Eq.~\ref{eq:global_cache_pred};

      compute $\hat{\textbf{y}}_l$ using Eq.~\ref{eq:local_cache_pred};

      \SetEndCharOfAlgoLine{}\Comment*[l]{5. Adapt}
      adapt $\hat{\textbf{y}}_{zs}$ and obtain new logits $\hat{\textbf{y}}$ using Eq.~\ref{eq:overall_pred};

      return $\hat{\textbf{y}}$;
    }
\end{algorithm}

\begin{itemize}
    \item The function `count($\textbf{C}_g, \hat{L}$)' calculates the number of cached fingerprints belonging to class $\hat{L}$ in the global cache $\textbf{C}_g$. 
    \item The function `retrieve($\textbf{C}_g, \hat{L}$)' returns the maximum entropy among the cached fingerprints of class $\hat{L}$ in $\textbf{C}_g$. 
    \item The function `locate($\textbf{C}_g$, $\hat{L}$, $h_{max}$)' identifies the fingerprint with the highest entropy for class $\hat{L}$ in $\textbf{C}_g$. Similarly, `locate($\textbf{C}_l$, $\hat{L}$, $h_{max}$)' performs the same operation in the local cache $\mathbf{C}_l$. 
    \item The operator $\leftarrow$ indicates that the fingerprint with the highest entropy is replaced by the fingerprint of the current sample $Q$ if $h<h_{max}$.  
\end{itemize}

\section{Implementation Details}

For the point encoder in ULIP~\cite{xue23ulip} and ULIP-2~\cite{xue24ulip2}, we use PointBert~\cite{yu22pointbert} as the backbone. 
For the point encoder in OpenShape~\cite{liu23openshape}, we utilize the scaled version of PointBert (32.1M parameters), as detailed in Table 4 of the Appendix in the corresponding paper. For Uni3D, we employ the giant version, where the point encoder has 1,016.5M parameters. The pre-trained weights for these models are obtained from their public 
GitHub repositories.
The zero-shot recognition accuracy (\%) of the various large 3D models are the baselines for comparison. %

Rather than relying on a single fixed template (\eg, `a point cloud object of a \{\verb|class|\}') to describe a point cloud,  
we adopt 64 text templates to generate diverse descriptions of 3D objects, as in ULIP~\cite{xue23ulip} and Point-PRC~\cite{sun24point-prc}. These 
descriptions are encoded into 64 text embeddings, which are then averaged to create a feature representation for a specific class.

\section{Additional Results and Analysis}

\subsection{Test-time Robustness and Generalization}

\textbf{Robustness against data corruptions.} 
We also create the corrupted versions for the three splits of ScanObjectNN according to the atomic operations in ModelNet-C~\cite{ren22modelnet-c} and 
conduct experiments on them. The results are reported in Tab.~\ref{tab:s_obj_only_c_robustness},~\ref{tab:s_obj_bg_c_robustness} and \ref{tab:s_hardest_c_robustness}. 
The proposed global and hierarchical cache models bring consistent and significant improvements across backbones, datasets and corruption types. 
For instance, +6.61\% for ULIP on S-OBJ\_ONLY-C, +6.05\% for Uni3D on S-OBJ\_BG-C, and +5.72\% for OpenShape on S-PB\_T50-\_RS-C 
across 7 corruptions, compared to the corresponding zero-shot predictions. The results verify the effectiveness of Point-Cache 
in strengthening the robustness of large 3D model against data corruptions. 
Likewise as the observations from Tab.~\ref{tab:modelnet_c_robustness}, the gains are not limited to 
corrupted data. Point-Cache also boosts the recognition accuracy of various models on original clean data. 

\begin{table*}[ht]
    \footnotesize
    \centering
    \caption{\textbf{Comparison of recognition accuracy on S-OBJ\_ONLY-C that includes 7 types of corruptions}. 
    Results are reported for a corruption severity level of 2. Each clean point cloud contains 1024 points. The last column is the average across the 7 types of corruptions. 
    SONN: ScanObjectNN.}
    \label{tab:s_obj_only_c_robustness}
    \begin{tabular}{l c c c c c c c c c}
       \toprule
       \multirow{2}{*}{Method} & \textbf{Original Data} & \multicolumn{7}{c}{\textbf{Corruption Type}} & \multirow{2}{*}{\textbf{Avg.}} \\\cline{3-9}
             & SONN & Add Global & Add Local & Drop Global & Drop Local & Rotate & Scale & Jitter & \\
       \midrule
       ULIP~\cite{xue23ulip} & 49.05 & 31.50 & 34.77 & 51.29 & 38.38 & 48.36 & 44.58 & 36.83 & 40.82 \\
       \ +\textbf{Global Cache}(Ours) & \textbf{52.15} & \textbf{35.80} & \underline{37.01} & \underline{54.39} & \underline{41.82} & \underline{49.74} & \underline{45.09} & \textbf{40.28} & \underline{43.45} \\ 
       \ +\textbf{Hierarchical Cache}(Ours) & \textbf{52.15} & \underline{32.01} & \textbf{38.04} & \textbf{54.56} & \textbf{45.27} & \textbf{50.95} & \textbf{45.96} & \underline{39.24} & \textbf{43.72} \\ 
       \midrule
       ULIP-2~\cite{xue24ulip2} & 42.00 & 40.45 & 41.31 & 37.69 & 30.29 & 38.21 & 44.45 & 22.89 & 36.47 \\ 
       \ +\textbf{Global Cache}(Ours) & \underline{48.19} & \textbf{49.05} & \textbf{46.30} & \underline{45.09} & \underline{37.18} & \underline{41.65} & \underline{44.41} & \textbf{25.99} & \underline{41.38} \\
       \ +\textbf{Hierarchical Cache}(Ours) & \textbf{51.98} & \textbf{49.05} & \textbf{46.30} & \textbf{48.88} & \textbf{40.45} & \textbf{45.78} & \textbf{45.09} & \textbf{25.99} & \textbf{43.08} \\ 
       \midrule
       O-Shape~\cite{liu23openshape} & 53.18 & 49.91 & 46.30 & 52.15 & 36.66 & 46.64 & 46.82 & 30.81 & 44.18 \\
       \ +\textbf{Global Cache}(Ours) & \underline{56.80} & \underline{56.45} & \underline{51.98} & \underline{54.56} & \underline{40.45} & \textbf{51.81} & \textbf{49.23} & \underline{37.69} & \underline{48.88} \\ 
       \ +\textbf{Hierarchical Cache}(Ours) & \textbf{58.69} & \textbf{59.04} & \textbf{53.01} & \textbf{55.94} & \textbf{41.82} & \underline{51.12} & \underline{48.54} & \textbf{39.41} & \textbf{49.84} \\ 
       \midrule
       Uni3D~\cite{zhou24uni3d} & 65.58 & 62.65 & 56.45 & 60.07 & 49.40 & 61.62 & 56.11 & 43.55 & 55.69 \\ %
       \ +\textbf{Global Cache}(Ours) & \underline{70.05} & \underline{65.06} & \textbf{59.38} & \underline{63.68} & \underline{54.39} & \textbf{63.34} & \underline{60.07} & \underline{51.29} & \underline{59.60} \\
       \ +\textbf{Hierarchical Cache}(Ours) & \textbf{70.22} & \textbf{65.40} & \underline{58.00} & \textbf{64.20} & \textbf{54.91} & \underline{61.96} & \textbf{62.13} & \textbf{53.18} & \textbf{59.97} \\ 
       \bottomrule
    \end{tabular}
\end{table*}

\begin{table*}[ht]
    \footnotesize
    \centering
    \caption{\textbf{Comparison of recognition accuracy on S-OBJ\_BG-C that includes 7 types of corruptions.} 
    The results are reported for a corruption severity level of 2. Each clean point cloud has 1024 points. The last column is the average across the 7 types of corruptions.}
    \label{tab:s_obj_bg_c_robustness}
    \begin{tabular}{l c c c c c c c c c}
        \toprule
        \multirow{2}{*}{Method} & \textbf{Original Data} & \multicolumn{7}{c}{\textbf{Corruption Type}} & \multirow{2}{*}{\textbf{Avg.}} \\\cline{3-9}
                & SONN & Add Global & Add Local & Drop Global & Drop Local & Rotate & Scale & Jitter & \\
        \midrule
        ULIP~\cite{xue23ulip} & 45.96 & 27.19 & 25.82 & 45.61 & 34.25 & 40.96 & 40.10 & 30.98 & 34.99 \\
        \ +\textbf{Global Cache}(Ours) & \underline{48.88} & \textbf{30.46} & \textbf{30.46} & \textbf{49.05} & \underline{39.59} & \textbf{44.92} & \textbf{42.17} & \underline{31.84} & \textbf{38.36} \\ 
        \ +\textbf{Hierarchical Cache}(Ours) & \textbf{49.74} & \underline{28.23} & \underline{30.12} & \underline{48.71} & \textbf{40.45} & \underline{43.55} & \underline{40.28} & \textbf{34.42} & \underline{37.97} \\ 
        \midrule
        ULIP-2~\cite{xue24ulip2} & 48.19 & 40.62 & 38.90 & 39.24 & 32.36 & 41.14 & 42.86 & 21.17 & 36.61 \\ 
        \ +\textbf{Global Cache}(Ours) & \underline{52.50} & \underline{48.19} & \underline{45.09} & \underline{46.82} & \underline{39.07} & \underline{46.64} & \underline{48.02} & \textbf{26.51} & \underline{42.91} \\
        \ +\textbf{Hierarchical Cache}(Ours) & \textbf{54.73} & \textbf{51.64} & \textbf{47.16} & \textbf{50.95} & \textbf{39.76} & \textbf{53.01} & \textbf{51.81} & \underline{22.72} & \textbf{45.29} \\ 
        \midrule
        O-Shape~\cite{liu23openshape} & 55.94 & 49.40 & 48.19 & 52.67 & 42.51 & 48.88 & 47.16 & 31.84 & 45.81 \\
        \ +\textbf{Global Cache}(Ours) & \underline{59.72} & \underline{57.49} & \underline{51.12} & \textbf{59.72} & \textbf{48.71} & \textbf{56.11} & \textbf{54.22} & \underline{35.28} & \textbf{51.81} \\ 
        \ +\textbf{Hierarchical Cache}(Ours) & \textbf{62.65} & \textbf{58.00} & \textbf{51.64} & \underline{59.55} & \underline{47.85} & \underline{54.91} & \underline{53.36} & \textbf{36.49} & \underline{51.69} \\ 
        \midrule
        Uni3D~\cite{zhou24uni3d} & 60.24 & 58.00 & 52.32 & 51.64 & 44.23 & 58.00 & 51.81 & 39.24 & 50.75 \\ %
        \ +\textbf{Global Cache}(Ours) & \textbf{63.86} & \textbf{66.27} & \textbf{57.83} & \underline{56.11} & \textbf{50.77} & \textbf{61.62} & \underline{56.11} & \underline{44.23} & \underline{56.13} \\
        \ +\textbf{Hierarchical Cache}(Ours) & \underline{62.82} & \underline{64.72} & \underline{57.14} & \textbf{58.52} & \underline{50.43} & \underline{60.93} & \textbf{59.55} & \textbf{46.30} & \textbf{56.80} \\ 
        \bottomrule
    \end{tabular}
\end{table*}

\begin{table*}[ht]
  \footnotesize
  \centering
  \caption{\textbf{Comparison of corruption generalization on S-PB\_T50-\_RS-C}, which is the hardest split of ScanObjectNN is used. Each clean point cloud is represented by 1024 points. SONN is short for ScanObjectNN.}
  \label{tab:s_hardest_c_robustness}
  \begin{tabular}{l c c c c c c c c c}
     \toprule
     \multirow{2}{*}{Method} & \textbf{Original Data} & \multicolumn{7}{c}{\textbf{Corruption Type}} & \multirow{2}{*}{\textbf{Avg.}} \\\cline{3-9}
           & SONN & Add Global & Add Local & Drop Global & Drop Local & Rotate & Scale & Jitter & \\
     \midrule
     ULIP~\cite{xue23ulip} & 29.29 & 19.26 & 18.39 & 30.99 & 23.91 & 27.48 & 26.34 & 21.44 & 23.97 \\
     \ +\textbf{Global Cache}(Ours) & \underline{32.37} & \underline{22.87} & \underline{20.85} & \underline{33.31} & \underline{27.90} & \underline{30.85} & \textbf{28.63} & \underline{24.53} & \underline{26.99} \\ 
     \ +\textbf{Hierarchical Cache}(Ours) & \textbf{32.48} & \textbf{23.46} & \textbf{22.69} & \textbf{34.70} & \textbf{31.75} & \textbf{33.00} & \underline{28.28} & \textbf{25.05} & \textbf{28.42} \\ 
     \midrule
     ULIP-2~\cite{xue24ulip2} & 33.38 & 30.29 & 29.42 & 28.24 & 24.91 & 28.56 & 30.22 & 12.98 & 26.37 \\ 
     \ +\textbf{Global Cache}(Ours) & \underline{40.28} & \underline{36.40} & \underline{33.80} & \underline{35.39} & \underline{30.88} & \underline{33.66} & \underline{35.01} & \underline{18.36} & \underline{31.93} \\
     \ +\textbf{Hierarchical Cache}(Ours) & \textbf{42.40} & \textbf{35.70} & \textbf{34.42} & \textbf{37.75} & \textbf{34.21} & \textbf{36.26} & \textbf{36.09} & \textbf{19.12} & \textbf{33.36} \\ 
     \midrule
     O-Shape~\cite{liu23openshape} & 41.12 & 32.41 & 35.60 & 37.80 & 27.34 & 36.61 & 35.22 & 18.88 & 31.98 \\
     \ +\textbf{Global Cache}(Ours) & \underline{42.16} & \underline{40.32} & \underline{37.58} & \underline{42.02} & \underline{33.76} & \underline{41.53} & \textbf{38.24} & \underline{24.12} & \underline{36.80} \\ 
     \ +\textbf{Hierarchical Cache}(Ours) & \textbf{43.72} & \textbf{40.91} & \textbf{39.24} & \textbf{43.03} & \textbf{35.22} & \textbf{43.06} & \underline{37.40} & \textbf{25.05} & \textbf{37.70} \\ 
     \midrule
     Uni3D~\cite{zhou24uni3d} & 46.04 & 48.23 & 37.99 & 36.75 & 31.47 & 44.00 & 37.37 & 28.66 & 37.38 \\
     \ +\textbf{Global Cache}(Ours) & \underline{50.28} & \textbf{52.57} & \textbf{42.23} & \underline{42.61} & \underline{36.29} & \underline{47.22} & \underline{39.83} & \underline{33.48} & \underline{42.03} \\
     \ +\textbf{Hierarchical Cache}(Ours) & \textbf{51.13} & \underline{51.67} & \underline{41.88} & \textbf{44.59} & \textbf{38.79} & \textbf{49.03} & \textbf{41.05} & \textbf{34.70} & \textbf{43.10} \\ 
     \bottomrule
  \end{tabular}
\end{table*}

\noindent\textbf{Generalization from simulated to real data.}
We investigate the performances of Point-Cache on Sim-to-Real~\cite{huang21metasets}, which is used to evaluate 
the generalization from simulated data (in the source domain) to real data (in the target domain). 
Sim-to-Real introduces two evaluation settings: MN\_11 $\rightarrow$ SONN\_11 and SN\_9 $\rightarrow$ SONN\_9. MN is short for 
ModelNet, SN is short for ShapeNet and SONN is short for ScanObjectNN. SONN has three splits, 
as shown in Tab.~\ref{tab:xset_sim2real_generalization}. 
The results suggest our global cache model substantially raise the zero-shot accuracy of various large 3D models, \eg, 
+3.77\% based on Uni3D. 
Also, the hierarchical cache model leads the global one by a clear margin, \eg, 
+3.46\% based on ULIP-2 across 6 datasets, 
revealing the effectiveness of local cache again. 
Note that we also compare with prior strong baselines that are trained on the source domain, such as MetaSets~\cite{huang21metasets}, PDG~\cite{wei22pdg} and I-OODG~\cite{zhang24invariantoodg}. 
In contrast, Point-Cache is directly transferred to the target datasets of Sim-to-Real and totally training-free. 
As a result, we attain competitive or even better performances compared to those learning-based baselines. 

\begin{table*}[ht]
    \centering\footnotesize
    \caption{\textbf{Comparison of recognition accuracy on Sim-to-Real}. Two evaluation settings are considered:  
    MN\_11 $\rightarrow$ SONN\_11 and SN\_9 $\rightarrow$ SONN\_9. 
    The dataset on the left side of $\rightarrow$ stands for simulated data, while the dataset on the right side indicates real-world data.  
    11 classes are shared between MN\_11 and SONN\_11, while 9 classes are common between SN\_9 and SONN\_9. 
    The last column shows the average accuracy across 6 datasets. In the experiments, each point cloud is represented by 2,048 points. 
    MN: ModelNet, SN: ShapeNet, -P: PointNet, -D: DGCNN. Note that our methods are \textbf{training-free} while prior methods (\eg, PDG, MetaSets) \textbf{use the full training set} to build their models.}
    \begin{tabular}{l c c c c c c c c c}
       \toprule
       \multirow{2}{*}{Method} & \multirow{2}{*}{Training?} & \multicolumn{3}{c}{MN\_11 $\rightarrow$ SONN\_11} & & \multicolumn{3}{c}{SN\_9 $\rightarrow$ SONN\_9} & \multirow{2}{*}{\textbf{Avg.}} \\\cline{3-5}\cline{7-9}
              & & OBJ & OBJ\_BG & PB\_T50\_RS & & OBJ & OBJ\_BG & PB\_T50\_RS\\
       \midrule
       MetaSets-P~\cite{huang21metasets} & \cmark & 60.3 & 52.4 & 47.4 & & 51.8 & 44.3 & 45.6 & 50.3 \\ %
       MetaSets-D~\cite{huang21metasets} & \cmark & 58.4 & 59.3 & 48.3 & & 49.8 & 47.4 & 42.7 & 51.0\\ %
       PDG-P~\cite{wei22pdg} & \cmark & 67.6 & 58.5 & 56.6 & & 57.3 & 51.3 & 51.3 & 57.1 \\ %
       PDG-D~\cite{wei22pdg} & \cmark & 65.3 & 65.4 & 55.2 & & 59.1 & 59.3 & 51.0 & 59.2 \\ %
       I-OODG~\cite{zhang24invariantoodg} & \cmark & - & 69.8 & - & & - & 59.8 & - & 64.8 \\ %
       \midrule
       ULIP~\cite{xue23ulip} & \xmark & 57.05 & 50.32 & 32.60 & & 61.00 & 61.00 & 44.38 & 51.06 \\
       +\textbf{Global Cache}(Ours) & \xmark & \underline{62.32} & \underline{52.63} & \underline{34.97} & & \underline{65.50} & \underline{62.50} & \underline{47.36} & \underline{54.21} \\
       +\textbf{Hierarchical Cache}(Ours) & \xmark & \textbf{64.42} & \textbf{56.63} & \textbf{35.77} & & \textbf{67.25} & \textbf{64.50} & \textbf{47.61} & \textbf{56.03} \\
       \midrule
       ULIP-2~\cite{xue24ulip2} & \xmark & 52.42 & 53.89 & 41.57 & & 51.50 & 59.25 & 46.35 & 50.83 \\ 
       +\textbf{Global Cache}(Ours) & \xmark & \underline{57.05} & \underline{59.37} & \underline{47.38} & & \underline{56.75} & \underline{65.75} & \underline{50.68} & \underline{56.16} \\ 
       +\textbf{Hierarchical Cache}(Ours) & \xmark & \textbf{59.16} & \textbf{60.84} & \textbf{49.87} & & \textbf{61.75} & \textbf{71.00} & \textbf{55.11} & \textbf{59.62} \\
       \midrule
       OpenShape~\cite{liu23openshape} & \xmark & 62.32 & 64.42 & 48.52 &  & 64.00 & 70.25 & 53.55 & 60.51 \\
       +\textbf{Global Cache}(Ours) & \xmark & \underline{65.68} & \underline{69.05} & \underline{49.36} & & \textbf{71.00} & \underline{71.50} & \underline{55.67} & \underline{63.71} \\
       + \textbf{Hierarchical Cache} & \xmark & \textbf{66.53} & \textbf{70.74} & \textbf{50.59} & & \underline{71.50} & 71.00 & \textbf{56.57} & \textbf{64.49} \\
       \midrule
       Uni3D~\cite{zhou24uni3d} & \xmark & 72.63 & 74.53 & 55.76 &  & 75.50 & 77.00 & 57.98 & 68.90 \\
       +\textbf{Global Cache}(Ours) & \xmark & \textbf{76.21} & \textbf{77.26} & \textbf{59.10} & & \underline{80.00} & \underline{81.00} & \underline{62.47} & \underline{72.67} \\
       +\textbf{Hierarchical Cache}(Ours) & \xmark & \underline{74.11} & \underline{76.00} & \underline{57.92} & & \textbf{83.00} & \textbf{81.50} & \textbf{63.98} & \textbf{72.75} \\
       \bottomrule
    \end{tabular}
    \label{tab:xset_sim2real_generalization}
\end{table*}

\begin{table*}[t]
    \footnotesize
    \centering
    \caption{\textbf{Comparison of recognition accuracy across a suite of datasets (no\_lvis weights)}. S-PB\_RS\_T50 is the hardest split of ScanObjectNN. O-LVIS: Objaverse-LVIS. Omni3D: OmniObject3D. In Omni3D, each point cloud can be represented by a different number of points (pts). Note that Omni3D has 216 classes and O-LVIS has 1,156 classes. The last column is the average accuracy on these datasets.}
    \label{tab:multi_dataset_generalization_no_lvis}
    \begin{tabular}{l c c c c c c c c }
       \toprule
       \multirow{2}{*}{Method} & \multirow{2}{*}{ModelNet40} & \multirow{2}{*}{S-PB\_RS\_T50} & \multirow{2}{*}{O-LVIS} & \multicolumn{3}{c}{Omni3D} & \multirow{2}{*}{\textbf{Avg.}} \\\cline{5-7} %
       & & & & 1024 pts & 4096 pts & 16384 pts \\
       \midrule
       O-Shape~\cite{liu23openshape} & 85.05 & 54.01 & \textbf{47.17} & 33.64 & 34.16 & 34.25 & 48.05 \\
       \ +\textbf{Global Cache}(Ours) & \textbf{85.74} & \textbf{57.06} & \underline{47.06} & \underline{37.11} & \textbf{38.53} & \textbf{38.07} & \textbf{50.60} \\
       \ +\textbf{Hierarchical Cache}(Ours) & \underline{85.70} & \underline{56.40} & 45.69 & \textbf{37.46} & \underline{38.36} & \underline{38.05} & \underline{50.28} \\ %
       \midrule
       Uni3D~\cite{zhou24uni3d} & 87.07 & 66.37 & \underline{47.24} & 30.08 & 38.10 & 38.04 & 51.15 \\ %
       \ +\textbf{Global Cache}(Ours) & \textbf{87.93} & \textbf{68.58} & \textbf{47.51} & \underline{33.23} & \textbf{39.51} & \underline{40.27} & \textbf{52.84} \\
       \ +\textbf{Hierarchical Cache}(Ours) & \underline{87.84} & \underline{67.96} & 46.81 & \textbf{33.91} & \underline{39.49} & \textbf{40.49} & \underline{52.75} \\ %
       \bottomrule
    \end{tabular}
\end{table*}

\subsection{Total Size of Full Hierarchical Cache}
Here we explain how to calculate the total size of full hierarchical cache. 
The variables listed in Tab.~\ref{tab:variables_for_cache_size_computation} are vital to decide the total size of full hierarchical cache $\textbf{C}_g \cup \textbf{C}_l$, including the feature dimension $d$ of $\textbf{e}_p^g$ and $\textbf{e}_p^l$, 
the upper bound $K$ on the number of samples in each category, the number of parts $m$ of a point cloud, and the number of classes $C$ in the dataset. 

Here we take (Uni3D, O-LVIS, Hierarchical Cache) as an example for computing the size of each item in the global and local cache. 
\begin{itemize}
  \item $\textbf{E}_g$: 1156*3*512 = 1,775,616
  \item $\hat{\textbf{L}}_g$: 1156*3 = 3,468
  \item $\textbf{h}_g$: 1156*3 = 3,468
  \item $\textbf{E}_l$: 1156*3*3*512 = 5,326,848
  \item $\hat{\textbf{L}}_l$: 1156*3*3 = 10,404
\end{itemize}
So the total size of full hierarchical cache for (Uni3D, O-LVIS) is sum of these items, approximately 7.1M. We present the 
parameter counting for other backbones in Tab.~\ref{tab:cache_size_count}. 
The results demonstrate the total size of a full hierarchical cache is very small, \eg, 7.1M, 
particularly when compared to the hundreds of 
millions of parameters in a large multimodal 3D model, \eg, 1016.5M in Uni3D. 
Therefore, Point-Cache introduces minimal additional computational and storage overhead, having little impact on memory usage and runtime efficiency, indicated by Tab.~\ref{tab:memory_comparison_uni3d} and Tab.~\ref{tab:efficiency_comparison_mn40} of the main paper. 

\begin{table}[t]
  \footnotesize
  \centering
  \caption{\textbf{Statistics of feature dimension $d$, number of shots $K$ per class, number of parts $m$ per point cloud, 
  and number of classes $C$ in the dataset}. The used dataset is O-LVIS.}
  \begin{tabular}{l r c c c}
    \toprule
    Backbone & $d$ dims & $K$ shots & $m$ parts & $C$ classes\\ 
    \midrule
    ULIP & 512 & 3 & 3 & 1,156 \\
    ULIP-2 & 512 & 3 & 3 & 1,156 \\
    OpenShape & 1,280 & 3 & 3 & 1,156 \\
    Uni3D & 512 & 3 & 3 & 1,156 \\
    \bottomrule
  \end{tabular}
  \label{tab:variables_for_cache_size_computation}
\end{table}

\begin{table}[t]
  \footnotesize
  \centering
  \caption{\textbf{Parameter count for the full hierarchical cache on O-LVIS, which covers 1,156 classes}. Capital letters in brackets indicate units of measurement.}
  \begin{tabular}{l c c c c r c r}
     \toprule
     \multirow{3}{*}{Backbone} & \multicolumn{3}{c}{Global Cache} & & \multicolumn{2}{c}{Local Cache} & Total \\\cline{2-4}\cline{6-7}
     &  $\textbf{E}_g$, & $\hat{\textbf{L}}_g$, & $\textbf{h}_g$ & & $\textbf{E}_l$, & $\hat{\textbf{L}}_l$ & Size \\
     & (K) & (K) & (K) & & (K) & (K) & (M) \\
     \midrule
     ULIP & 1775.6 & 3.5 & 3.5 & & 5326.8 & 10.4 & 7.1 \\
     ULIP-2 & 1775.6 & 3.5 & 3.5 & & 5326.8 & 10.4 & 7.1 \\
     O-Shape & 4439.0 & 3.5 & 3.5 & & 13,317.1 & 10.4 & 17.8 \\
     Uni3D & 1775.6 & 3.5 & 3.5 & & 5326.8 & 10.4 & 7.1 \\ 
     \bottomrule
  \end{tabular}
  \label{tab:cache_size_count}
\end{table}

\subsection{Memory Usage and Throughput}
\noindent\textbf{Memory.}
We compare the memory usage based on other large 3D models such as ULIP, ULIP-2 and OpenShape. 
In the experiments, a point cloud contains 1,024 points. 
The results are recorded in 
Tab.~\ref{tab:memory_comparison_ulip},~\ref{tab:memory_comparison_ulip2} and~\ref{tab:memory_comparison_openshape}. 
\#Params count the total parameters in a large multimodal 3D model. 
We observe that our global and hierarchical cache model utilize same or slightly higher memory compared to the zero-shot baseline 
across backbones and datasets. For instance, OpenShape powered by our global cache consumes 7,058 MB GPU memory, same as the usage 
of zero-shot OpenShape. 
Moreover, with the number of 3D classes increasing rapidly, \eg, $40\rightarrow216\rightarrow1,156$, 
the memory rises slowly, \eg, $1,556\rightarrow1,558\rightarrow1,570$ for ULIP-2 with our hierarchical cache. 
The reason is same as we explained in the main paper: memory consumption is dominated by the numerous parameters of 
the large 3D model (\eg, 32.3M \#Params in the OpenShape point encoder alone) and the overhead of Point-Cache is ignorable. 

\begin{table}[t]
    \footnotesize
    \centering
    \caption{\textbf{Comparison of memory usage (MB) based on ULIP.} 
    The batch size is set to 1, and the experiments are conducted on an RTX 4090. The number below each dataset name indicates \#Classes.}
    \begin{tabular}{l c c c c}
        \toprule
        \multirow{2}{*}{Method} & ModelNet-C & Omni3D & O-LVIS & \multirow{2}{*}{\#Params} \\
        & (40) & (216) & (1,156) & \\
        \midrule
        ULIP & \textbf{1,556} & \textbf{1,558} & \textbf{1,556} & 85.7M \\ %
        \ +\textbf{Global}(Ours) & \textbf{1,556} & \textbf{1,558} & \underline{1,560} & 85.7M \\
        \ +\textbf{Hierar}(Ours) &\textbf{1,556} & \textbf{1,558} & 1,566 & 85.7M \\
        \bottomrule
    \end{tabular}
    \label{tab:memory_comparison_ulip}
\end{table}

\begin{table}[t]
    \footnotesize
    \centering
    \caption{\textbf{Comparison of memory usage (MB) based on ULIP-2.} 
    The batch size is set to 1, and the experiments are conducted on an RTX 4090. The number below each dataset name indicates \#Classes.}
    \begin{tabular}{l c c c c}
        \toprule
        \multirow{2}{*}{Method} & ModelNet-C & Omni3D & O-LVIS & \multirow{2}{*}{\#Params} \\
        & (40) & (216) & (1,156) & \\
        \midrule
        ULIP-2 & \textbf{1,556} & \textbf{1,558} & \textbf{1,556} & 85.7M \\ %
        \ +\textbf{Global}(Ours) & \textbf{1,556} & \textbf{1,558} & \underline{1,560} & 85.7M \\
        \ +\textbf{Hierar}(Ours) & \textbf{1,556} & \textbf{1,558} & 1,570 & 85.7M\\
        \bottomrule
    \end{tabular}
    \label{tab:memory_comparison_ulip2}
\end{table}

\begin{table}[t]
  \footnotesize
  \centering
  \caption{\textbf{Comparison of memory usage (MB) based on OpenShape.} 
  The batch size is set to 1, and the experiments are conducted on an RTX 4090. The number below each dataset name indicates \#Classes.}
  \begin{tabular}{l c c c c}
     \toprule
     \multirow{2}{*}{Method} & ModelNet-C & Omni3D & O-LVIS & \multirow{2}{*}{\#Params} \\
     & (40) & (216) & (1,156) &  \\
     \midrule
     OpenShape & \textbf{7,056} & \textbf{7,058} & \textbf{7,116} & 2,571.9M \\ %
     \ +\textbf{Global}(Ours) & \textbf{7,056} & \textbf{7,058} & \underline{7,126} & 2,571.9M \\
     \ +\textbf{Hierar}(Ours) & \underline{7,058} & \underline{7,062} & 7,150 & 2,571.9M\\
     \bottomrule
  \end{tabular}
  \label{tab:memory_comparison_openshape}
\end{table}

\begin{table}[t]
    \footnotesize
    \centering
    \caption{\textbf{Comparison of memory usage (MB) based on Uni3D.} 
    The batch size is set to 1, and the experiments are conducted on an RTX 4090. The number below each dataset name indicates \#Classes.}
    \begin{tabular}{l c c c c}
       \toprule
       \multirow{2}{*}{Method} & ModelNet-C & Omni3D & O-LVIS & \multirow{2}{*}{\#Params} \\
       & (40) & (216) & (1,156) &  \\
       \midrule
       Uni3D & \textbf{5,062} & \textbf{5,062} & \textbf{5,062} & 1711.7M \\ %
       \ +\textbf{Global}(Ours) & \textbf{5,062} & \underline{5,064} & \underline{5,070} & 1711.7M \\
       \ +\textbf{Hierar}(Ours) & \underline{5,064} & 5,068 & 5,090 & 1711.7M\\
       \bottomrule
    \end{tabular}
    \label{tab:memory_comparison_uni3d_suppl}
\end{table}

\noindent\textbf{Throughput.}
We test the throughput of Point-Cache on S-OBJ\_ONLY and report the results in Tab.~\ref{tab:efficiency_comparison_sonn_c_obj_only_clean}. 
The throughput is measured by the number of test samples per second ($t/s$) the model can process. 
Models with our global and hierarchical cache run slightly slower than zero-shot inference, 
\eg, a 0.03 $t/s$ drop for OpenShape with global cache and a 0.05 $t/s$ drop for OpenShape with hierarchical cache, 
suggesting little computational overhead introduced by Point-Cache. 
In theory, the throughput is decided by the model itself and the GPU device used, instead of the dataset. 
In practice, the throughput on S-OBJ\_ONLY is consistent with that on ModelNet40, as shown in Tab.~\ref{tab:efficiency_comparison_mn40} of the main paper. 

\begin{table}[t]
  \footnotesize
  \centering
  \caption{\textbf{Comparison of running throughput ($t/s$) for different models on S-OBJ\_ONLY.} 
  Each point cloud contains 1024 points. The batch size is set to 1 and the used device is an RTX 4090. 
  The results are averaged over all test samples.}
  \begin{tabular}{l r r r}
     \toprule
     Method & Zero-shot & +\textbf{Global}(Ours) & +\textbf{Hierar}(Ours) \\
     \midrule
     ULIP  & \textbf{11.19} & \underline{11.16} & 11.14 \\ %
     ULIP-2 & \textbf{11.19} & \underline{11.15} & 11.14 \\
     OpenShape & \textbf{9.86} & \underline{9.83} & 9.81 \\
     Uni3D & \textbf{9.62} & \underline{9.59} & 9.58 \\
     \bottomrule
  \end{tabular}
  \label{tab:efficiency_comparison_sonn_c_obj_only_clean}
\end{table}

\subsection{Other Cache Models}
\textbf{Comparison with other cache models.} 
There are only a few 3D point cloud cache models (Point-PEFT [46], BFTT3D [59] and Point-NN [73]). They have different pipelines and settings, making fair comparisons difficult, \eg, (1) they use the entire training set (\emph{with} real labels) to construct the cache offline, 
whereas Point-Cache constructs the cache using test data (\emph{without} real labels) online; 
(2) they are not based on large 3D models and cannot recognize new classes in Omni3D and O-LVIS. 
In Tab.~\ref{tab:comparison_with_other_cache_model}, we add comparisons with Point-NN (not a test-time method). 
The performance of Point-NN is expectedly better since it uses the real labels and the whole training set to build the cache. 

\begin{table}[t]
    \scriptsize
    \centering
    \caption{\textbf{Comparison with Point-NN}. The comparison is unfair
    since Point-NN uses the training set to construct an offline cache.}
    \begin{tabular}{l c c c c}
       \toprule
       \multirow{2}{*}{Model} & ModelNet & \multicolumn{3}{c}{ScanObjectNN (1,024 points)} \\\cline{3-5}
        & (1,024 points) & OBJ\_ONLY & OBJ\_BG & PB\_T50\_RS \\ 
       \toprule
       Point-NN & 81.65 & 72.46 & 71.26 & 62.80  \\ %
       \midrule
       Uni3D & 81.81 & 65.58 & 60.24 & 46.04 \\ %
       \ +\textbf{Global} & \underline{83.14} & \underline{70.05} & \textbf{63.86} & \underline{50.28} \\ %
       \ +\textbf{Hierar} & \textbf{83.87} & \textbf{70.22} & \underline{62.82} & \textbf{51.53} \\ %
       \bottomrule
    \end{tabular}
    \label{tab:comparison_with_other_cache_model}
\end{table}

\section{Visualization}

\subsection{Relation with Previous Methods}
Tab.~\ref{tab:comparison_with_existing_cache_models} highlights the differences 
between existing cache models and Point-Cache. The proposed approach is a \emph{dynamic} and \emph{hierarchical} cache model that 
is constructed \emph{entirely based on test data} for test-time point cloud recognition. 

\begin{table}[t]
    \footnotesize
    \centering
    \caption{\textbf{Comparison with other cache models}. In the first row, we select several key attributes of 
    the cache models for comparison. `Test-time' means whether the model is developed for test-time adaptation. `T-set' 
    indicates whether the cache model is solely built on the test set. `Dynamic' and `Hierarchical' 
    represent whether the cache is dynamically managed and designed in a coarse-to-fine manner, respectively.}
    \begin{tabular}{l c c c c}
       \toprule
       Cache Model & Test-time & T-set only & Dynamic & Hierarchical \\
       \midrule
       Point-NN~\cite{zhang23pointnn} & \xmark & \xmark & \xmark & \xmark \\
       Point-PEFT~\cite{tang24pointpeft} & \xmark & \xmark & \xmark & \xmark \\
       BFTT3D~\cite{wang24bftt3d} & \checkmark & \xmark & \xmark & \xmark\\
       \textbf{Point-Cache} & \checkmark & \checkmark & \checkmark & \checkmark \\
       \bottomrule
    \end{tabular}
    \label{tab:comparison_with_existing_cache_models}
\end{table}

\subsection{Point Cloud Encoding} 
\begin{figure*}[b]
  \centering
  \includegraphics[width=\textwidth]{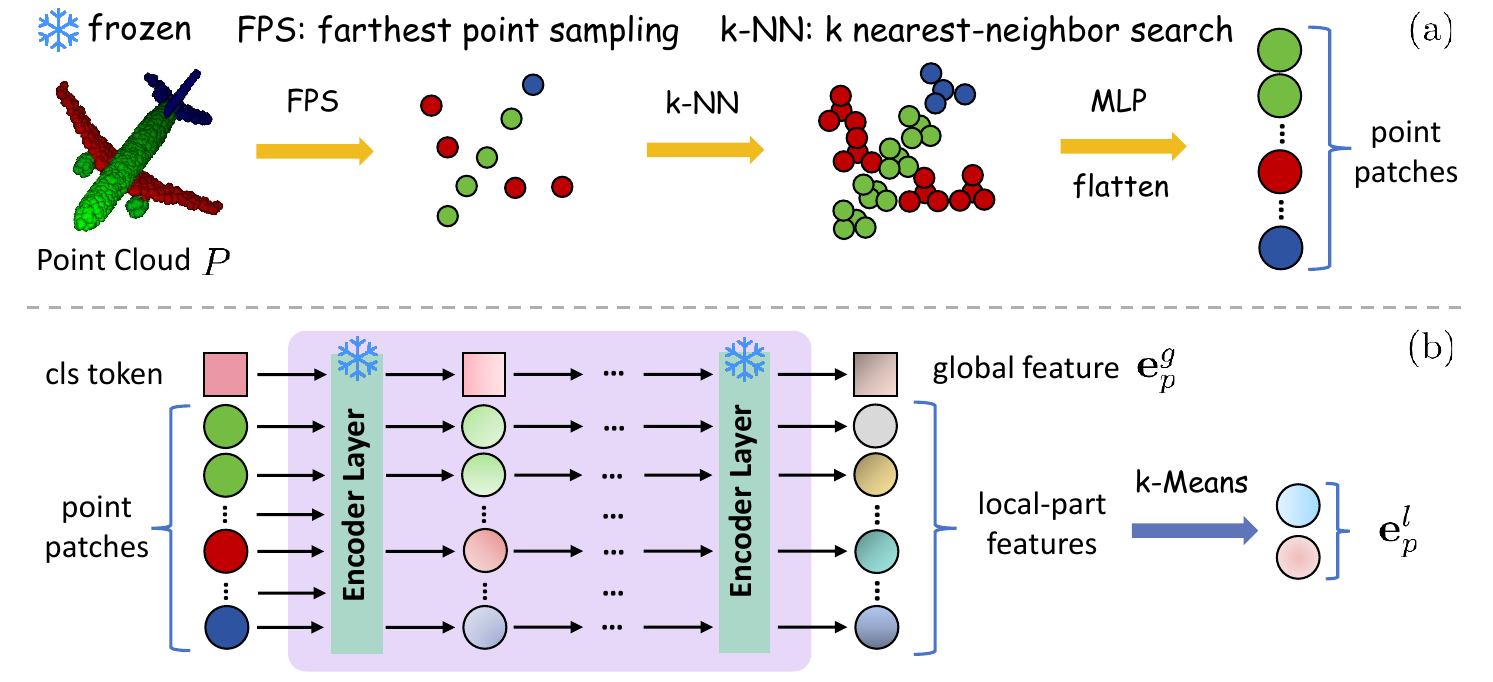}
  \caption{\textbf{Visualization of point cloud encoding}. Subfigure (a) illustrates 
  the process of producing point patches, while subfigure (b) explains 
  how the global feature and local-part features are generated.}
  \label{fig:point_cloud_encoding}
\end{figure*}

Fig.~\ref{fig:point_cloud_encoding} illustrates the detailed process of 
point cloud encoding, corresponding to the `Encode' component of Fig.~\ref{fig:architecture} in 
the main paper. 
For an input point cloud $P \in \mathbb{R}^{N\times 3}$, we first perform \emph{farthest point 
sampling} to obtain $M$ key points. Next, we search for $k$ \emph{nearest neighbors} for each key point to form 
$M$ local point patches, which are transformed by a lightweight neural network (\eg, 2-layer MLP in ViPFormer~\cite{sun23vipformer} or mini-PointNet~\cite{qi17pointnet} in PointBert~\cite{yu22pointbert}),
as shown in Fig.~\ref{fig:point_cloud_encoding} (a). 
Subsequently, a class token, along with the flattened point patches, is fed into the Transformer-based point encoder, generating the global feature $\mathbf{e}_p^g \in \mathbb{R}^d$ and local-part features. 
To save memory and computation, these local-part features (\eg. 512 in ULIP-2\cite{xue24ulip2}) into $m$ (\eg, 5) centers using K-Means, resulting in
$\mathbf{e}_p^l \in \mathbb{R}^{m\times d}$, as depicted in Fig.~\ref{fig:point_cloud_encoding} (b). 

\begin{figure*}[b]
    \centering
    \includegraphics[width=\columnwidth]{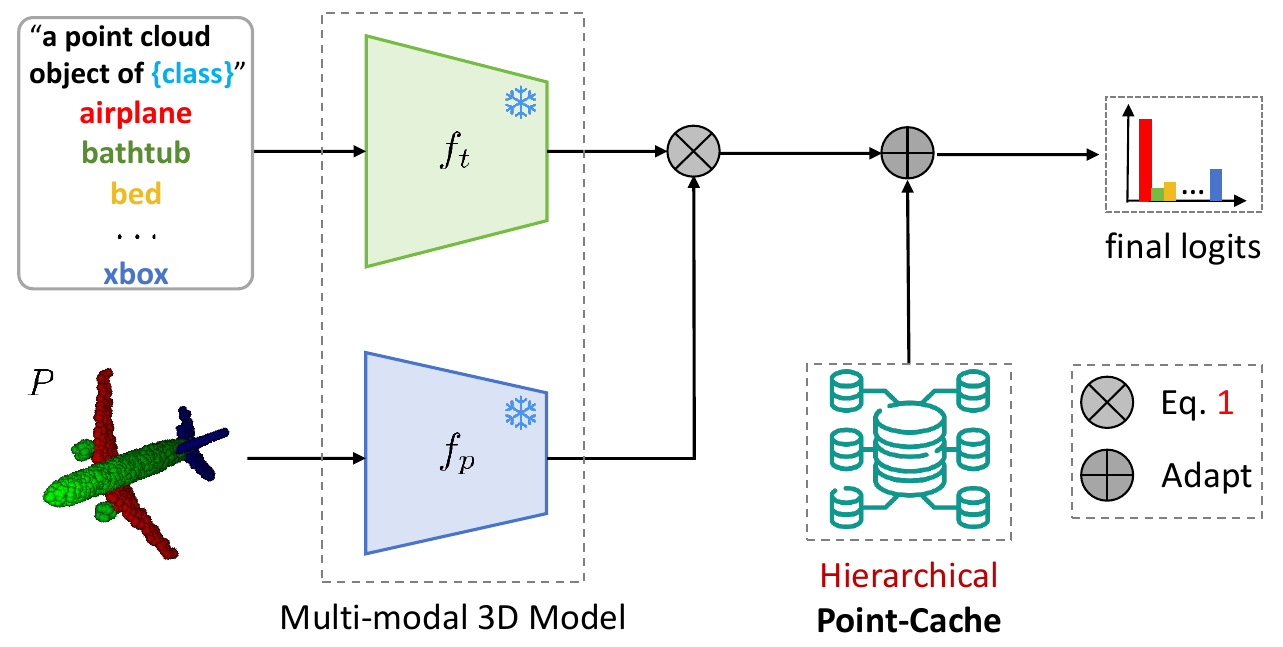}
    \caption{\textbf{Illustration of Point-Cache as a plug-and-play module}. We record global and local-part 3D 
    features of high-quality test samples in the hierarchical cache. Then the hierarchical cache can be employed to adapt the zero-shot predictions of
    various large multi-modal 3D models.}
    \label{fig:illustration_of_plug_and_play_module}
\end{figure*}

\subsection{The Global and Local Cache}
\begin{figure*}[b]
  \centering
  \includegraphics[width=\textwidth]{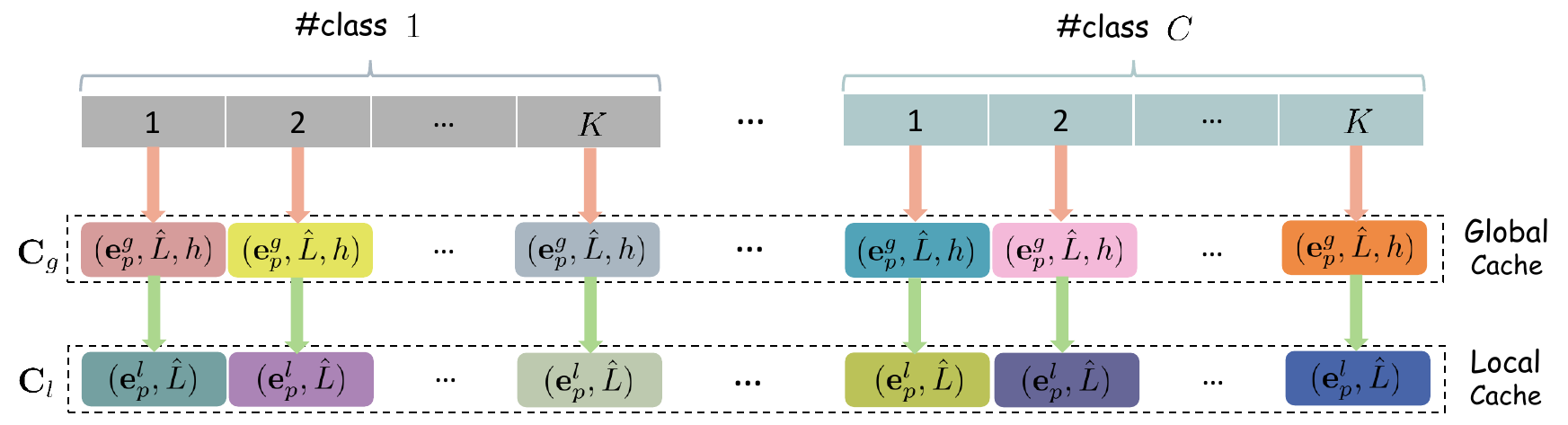}
  \caption{\textbf{Simulation of the global cache and local cache.} An upper bound $K$ is set for the number of samples per class in the cache. The local fingerprint $(\mathbf{e}_p^l, \hat{L})$ of a sample $P$ is stored in the local cache only if its global fingerprint $(\mathbf{e}_p^g, \hat{L}, h)$ 
  qualifies for inclusion in the global cache. The global and local caches are initially empty and then updated according to Alg.~\ref{alg:hierarchical_cache_building_and_adaptation}. However, the global and local caches are not necessarily full. The full status of Point-Cache is determined by storing $K$ global and local fingerprints in each of $C$ categories.}
  \label{fig:global_and_local_cache}
\end{figure*}

Fig.~\ref{fig:global_and_local_cache} visualizes the global cache $\mathbf{C}_g$ and the local cache $\mathbf{C}_l$. 
$\mathbf{C}_g$ stores up to $K$ global fingerprints $(\mathbf{e}_p^g, \hat{L}, h)$ per class from online test samples,  
while $\mathbf{C}_l$ records the local fingerprints $(\mathbf{e}_p^l, \hat{L})$ of corresponding samples. 
The global and local caches are empty at the beginning and then accept the fingerprints of online test samples.  
Both $\mathbf{C}_g$ and $\mathbf{C}_l$ are dynamically managed to prioritize high-quality samples, as outlined in Alg.~\ref{alg:hierarchical_cache_building_and_adaptation}. 
Note that the global and local caches are not necessarily full. 
This hierarchical design and the selective mechanism enable the creation of an more accurate profile for  
test data than previous cache methods, facilitating robust and generalizable point cloud analysis at test time. 

\subsection{Qualitative Analysis}
\begin{figure*}[b]
  \centering
  \begin{subfigure}{0.25\textwidth}
     \centering
     \includegraphics[width=\columnwidth]{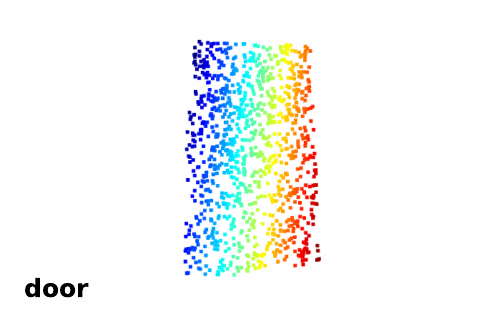}
     \caption{Ground truth.}
     \label{fig:vis_ulip1_sonn_c_obj_only_rotate_2_gt_door}
  \end{subfigure}%
  \begin{subfigure}{0.25\textwidth}
     \centering
     \includegraphics[width=\columnwidth]{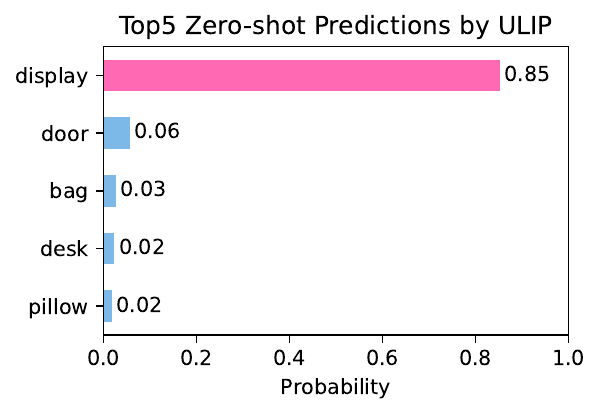}
     \caption{ULIP.}
     \label{fig:vis_ulip1_sonn_c_obj_only_rotate_2_zero_door}
  \end{subfigure}%
  \begin{subfigure}{0.25\textwidth}
     \centering
     \includegraphics[width=\columnwidth]{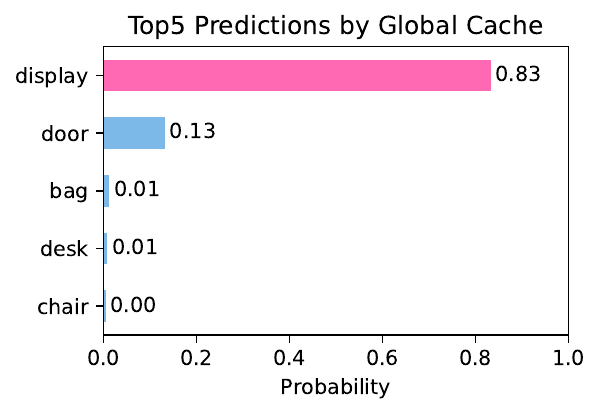}
     \caption{ULIP+\textbf{GC}.}
     \label{fig:vis_ulip1_sonn_c_obj_only_rotate_2_global_door}
  \end{subfigure}%
  \begin{subfigure}{0.25\textwidth}
     \centering
     \includegraphics[width=\columnwidth]{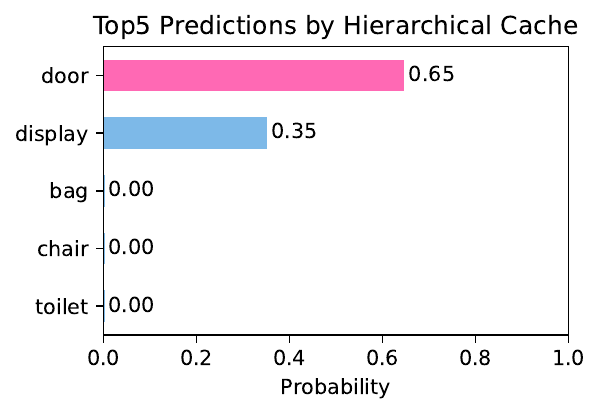}
     \caption{ULIP+\textbf{HC}.}
     \label{fig:vis_ulip1_sonn_c_obj_only_rotate_2_hierar_door}
  \end{subfigure}%

  \begin{subfigure}{0.25\textwidth}
      \centering
      \includegraphics[width=\columnwidth]{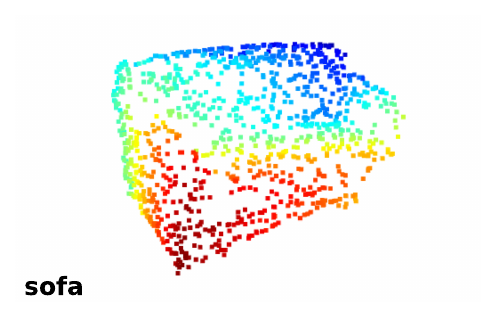}
      \caption{Ground truth.}
      \label{fig:vis_ulip1_sonn_c_obj_only_rotate_2_gt_sofa}
  \end{subfigure}%
  \begin{subfigure}{0.25\textwidth}
      \centering
      \includegraphics[width=\columnwidth]{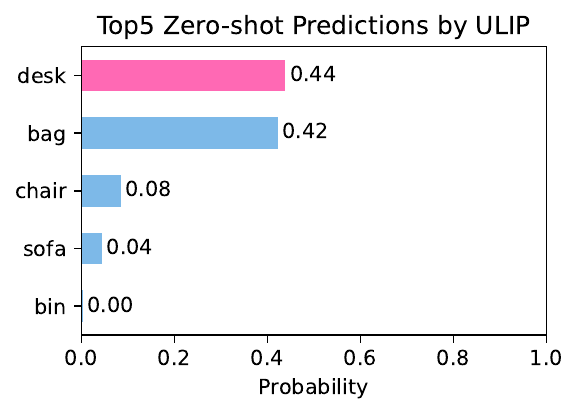}
      \caption{ULIP.}
      \label{fig:vis_ulip1_sonn_c_obj_only_rotate_2_zero_sofa}
  \end{subfigure}%
  \begin{subfigure}{0.25\textwidth}
      \centering
      \includegraphics[width=\columnwidth]{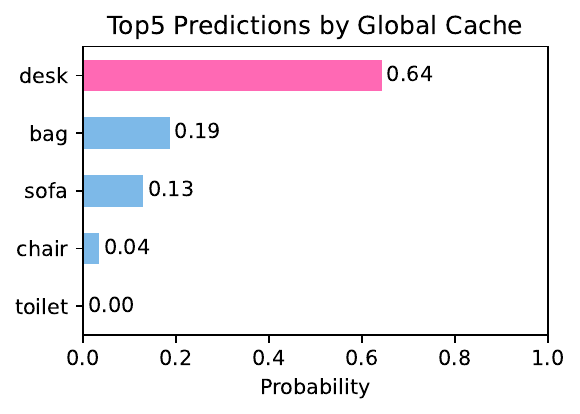}
      \caption{ULIP+\textbf{GC}.}
      \label{fig:vis_ulip1_sonn_c_obj_only_rotate_2_global_sofa}
  \end{subfigure}%
  \begin{subfigure}{0.25\textwidth}
      \centering
      \includegraphics[width=\columnwidth]{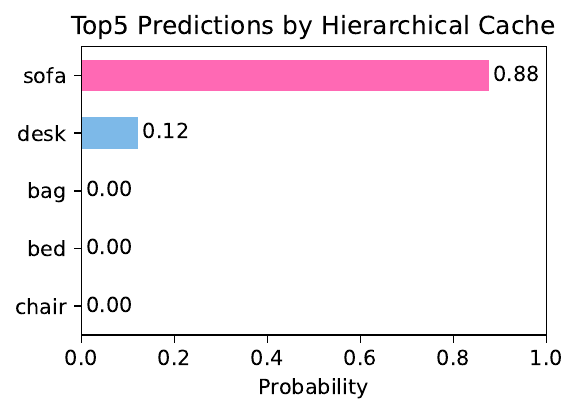}
      \caption{ULIP+\textbf{HC}.}
      \label{fig:vis_ulip1_sonn_c_obj_only_rotate_2_hierar_sofa}
  \end{subfigure}%

  \begin{subfigure}{0.25\textwidth}
    \centering
    \includegraphics[width=\columnwidth]{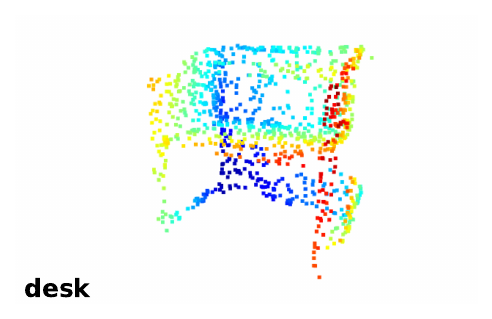}
    \caption{Ground truth.}
    \label{fig:vis_ulip1_sonn_c_obj_only_rotate_2_gt_desk}
  \end{subfigure}%
  \begin{subfigure}{0.25\textwidth}
      \centering
      \includegraphics[width=\columnwidth]{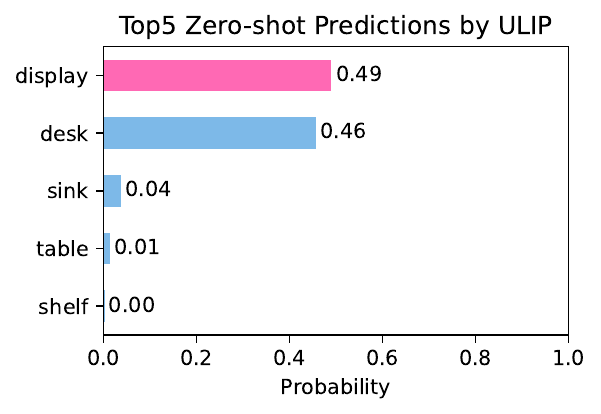}
      \caption{ULIP.}
      \label{fig:vis_ulip1_sonn_c_obj_only_rotate_2_zero_desk}
  \end{subfigure}%
  \begin{subfigure}{0.25\textwidth}
      \centering
      \includegraphics[width=\columnwidth]{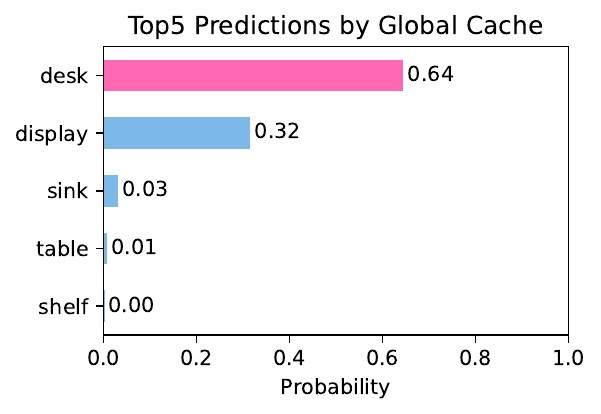}
      \caption{ULIP+\textbf{GC}.}
      \label{fig:vis_ulip1_sonn_c_obj_only_rotate_2_global_desk}
  \end{subfigure}%
  \begin{subfigure}{0.25\textwidth}
      \centering
      \includegraphics[width=\columnwidth]{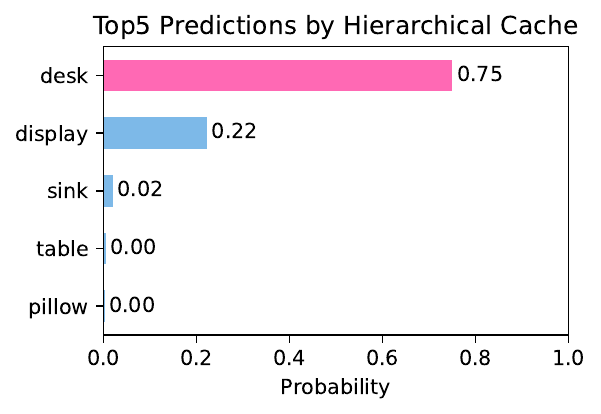}
      \caption{ULIP+\textbf{HC}.}
      \label{fig:vis_ulip1_sonn_c_obj_only_rotate_2_hierar_desk}
  \end{subfigure}%

  \begin{subfigure}{0.25\textwidth}
     \centering
     \includegraphics[width=\columnwidth]{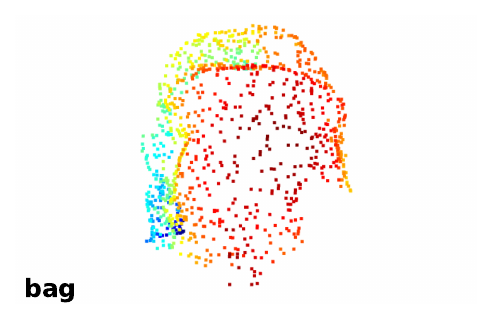}
     \caption{Ground truth.}
     \label{fig:vis_ulip1_sonn_c_obj_only_rotate_2_gt_bag}
  \end{subfigure}%
  \begin{subfigure}{0.25\textwidth}
     \centering
     \includegraphics[width=\columnwidth]{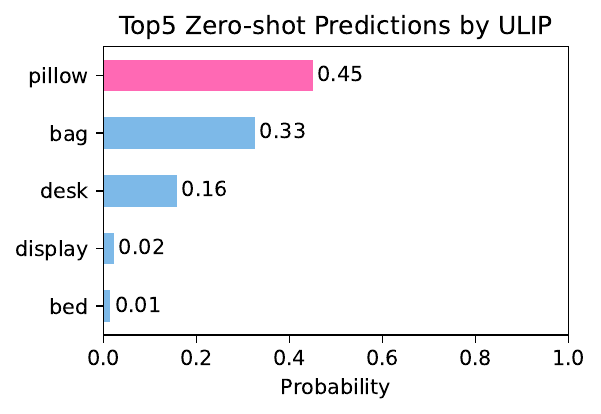}
     \caption{ULIP.}
     \label{fig:vis_ulip1_sonn_c_obj_only_rotate_2_zero_bag}
  \end{subfigure}%
  \begin{subfigure}{0.25\textwidth}
     \centering
     \includegraphics[width=\columnwidth]{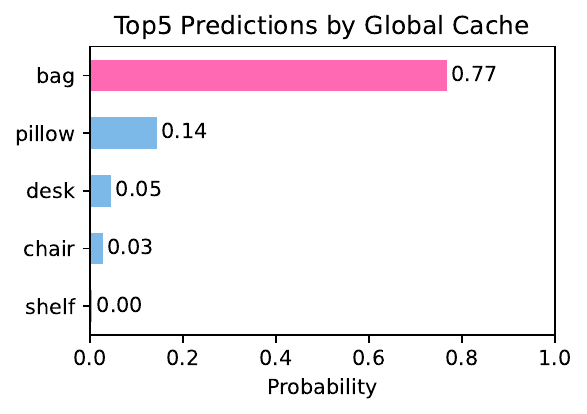}
     \caption{ULIP+\textbf{GC}.}
     \label{fig:vis_ulip1_sonn_c_obj_only_rotate_2_global_bag}
  \end{subfigure}%
  \begin{subfigure}{0.25\textwidth}
     \centering
     \includegraphics[width=\columnwidth]{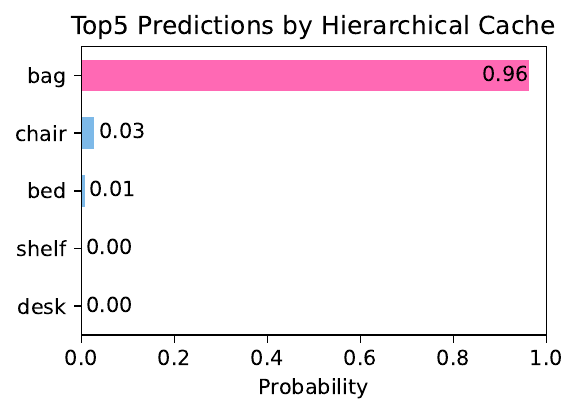}
     \caption{ULIP+\textbf{HC}.}
     \label{fig:vis_ulip1_sonn_c_obj_only_rotate_2_hierar_bag}
  \end{subfigure}%

  \begin{subfigure}{0.25\textwidth}
    \centering
    \includegraphics[width=\columnwidth]{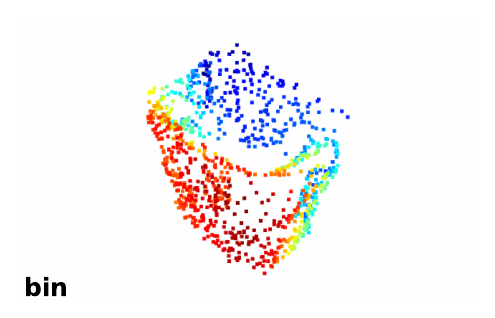}
    \caption{Ground truth.}
    \label{fig:vis_ulip1_sonn_c_obj_only_rotate_2_gt_bin}
  \end{subfigure}%
  \begin{subfigure}{0.25\textwidth}
      \centering
      \includegraphics[width=\columnwidth]{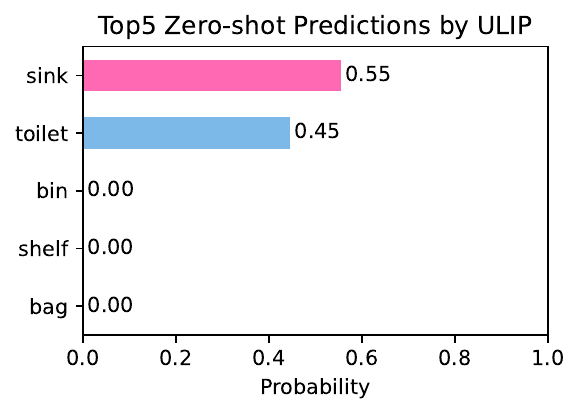}
      \caption{ULIP.}
      \label{fig:vis_ulip1_sonn_c_obj_only_rotate_2_zero_bin}
  \end{subfigure}%
  \begin{subfigure}{0.25\textwidth}
      \centering
      \includegraphics[width=\columnwidth]{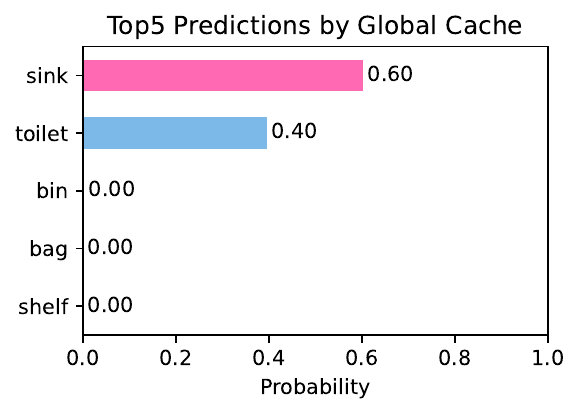}
      \caption{ULIP+\textbf{GC}.}
      \label{fig:vis_ulip1_sonn_c_obj_only_rotate_2_global_bin}
  \end{subfigure}%
  \begin{subfigure}{0.25\textwidth}
      \centering
      \includegraphics[width=\columnwidth]{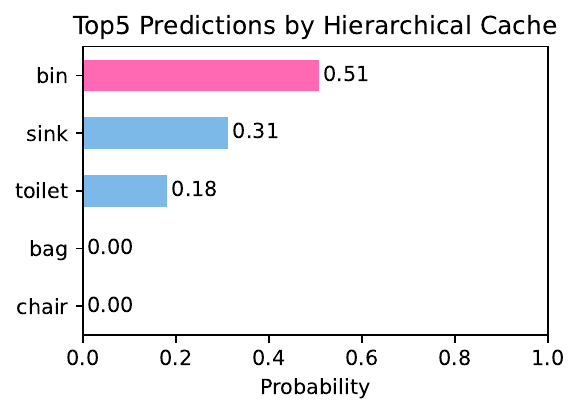}
      \caption{ULIP+\textbf{HC}.}
      \label{fig:vis_ulip1_sonn_c_obj_only_rotate_2_hierar_bin}
  \end{subfigure}%

  \begin{subfigure}{0.25\textwidth}
    \centering
    \includegraphics[width=\columnwidth]{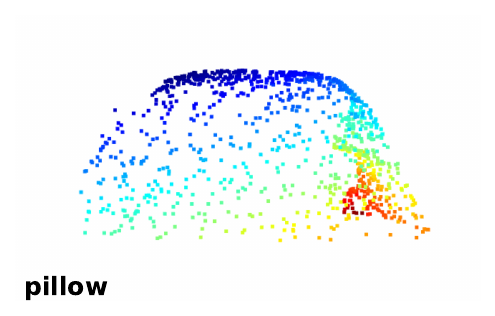}
    \caption{Ground truth.}
    \label{fig:vis_ulip1_sonn_c_obj_only_rotate_2_gt_pillow}
  \end{subfigure}%
  \begin{subfigure}{0.25\textwidth}
      \centering
      \includegraphics[width=\columnwidth]{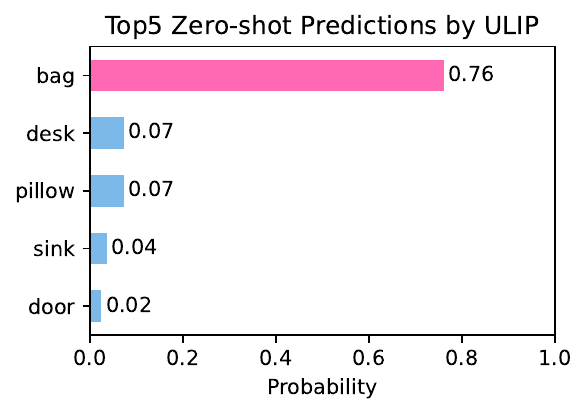}
      \caption{ULIP.}
      \label{fig:vis_ulip1_sonn_c_obj_only_rotate_2_zero_pillow}
  \end{subfigure}%
  \begin{subfigure}{0.25\textwidth}
      \centering
      \includegraphics[width=\columnwidth]{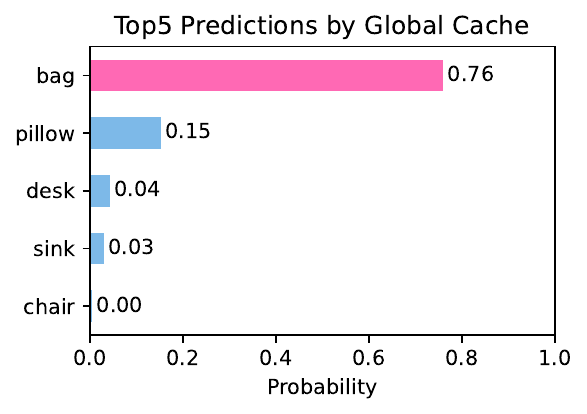}
      \caption{ULIP+\textbf{GC}.}
      \label{fig:vis_ulip1_sonn_c_obj_only_rotate_2_global_pillow}
  \end{subfigure}%
  \begin{subfigure}{0.25\textwidth}
      \centering
      \includegraphics[width=\columnwidth]{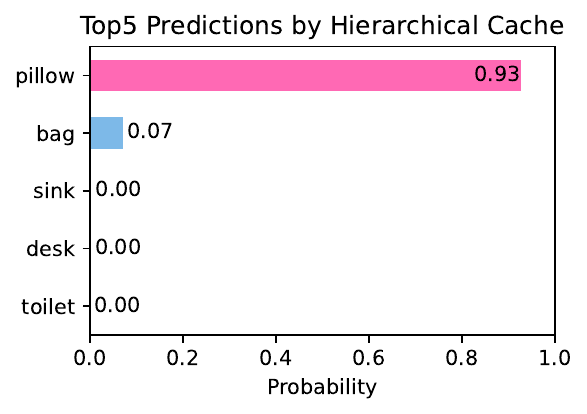}
      \caption{ULIP+\textbf{HC}.}
      \label{fig:vis_ulip1_sonn_c_obj_only_rotate_2_hierar_pillow}
  \end{subfigure}%
  \caption{\textbf{ULIP zero-shot predictions before and after adaptation by Point-Cache.} The used dataset S-OBJ\_ONLY-C 
  (rotate, severity=2). Each 3D object contains 1,024 points. \textbf{GC}: global cache. \textbf{HC}: hierarchical cache.}
  \label{fig:vis_lm3d_dataset_adaptation_suppl_ulip}
\end{figure*}

\begin{figure*}[b]
  \centering
  \begin{subfigure}{0.25\textwidth}
     \centering
     \includegraphics[width=\columnwidth]{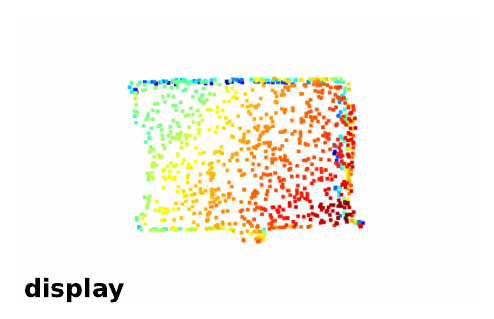}
     \caption{Ground truth.}
     \label{fig:vis_ulip2_sonn_hardest_gt_display}
  \end{subfigure}%
  \begin{subfigure}{0.25\textwidth}
     \centering
     \includegraphics[width=\columnwidth]{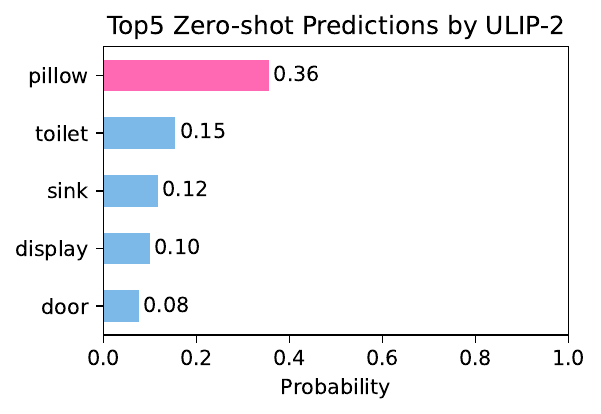}
     \caption{ULIP-2.}
     \label{fig:vis_ulip2_sonn_hardest_zero_display}
  \end{subfigure}%
  \begin{subfigure}{0.25\textwidth}
     \centering
     \includegraphics[width=\columnwidth]{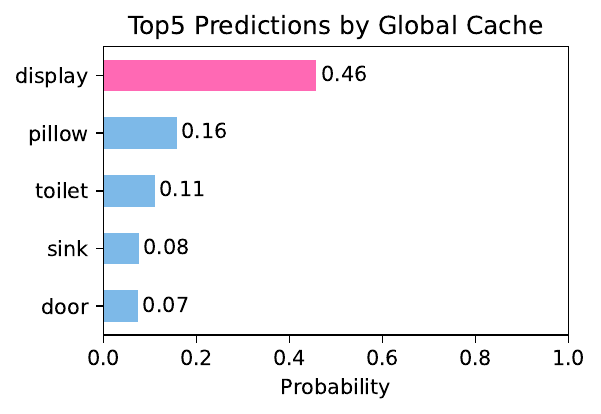}
     \caption{ULIP-2+\textbf{GC}.}
     \label{fig:vis_ulip2_sonn_hardest_global_display}
  \end{subfigure}%
  \begin{subfigure}{0.25\textwidth}
     \centering
     \includegraphics[width=\columnwidth]{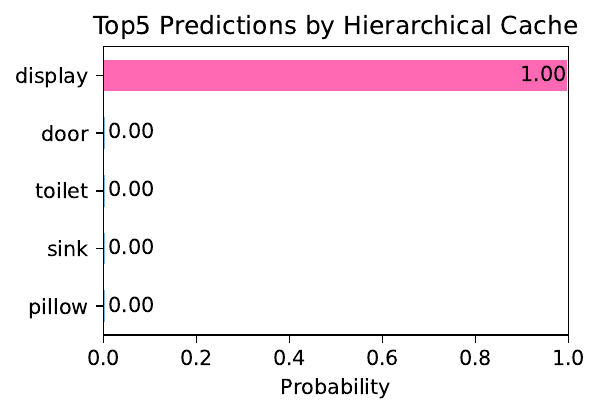}
     \caption{ULIP-2+\textbf{HC}.}
     \label{fig:vis_ulip2_sonn_hardest_hierar_display}
  \end{subfigure}%

  \begin{subfigure}{0.25\textwidth}
      \centering
      \includegraphics[width=\columnwidth]{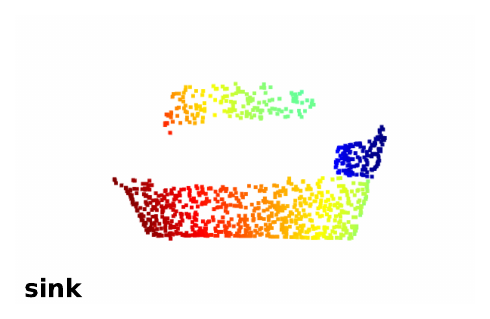}
      \caption{Ground truth.}
      \label{fig:vis_ulip2_sonn_hardest_gt_sink}
  \end{subfigure}%
  \begin{subfigure}{0.25\textwidth}
      \centering
      \includegraphics[width=\columnwidth]{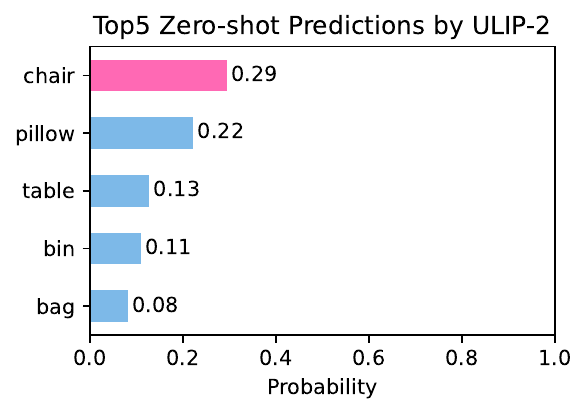}
      \caption{ULIP-2.}
      \label{fig:vis_ulip2_sonn_hardest_zero_sink}
  \end{subfigure}%
  \begin{subfigure}{0.25\textwidth}
      \centering
      \includegraphics[width=\columnwidth]{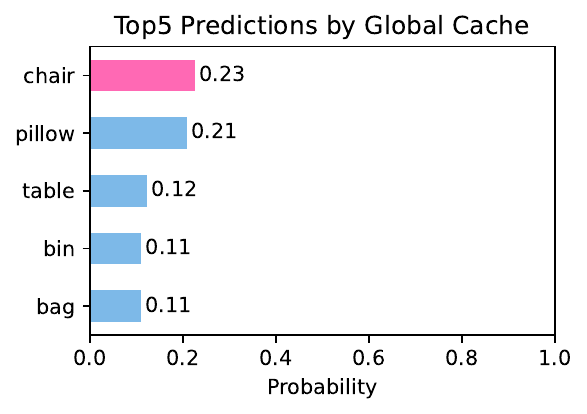}
      \caption{ULIP-2+\textbf{GC}.}
      \label{fig:vis_ulip2_sonn_hardest_global_sink}
  \end{subfigure}%
  \begin{subfigure}{0.25\textwidth}
      \centering
      \includegraphics[width=\columnwidth]{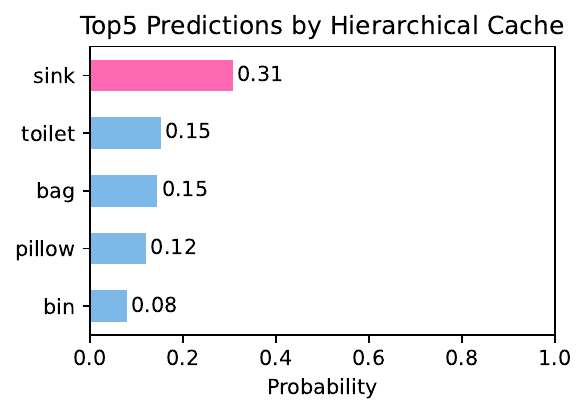}
      \caption{ULIP-2+\textbf{HC}.}
      \label{fig:vis_ulip2_sonn_hardest_hierar_sink}
  \end{subfigure}%

  \begin{subfigure}{0.25\textwidth}
    \centering
    \includegraphics[width=\columnwidth]{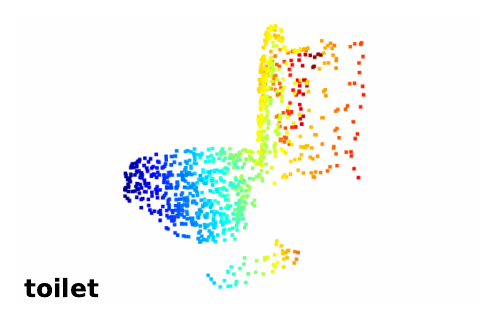}
    \caption{Ground truth.}
    \label{fig:vis_ulip2_sonn_hardest_gt_toilet}
  \end{subfigure}%
  \begin{subfigure}{0.25\textwidth}
      \centering
      \includegraphics[width=\columnwidth]{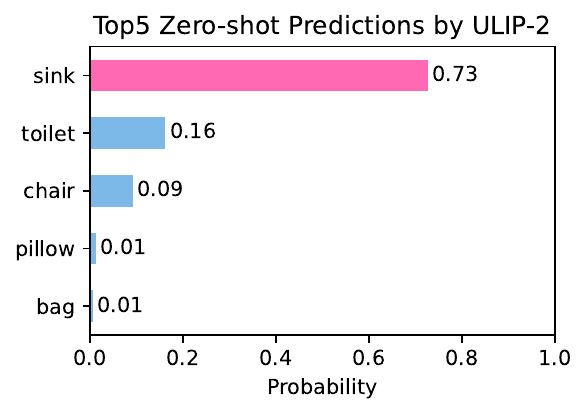}
      \caption{ULIP-2.}
      \label{fig:vis_ulip2_sonn_hardest_zero_toilet}
  \end{subfigure}%
  \begin{subfigure}{0.25\textwidth}
      \centering
      \includegraphics[width=\columnwidth]{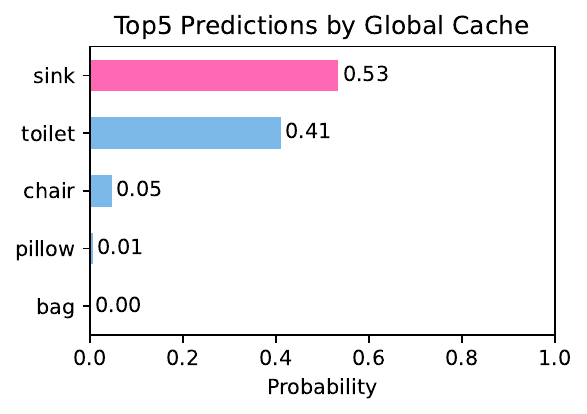}
      \caption{ULIP-2+\textbf{GC}.}
      \label{fig:vis_ulip2_sonn_hardest_global_toilet}
  \end{subfigure}%
  \begin{subfigure}{0.25\textwidth}
      \centering
      \includegraphics[width=\columnwidth]{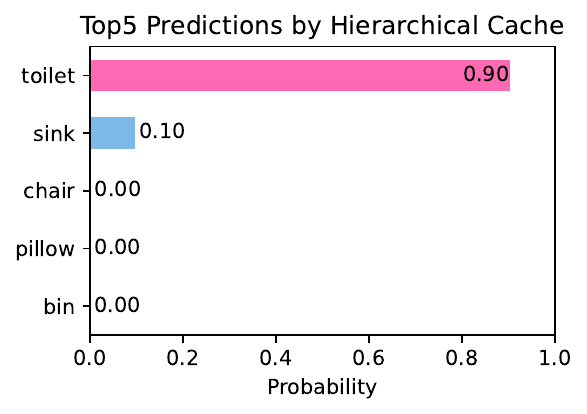}
      \caption{ULIP-2+\textbf{HC}.}
      \label{fig:vis_ulip2_sonn_hardest_hierar_toilet}
  \end{subfigure}%

  \begin{subfigure}{0.25\textwidth}
     \centering
     \includegraphics[width=\columnwidth]{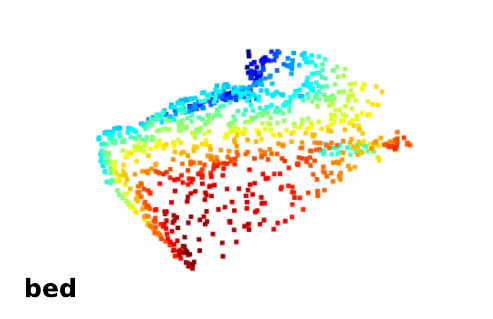}
     \caption{Ground truth.}
     \label{fig:vis_ulip2_sonn_hardest_gt_bed}
  \end{subfigure}%
  \begin{subfigure}{0.25\textwidth}
     \centering
     \includegraphics[width=\columnwidth]{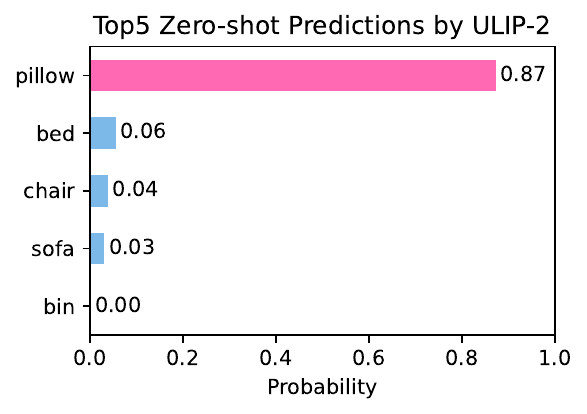}
     \caption{ULIP-2.}
     \label{fig:vis_ulip2_sonn_hardest_zero_bed}
  \end{subfigure}%
  \begin{subfigure}{0.25\textwidth}
     \centering
     \includegraphics[width=\columnwidth]{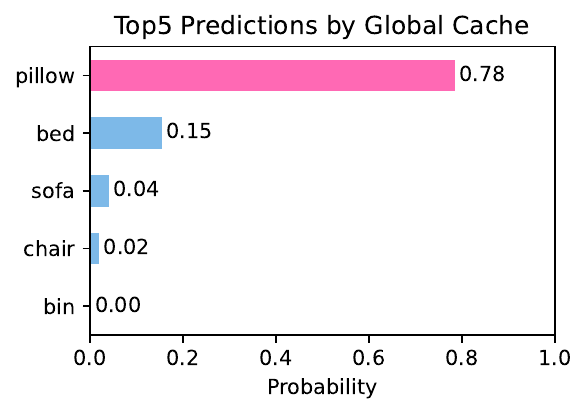}
     \caption{ULIP-2+\textbf{GC}.}
     \label{fig:vis_ulip2_sonn_hardest_global_bed}
  \end{subfigure}%
  \begin{subfigure}{0.25\textwidth}
     \centering
     \includegraphics[width=\columnwidth]{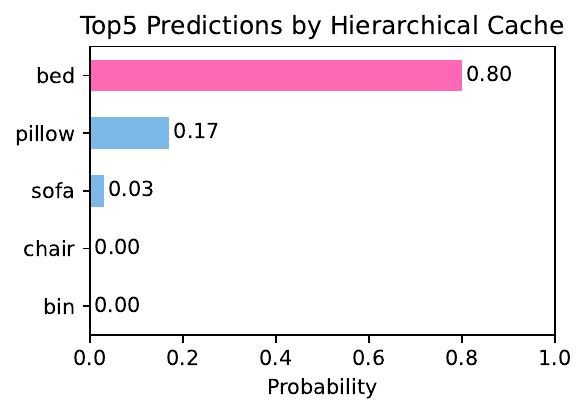}
     \caption{ULIP-2+\textbf{HC}.}
     \label{fig:vis_ulip2_sonn_hardest_hierar_bed}
  \end{subfigure}%

  \begin{subfigure}{0.25\textwidth}
    \centering
    \includegraphics[width=\columnwidth]{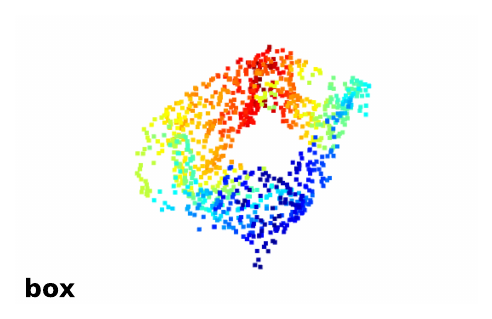}
    \caption{Ground truth.}
    \label{fig:vis_ulip2_sonn_hardest_gt_box}
  \end{subfigure}%
  \begin{subfigure}{0.25\textwidth}
      \centering
      \includegraphics[width=\columnwidth]{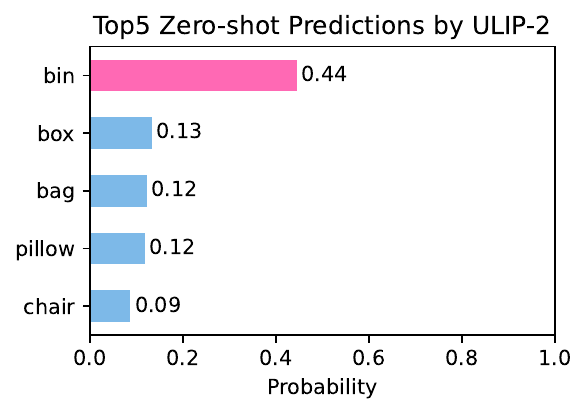}
      \caption{ULIP-2.}
      \label{fig:vis_ulip2_sonn_hardest_zero_box}
  \end{subfigure}%
  \begin{subfigure}{0.25\textwidth}
      \centering
      \includegraphics[width=\columnwidth]{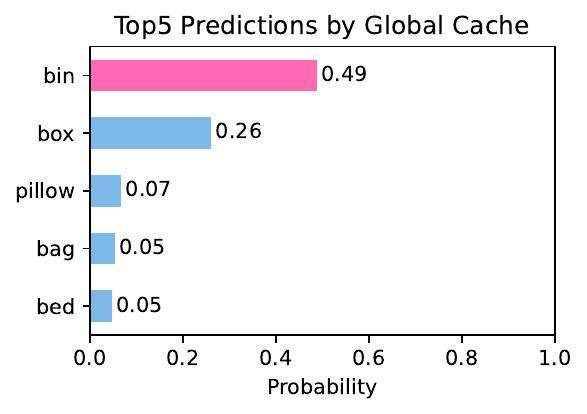}
      \caption{ULIP-2+\textbf{GC}.}
      \label{fig:vis_ulip2_sonn_hardest_global_box}
  \end{subfigure}%
  \begin{subfigure}{0.25\textwidth}
      \centering
      \includegraphics[width=\columnwidth]{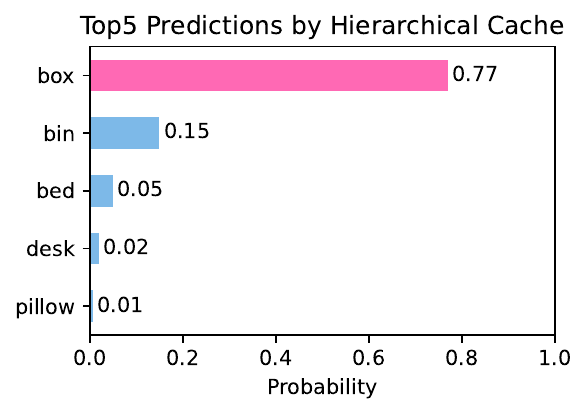}
      \caption{ULIP-2+\textbf{HC}.}
      \label{fig:vis_ulip2_sonn_hardest_hierar_box}
  \end{subfigure}%

  \begin{subfigure}{0.25\textwidth}
    \centering
    \includegraphics[width=\columnwidth]{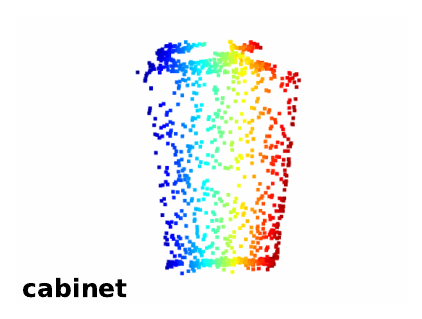}
    \caption{Ground truth.}
    \label{fig:ulip2_sonn_hardest_gt_cabinet}
  \end{subfigure}%
  \begin{subfigure}{0.25\textwidth}
      \centering
      \includegraphics[width=\columnwidth]{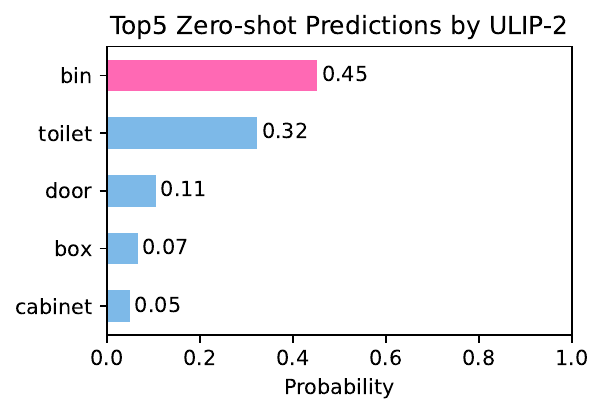}
      \caption{ULIP-2.}
      \label{fig:vis_ulip2_sonn_hardest_zero_cabinet}
  \end{subfigure}%
  \begin{subfigure}{0.25\textwidth}
      \centering
      \includegraphics[width=\columnwidth]{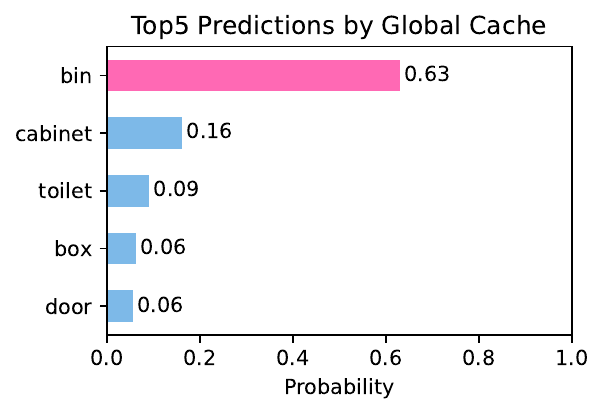}
      \caption{ULIP-2+\textbf{GC}.}
      \label{fig:vis_ulip2_sonn_hardest_global_cabinet}
  \end{subfigure}%
  \begin{subfigure}{0.25\textwidth}
      \centering
      \includegraphics[width=\columnwidth]{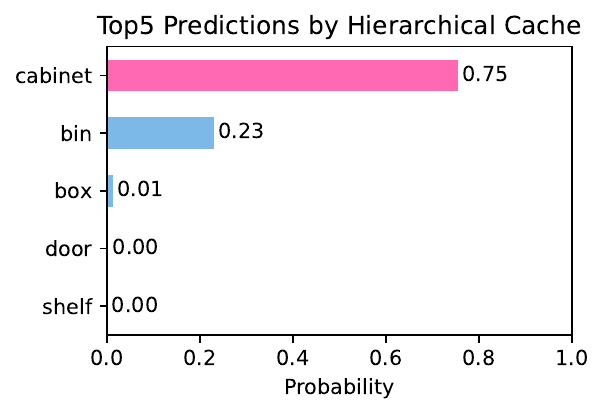}
      \caption{ULIP-2+\textbf{HC}.}
      \label{fig:vis_ulip2_sonn_hardest_hierar_cabinet}
  \end{subfigure}%
  \caption{\textbf{ULIP-2 zero-shot predictions before and after adaptation by Point-Cache.} The used dataset is S-PB\_T50\_RS. Each 3D object contains 1,024 points. 
  \textbf{GC}: global cache. \textbf{HC}: hierarchical cache.}
  \label{fig:vis_lm3d_dataset_adaptation_suppl_ulip2}
\end{figure*}

\begin{figure*}[b]
  \centering
  \begin{subfigure}{0.25\textwidth}
      \centering
      \includegraphics[width=\columnwidth]{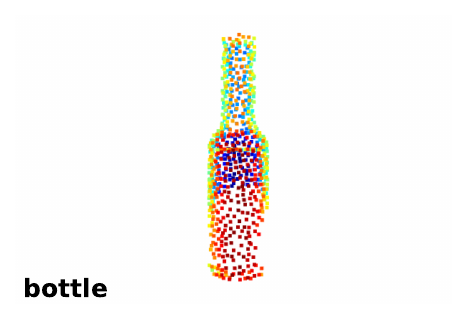}
      \caption{Ground truth.}
      \label{fig:vis_os_mn_c_dropout_local_gt_bottle}
  \end{subfigure}%
  \begin{subfigure}{0.25\textwidth}
      \centering
      \includegraphics[width=\columnwidth]{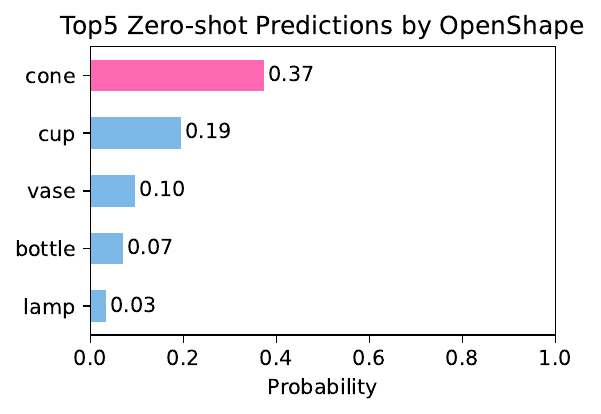}
      \caption{OpenShape.}
      \label{fig:vis_os_mn_c_dropout_local_zero_bottle}
  \end{subfigure}%
  \begin{subfigure}{0.25\textwidth}
      \centering
      \includegraphics[width=\columnwidth]{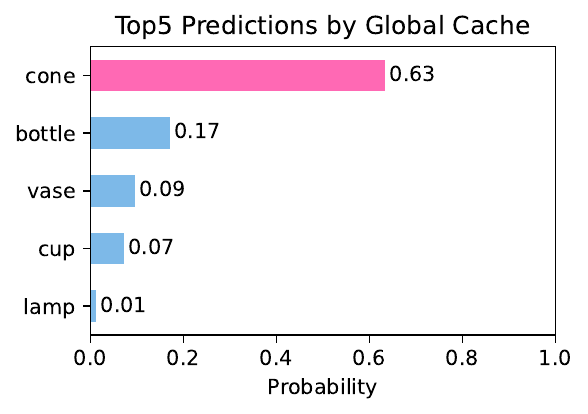}
      \caption{OpenShape+\textbf{GC}.}
      \label{fig:vis_os_mn_c_dropout_local_global_bottle}
  \end{subfigure}%
  \begin{subfigure}{0.25\textwidth}
      \centering
      \includegraphics[width=\columnwidth]{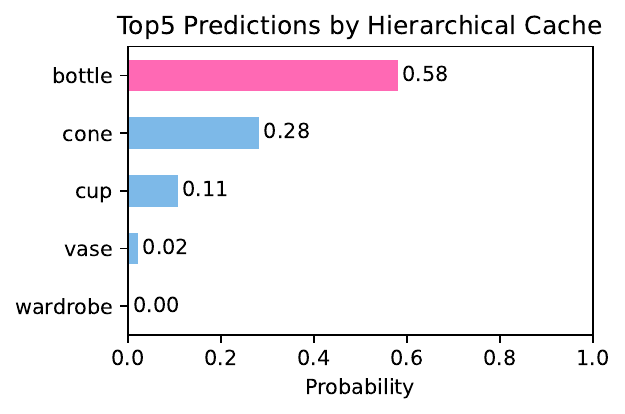}
      \caption{OpenShape+\textbf{HC}.}
      \label{fig:vis_os_mn_c_dropout_local_hierar_bottle}
  \end{subfigure}%

  \begin{subfigure}{0.25\textwidth}
    \centering
    \includegraphics[width=\columnwidth]{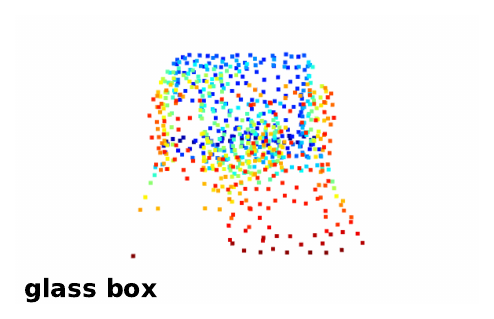}
    \caption{Ground truth.}
    \label{fig:vis_os_mn_c_dropout_local_gt_glass_box}
  \end{subfigure}%
  \begin{subfigure}{0.25\textwidth}
      \centering
      \includegraphics[width=\columnwidth]{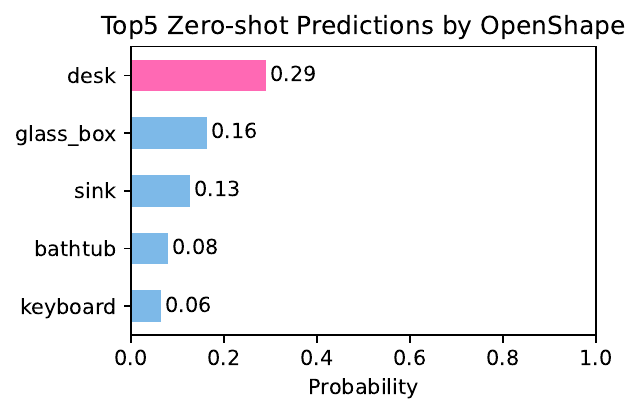}
      \caption{OpenShape.}
      \label{fig:vis_os_mn_c_dropout_local_zero_glass_box}
  \end{subfigure}%
  \begin{subfigure}{0.25\textwidth}
      \centering
      \includegraphics[width=\columnwidth]{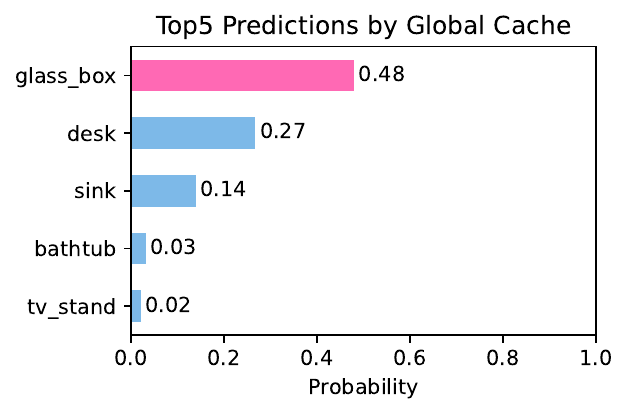}
      \caption{OpenShape+\textbf{GC}.}
      \label{fig:vis_os_mn_c_dropout_local_global_glass_box}
  \end{subfigure}%
  \begin{subfigure}{0.25\textwidth}
      \centering
      \includegraphics[width=\columnwidth]{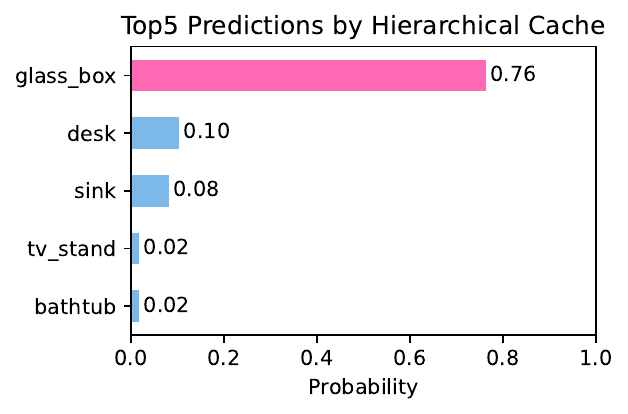}
      \caption{OpenShape+\textbf{HC}.}
      \label{fig:vis_os_mn_c_dropout_local_hierar_glass_box}
  \end{subfigure}%

  \begin{subfigure}{0.25\textwidth}
    \centering
    \includegraphics[width=\columnwidth]{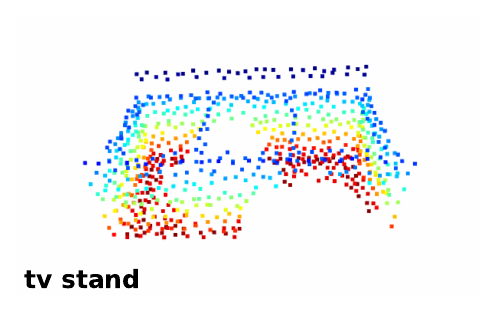}
    \caption{Ground truth.}
    \label{fig:vis_os_mn_c_dropout_local_gt_tv_stand}
  \end{subfigure}%
  \begin{subfigure}{0.25\textwidth}
      \centering
      \includegraphics[width=\columnwidth]{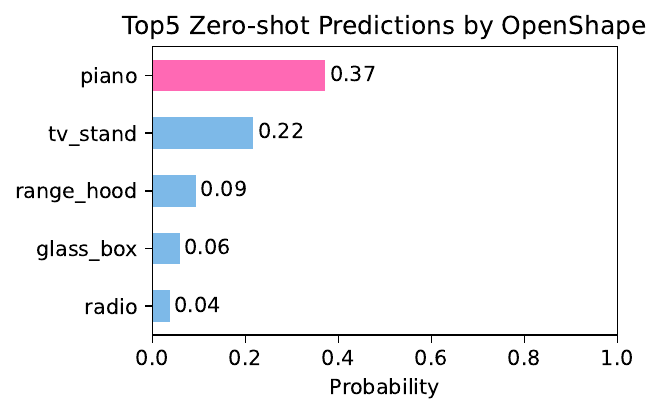}
      \caption{OpenShape.}
      \label{fig:vis_os_mn_c_dropout_local_zero_tv_stand}
  \end{subfigure}%
  \begin{subfigure}{0.25\textwidth}
      \centering
      \includegraphics[width=\columnwidth]{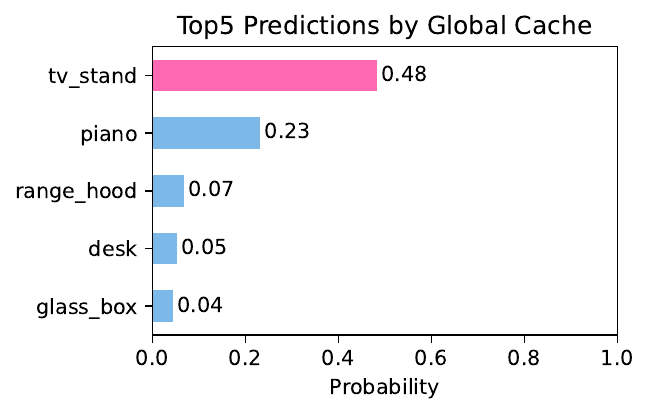}
      \caption{OpenShape+\textbf{GC}.}
      \label{fig:vis_os_mn_c_dropout_local_global_tv_stand}
  \end{subfigure}%
  \begin{subfigure}{0.25\textwidth}
      \centering
      \includegraphics[width=\columnwidth]{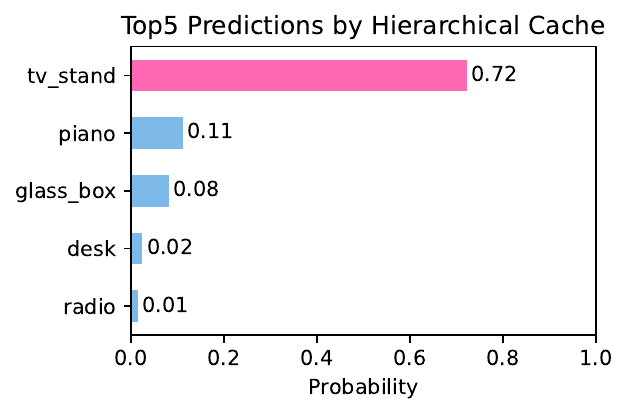}
      \caption{OpenShape+\textbf{HC}.}
      \label{fig:vis_os_mn_c_dropout_local_hierar_tv_stand}
  \end{subfigure}%

  \begin{subfigure}{0.25\textwidth}
     \centering
     \includegraphics[width=\columnwidth]{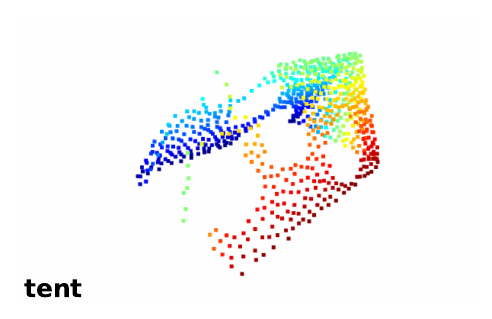}
     \caption{Ground truth.}
     \label{fig:vis_os_mn_c_dropout_local_gt_tent}
  \end{subfigure}%
  \begin{subfigure}{0.25\textwidth}
     \centering
     \includegraphics[width=\columnwidth]{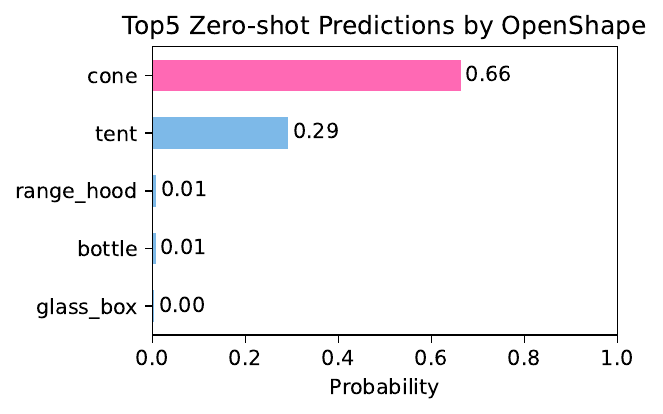}
     \caption{OpenShape.}
     \label{fig:vis_os_mn_c_dropout_local_zero_tent}
  \end{subfigure}%
  \begin{subfigure}{0.25\textwidth}
     \centering
     \includegraphics[width=\columnwidth]{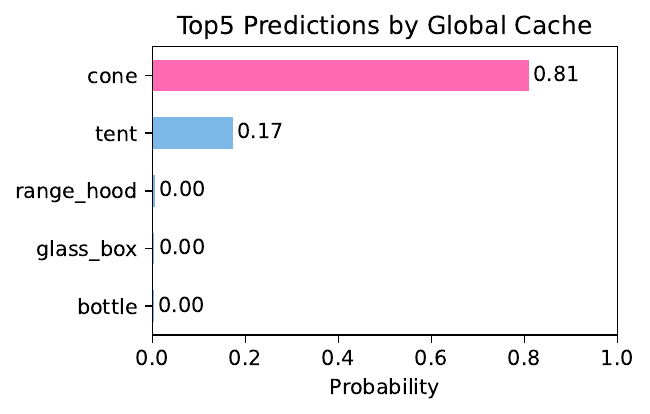}
     \caption{OpenShape+\textbf{GC}.}
     \label{fig:vis_os_mn_c_dropout_local_global_tent}
  \end{subfigure}%
  \begin{subfigure}{0.25\textwidth}
     \centering
     \includegraphics[width=\columnwidth]{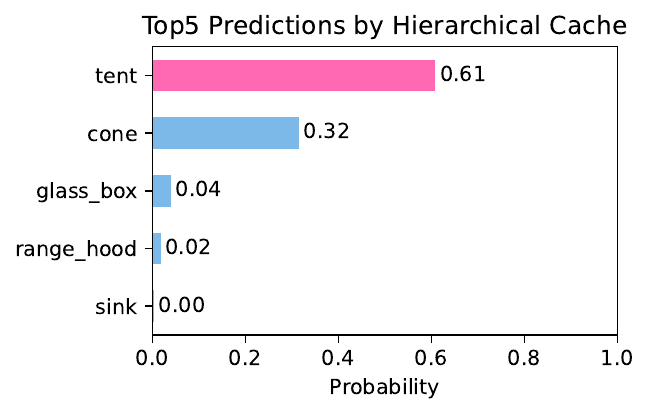}
     \caption{OpenShape+\textbf{HC}.}
     \label{fig:vis_os_mn_c_dropout_local_hierar_tent}
  \end{subfigure}%

  \begin{subfigure}{0.25\textwidth}
      \centering
      \includegraphics[width=\columnwidth]{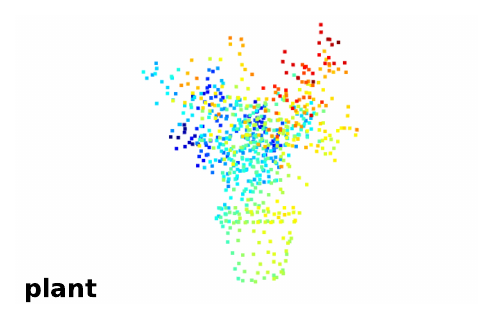}
      \caption{Ground truth.}
      \label{fig:vis_os_mn_c_dropout_local_gt_plant}
  \end{subfigure}%
  \begin{subfigure}{0.25\textwidth}
      \centering
      \includegraphics[width=\columnwidth]{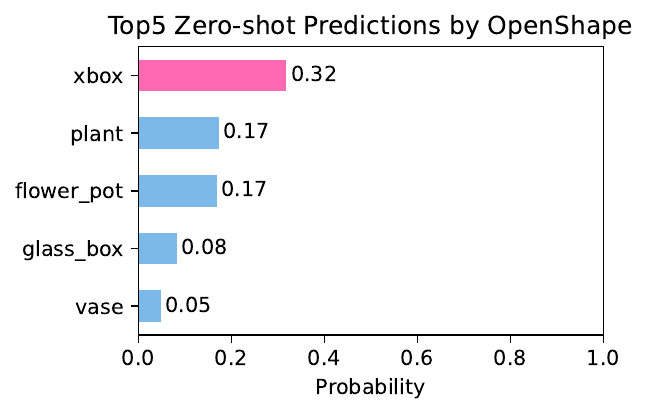}
      \caption{OpenShape.}
      \label{fig:vis_os_mn_c_dropout_local_zero_plant}
  \end{subfigure}%
  \begin{subfigure}{0.25\textwidth}
      \centering
      \includegraphics[width=\columnwidth]{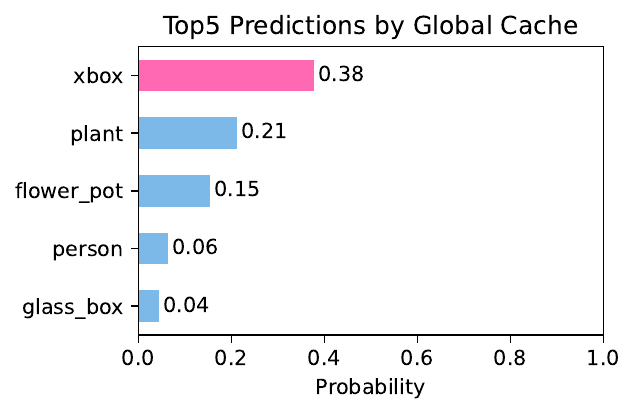}
      \caption{OpenShape+\textbf{GC}.}
      \label{fig:vis_os_mn_c_dropout_local_global_plant}
  \end{subfigure}%
  \begin{subfigure}{0.25\textwidth}
      \centering
      \includegraphics[width=\columnwidth]{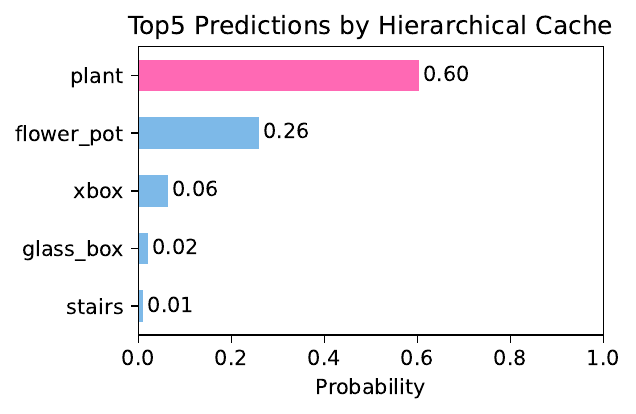}
      \caption{OpenShape+\textbf{HC}.}
      \label{fig:vis_os_mn_c_dropout_local_hierar_plant}
  \end{subfigure}%

  \begin{subfigure}{0.25\textwidth}
    \centering
    \includegraphics[width=\columnwidth]{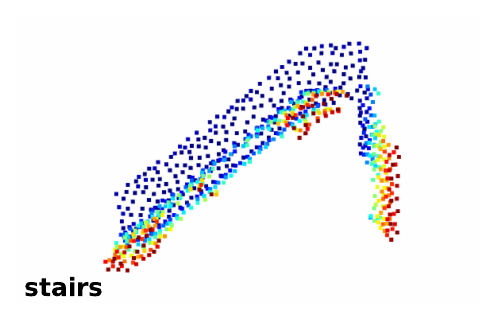}
    \caption{Ground truth.}
    \label{fig:vis_os_mn_c_dropout_local_gt_stairs}
  \end{subfigure}%
  \begin{subfigure}{0.25\textwidth}
      \centering
      \includegraphics[width=\columnwidth]{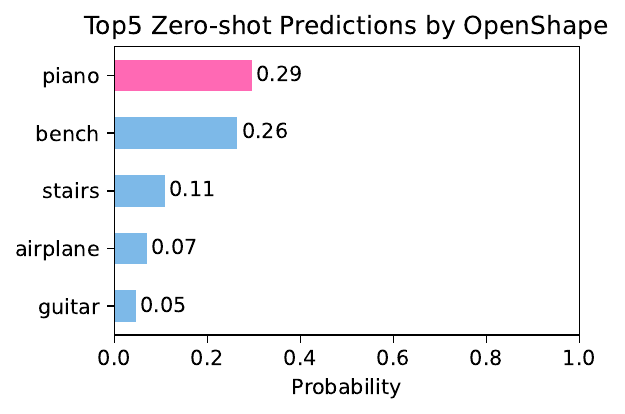}
      \caption{OpenShape.}
      \label{fig:vis_os_mn_c_dropout_local_zero_stairs}
  \end{subfigure}%
  \begin{subfigure}{0.25\textwidth}
      \centering
      \includegraphics[width=\columnwidth]{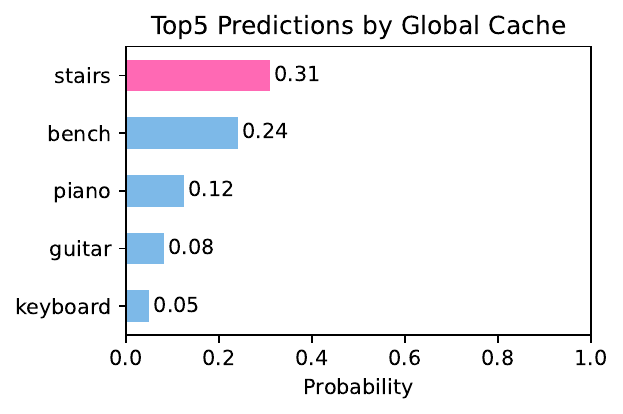}
      \caption{OpenShape+\textbf{GC}.}
      \label{fig:vis_os_mn_c_dropout_local_global_stairs}
  \end{subfigure}%
  \begin{subfigure}{0.25\textwidth}
      \centering
      \includegraphics[width=\columnwidth]{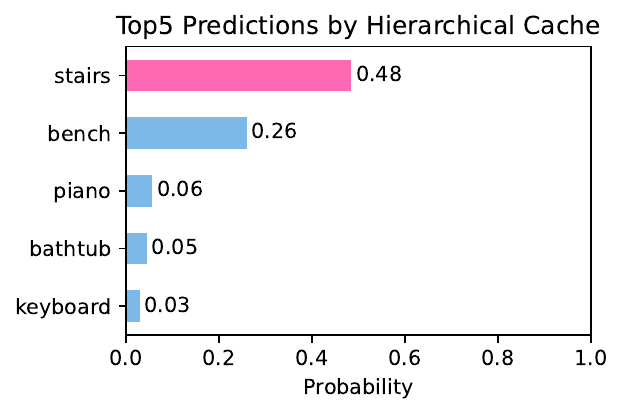}
      \caption{OpenShape+\textbf{HC}.}
      \label{fig:vis_os_mn_c_dropout_local_hierar_stairs}
  \end{subfigure}%

  \caption{\textbf{OpenShape zero-shot predictions before and after adaptation by Point-Cache.} The used dataset is ModelNet-C (drop\_local, severity=2). Each 3D object contains 1,024 points. 
  \textbf{GC}: global cache. \textbf{HC}: hierarchical cache.}
  \label{fig:vis_lm3d_dataset_adaptation_suppl_os}
\end{figure*}

\begin{figure*}[b]
  \centering
  \begin{subfigure}{0.25\textwidth}
      \centering
      \includegraphics[width=\columnwidth]{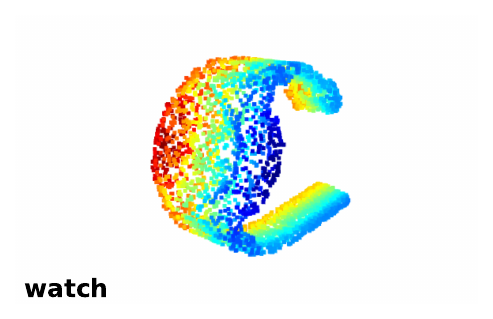}
      \caption{Ground truth.}
      \label{fig:vis_uni3d_omni3d_4096_gt_watch}
  \end{subfigure}%
  \begin{subfigure}{0.25\textwidth}
      \centering
      \includegraphics[width=\columnwidth]{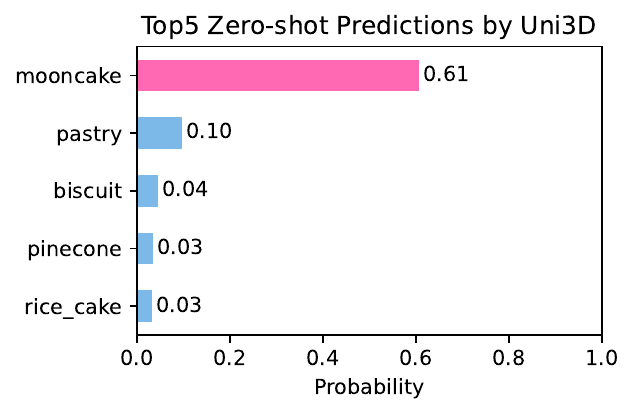}
      \caption{Uni3D.}
      \label{fig:vis_uni3d_omni3d_4096_zero_watch}
  \end{subfigure}%
  \begin{subfigure}{0.25\textwidth}
      \centering
      \includegraphics[width=\columnwidth]{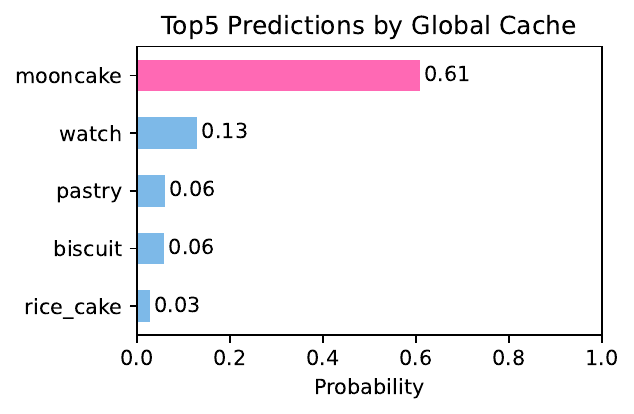}
      \caption{Uni3D+\textbf{GC}.}
      \label{fig:vis_uni3d_omni3d_4096_global_watch}
  \end{subfigure}%
  \begin{subfigure}{0.25\textwidth}
      \centering
      \includegraphics[width=\columnwidth]{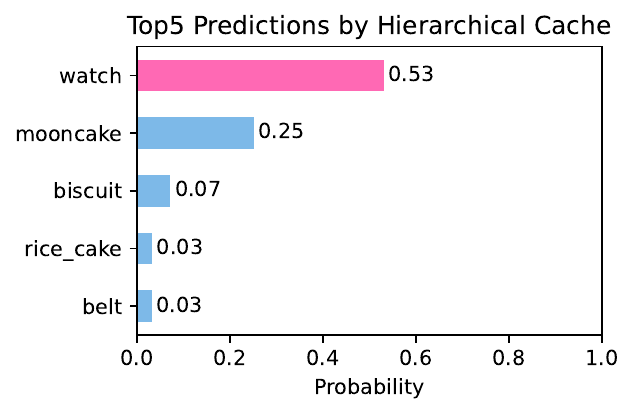}
      \caption{Uni3D+\textbf{HC}.}
      \label{fig:vis_uni3d_omni3d_4096_hierar_watch}
  \end{subfigure}%

  \begin{subfigure}{0.25\textwidth}
    \centering
    \includegraphics[width=\columnwidth]{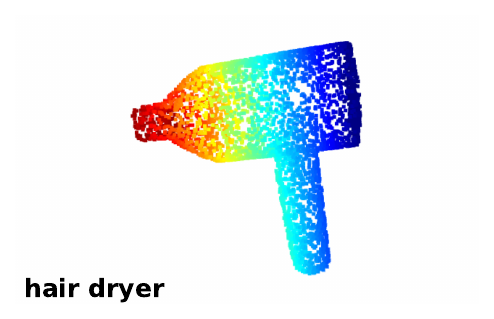}
    \caption{Ground truth.}
    \label{fig:uni3d_omni3d_4096_gt_hair_dryer}
  \end{subfigure}%
  \begin{subfigure}{0.25\textwidth}
      \centering
      \includegraphics[width=\columnwidth]{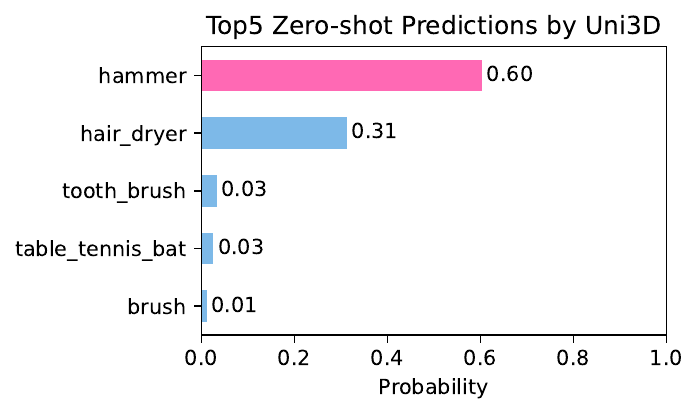}
      \caption{Uni3D.}
      \label{fig:vis_uni3d_omni3d_4096_zero_hair_dryer}
  \end{subfigure}%
  \begin{subfigure}{0.25\textwidth}
      \centering
      \includegraphics[width=\columnwidth]{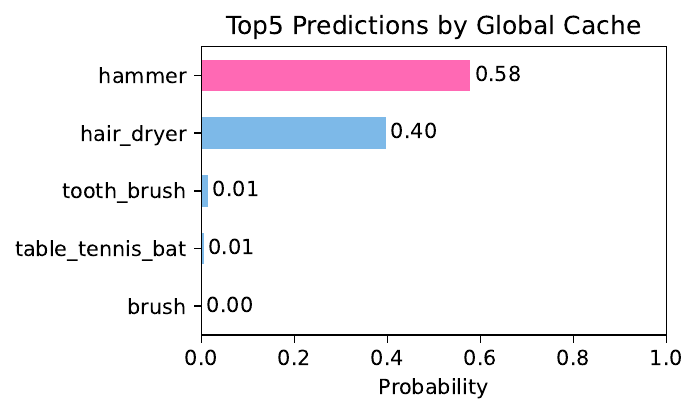}
      \caption{Uni3D+\textbf{GC}.}
      \label{fig:vis_uni3d_omni3d_4096_global_hair_dryer}
  \end{subfigure}%
  \begin{subfigure}{0.25\textwidth}
      \centering
      \includegraphics[width=\columnwidth]{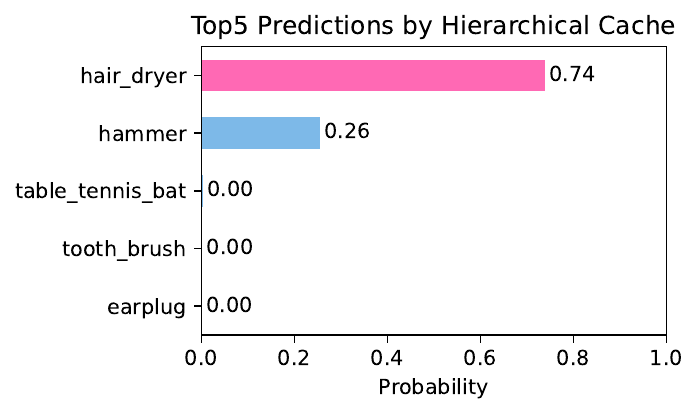}
      \caption{Uni3D+\textbf{HC}.}
      \label{fig:vis_uni3d_omni3d_4096_hierar_hair_dryer}
  \end{subfigure}%

  \begin{subfigure}{0.25\textwidth}
    \centering
    \includegraphics[width=\columnwidth]{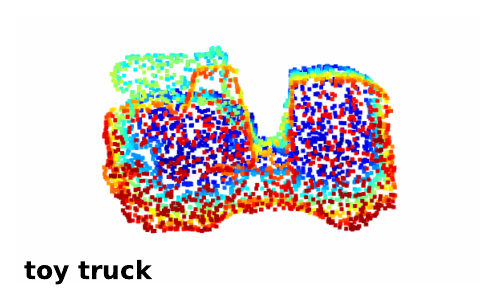}
    \caption{Ground truth.}
    \label{fig:vis_uni3d_omni3d_4096_gt_toy_truck}
  \end{subfigure}%
  \begin{subfigure}{0.25\textwidth}
      \centering
      \includegraphics[width=\columnwidth]{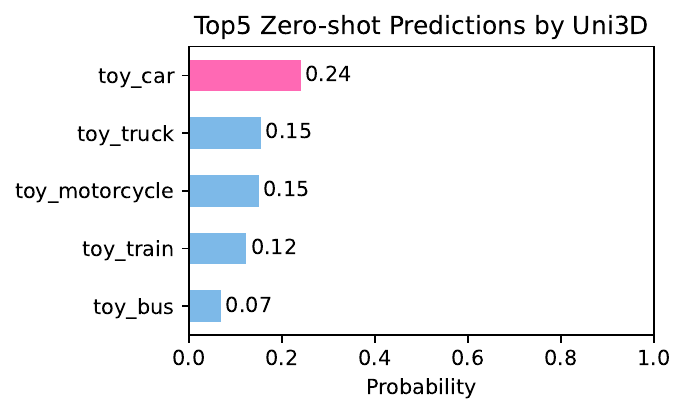}
      \caption{Uni3D.}
      \label{fig:vis_uni3d_omni3d_4096_zero_toy_truck}
  \end{subfigure}%
  \begin{subfigure}{0.25\textwidth}
      \centering
      \includegraphics[width=\columnwidth]{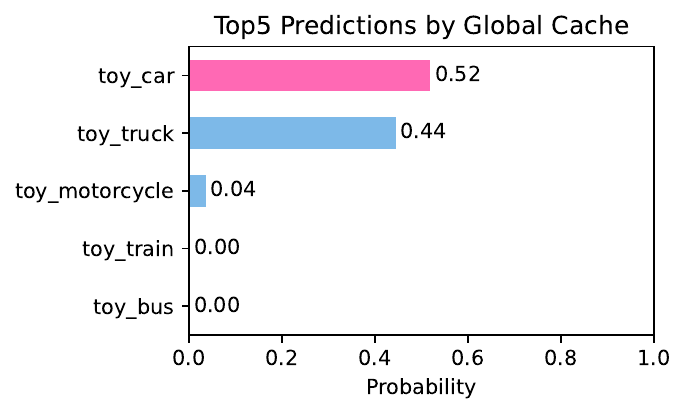}
      \caption{Uni3D+\textbf{GC}.}
      \label{fig:vis_uni3d_omni3d_4096_global_toy_truck}
  \end{subfigure}%
  \begin{subfigure}{0.25\textwidth}
      \centering
      \includegraphics[width=\columnwidth]{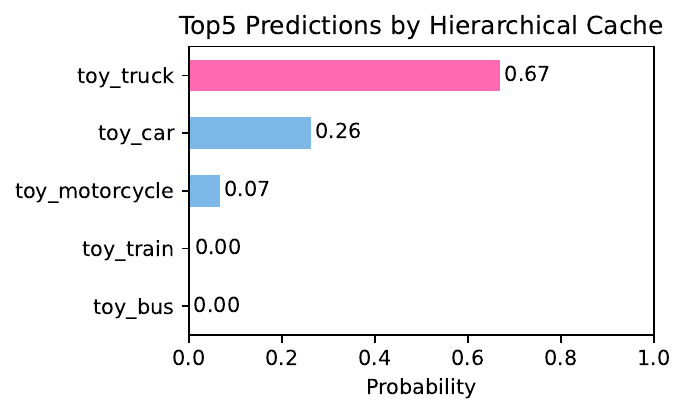}
      \caption{Uni3D+\textbf{HC}.}
      \label{fig:vis_uni3d_omni3d_4096_hierar_toy_truck}
  \end{subfigure}%

  \begin{subfigure}{0.25\textwidth}
     \centering
     \includegraphics[width=\columnwidth]{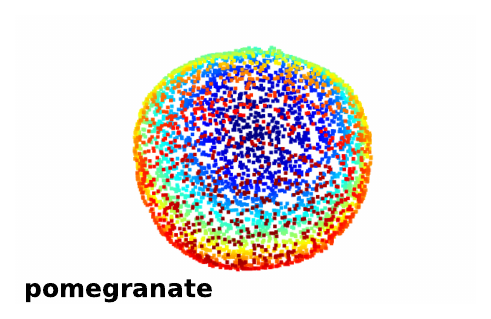}
     \caption{Ground truth.}
     \label{fig:vis_uni3d_omni3d_4096_gt_pomegranate}
  \end{subfigure}%
  \begin{subfigure}{0.25\textwidth}
     \centering
     \includegraphics[width=\columnwidth]{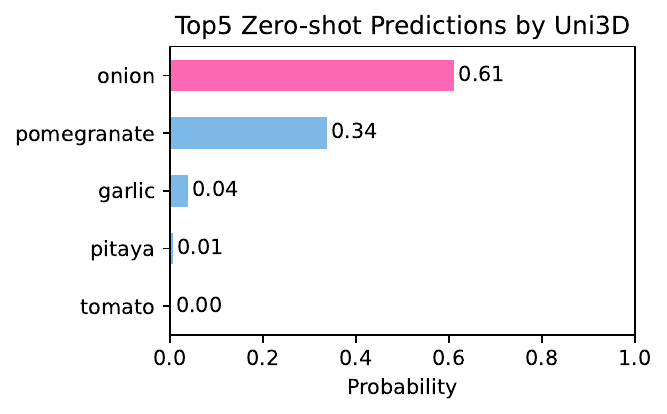}
     \caption{Uni3D.}
     \label{fig:vis_uni3d_omin3d_4096_zero_pomegranate}
  \end{subfigure}%
  \begin{subfigure}{0.25\textwidth}
     \centering
     \includegraphics[width=\columnwidth]{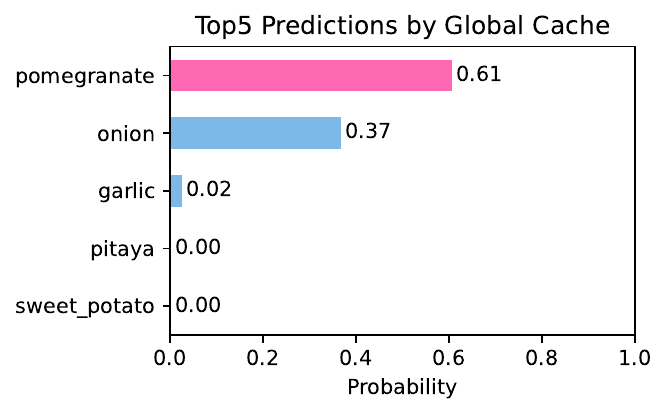}
     \caption{Uni3D+\textbf{GC}.}
     \label{fig:vis_uni3d_omni3d_4096_global_pomegranate}
  \end{subfigure}%
  \begin{subfigure}{0.25\textwidth}
     \centering
     \includegraphics[width=\columnwidth]{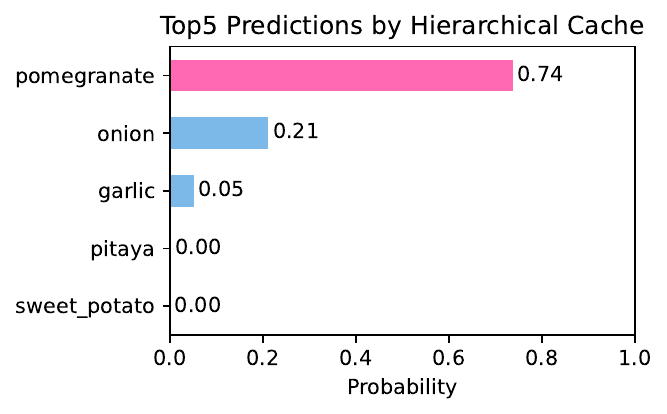}
     \caption{Uni3D+\textbf{HC}.}
     \label{fig:vis_uni3d_omni3d_4096_hierar_pomegranate}
  \end{subfigure}%

  \begin{subfigure}{0.25\textwidth}
      \centering
      \includegraphics[width=\columnwidth]{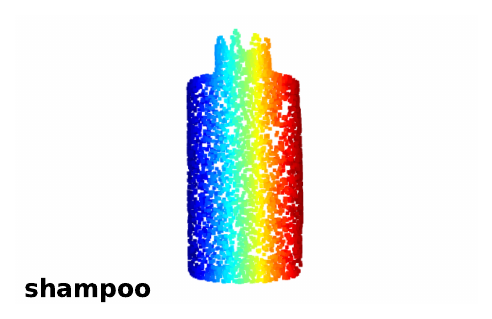}
      \caption{Ground truth.}
      \label{fig:vis_uni3d_omni3d_4096_gt_shampoo}
  \end{subfigure}%
  \begin{subfigure}{0.25\textwidth}
      \centering
      \includegraphics[width=\columnwidth]{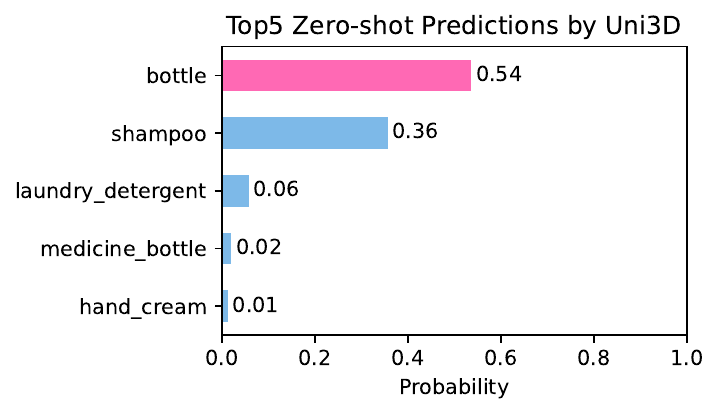}
      \caption{Uni3D.}
      \label{fig:vis_uni3d_omin3d_4096_zero_shampoo}
  \end{subfigure}%
  \begin{subfigure}{0.25\textwidth}
      \centering
      \includegraphics[width=\columnwidth]{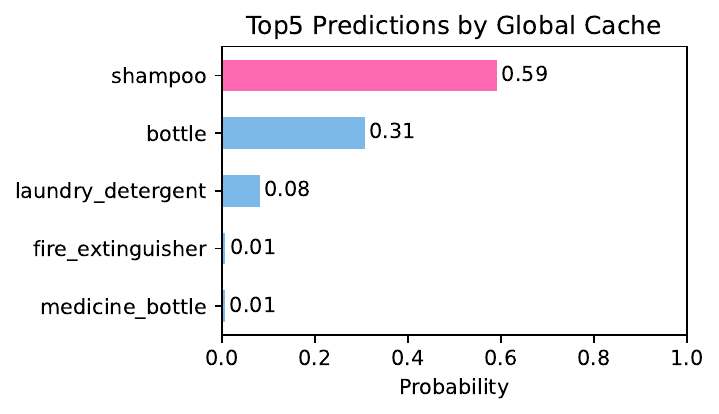}
      \caption{Uni3D+\textbf{GC}.}
      \label{fig:vis_uni3d_omni3d_4096_global_shampoo}
  \end{subfigure}%
  \begin{subfigure}{0.25\textwidth}
      \centering
      \includegraphics[width=\columnwidth]{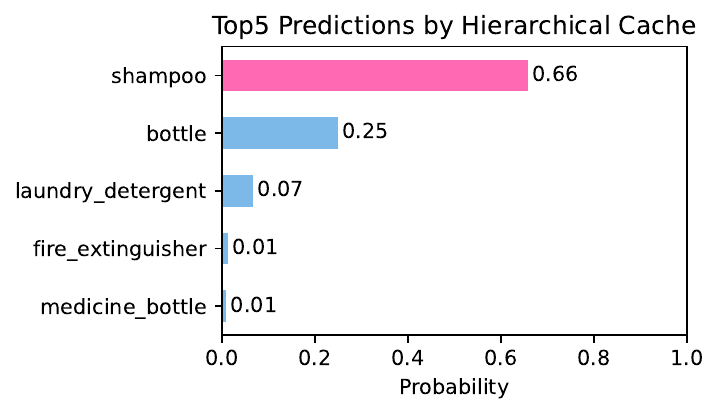}
      \caption{Uni3D+\textbf{HC}.}
      \label{fig:vis_uni3d_omni3d_4096_hierar_shampoo}
  \end{subfigure}%

  \begin{subfigure}{0.25\textwidth}
    \centering
    \includegraphics[width=\columnwidth]{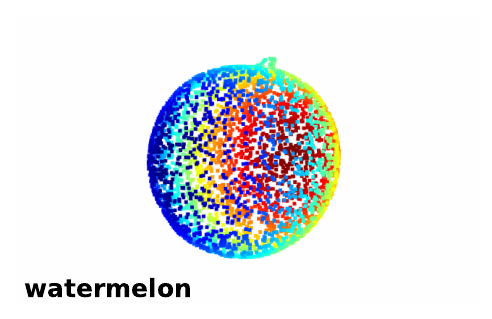}
    \caption{Ground truth.}
    \label{fig:vis_uni3d_omni3d_4096_gt_watermelon}
  \end{subfigure}%
  \begin{subfigure}{0.25\textwidth}
      \centering
      \includegraphics[width=\columnwidth]{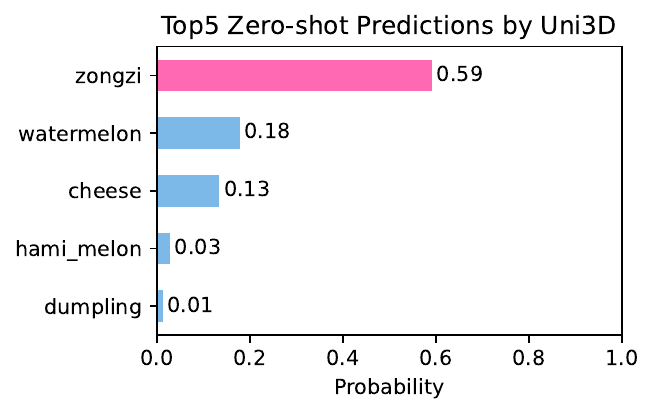}
      \caption{Uni3D.}
      \label{fig:vis_uni3d_omin3d_4096_zero_watermelon}
  \end{subfigure}%
  \begin{subfigure}{0.25\textwidth}
      \centering
      \includegraphics[width=\columnwidth]{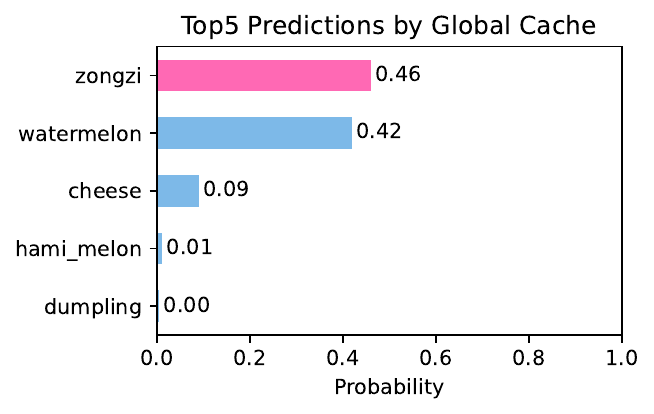}
      \caption{Uni3D+\textbf{GC}.}
      \label{fig:vis_uni3d_omni3d_4096_global_watermelon}
  \end{subfigure}%
  \begin{subfigure}{0.25\textwidth}
      \centering
      \includegraphics[width=\columnwidth]{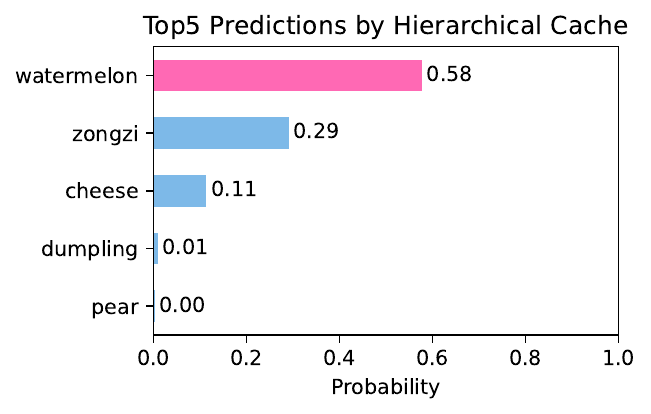}
      \caption{Uni3D+\textbf{HC}.}
      \label{fig:vis_uni3d_omni3d_4096_hierar_watermelon}
  \end{subfigure}%

  \caption{\textbf{Uni3D zero-shot predictions before and after adaptation by Point-Cache.} 
  The used dataset is Omni3D. Each 3D object contains 4,096 points. \textbf{GC}: global cache. \textbf{HC}: hierarchical cache.}
  \label{fig:vis_lm3d_dataset_adaptation_suppl_uni3d}
\end{figure*}

We provide additional qualitative examples to demonstrate the step-by-step adaptation process of various large 3D models with Point-Cache, 
exhibited in Fig.~\ref{fig:vis_lm3d_dataset_adaptation_suppl_ulip},
~\ref{fig:vis_lm3d_dataset_adaptation_suppl_ulip2},
~\ref{fig:vis_lm3d_dataset_adaptation_suppl_os} and 
~\ref{fig:vis_lm3d_dataset_adaptation_suppl_uni3d}. 
The results confirm that Point-Cache effectively assists large 3D models in correcting erroneous zero-shot predictions, 
reducing the classification entropy, and improving the recognition accuracy at test time. 
Notably, there are cases where the global cache model fails to adapt zero-shot predictions. 
For example, in the third row of Fig.~\ref{fig:vis_lm3d_dataset_adaptation_suppl_ulip2}, 
although ULIP-2+\textbf{GC} reduces the class probability for `sink' from 73\% to 53\%, it still identifies `sink' as the top-1 class. 
In contrast, ULIP-2+\textbf{HC} makes a sharp adjustment to the logits of ULIP-2+\textbf{GC} after incorporating local-part 
knowledge, promoting `toilet' to the top-1 class (from 41\% to 90\%) 
and achieving a successful correction. 
Similar adaptations are observed in 
Fig.~\ref{fig:vis_lm3d_dataset_adaptation_suppl_ulip} (1st, 2nd, 4th, and 5th rows), 
Fig.~\ref{fig:vis_lm3d_dataset_adaptation_suppl_os} (1st, 4th, and 5th rows), and  
Fig.~\ref{fig:vis_lm3d_dataset_adaptation_suppl_uni3d} (1st, 2nd, 3rd, and 6th rows), 
suggesting that the coarse-to-fine cache design is highly effective in capturing subtle differences among point cloud objects. 

\end{document}